\documentclass{article}

\PassOptionsToPackage{numbers, compress, square}{natbib}
\usepackage[final]{neurips_2022}
\usepackage[dvipsnames]{xcolor}
\newcommand{\beginsupplement}{%
        \setcounter{table}{0}
        \renewcommand{\thetable}{S\arabic{table}}%
        \setcounter{figure}{0}
        \renewcommand{\thefigure}{S\arabic{figure}}%
        \setcounter{subsection}{0}
        \renewcommand{\thesubsection}{S\arabic{subsection}}%
        
     }
\usepackage{enumitem}     
\usepackage{algorithm}
\usepackage{algpseudocode}
\usepackage{tikz}

\usepackage[utf8]{inputenc} 
\usepackage[T1]{fontenc}    
\usepackage{hyperref}       
\usepackage{url}            
\usepackage{booktabs}       
\usepackage{amsfonts}       
\usepackage{nicefrac}       
\usepackage{microtype}      
\usepackage{xcolor}         
\usepackage{wrapfig}
\usepackage{tikz}
\usepackage{caption}
\usepackage{subcaption}
\usepackage{amsmath, amssymb}
\usepackage{multirow}
\usepackage{array}
\usepackage{makecell}

\newcommand{\VAENS}{\textbf{\textcolor{BlueViolet}{VAE-NS}}}
\newcommand{\VAESTN}{\textbf{\textcolor{SkyBlue}{VAE-STN}}}
\newcommand{\DAGANUN}{\textbf{\textcolor{Mahogany}{DA-GAN-UN}}}
\newcommand{\DAGANRN}{\textbf{\textcolor{YellowOrange}{DA-GAN-RN}}}
\newcommand{\human}{\textbf{\textcolor{Gray}{human}}}

\def\checklist{1}

\DeclareMathOperator*{\argmin}{argmin}
\newcommand{\norm}[1]{\left\lVert#1\right\rVert}

\DeclareMathOperator*{\KL}{KL}

\title{Diversity vs. Recognizability: Human-like generalization in one-shot generative models}

\author{%
  Victor Boutin\textsuperscript{\textnormal{1,2}}, Lakshya Singhal\textsuperscript{\textnormal{2}}, Xavier Thomas\textsuperscript{\textnormal{2}} and  Thomas Serre\textsuperscript{\textnormal{1,2}}\\
  \textsuperscript{1} Artificial and Natural Intelligence Toulouse Institute, Universit\'e de Toulouse, France\\
   \textsuperscript{2} Carney Institute for Brain Science, Dpt. of Cognitive Linguistic \& Psychological Sciences \\ 
   Brown University, Providence, RI 02912\\
   \texttt{\{victor\_boutin, thomas\_serre\}@brown.edu}\\
}

\begin{document}

\maketitle

\begin{abstract}

Robust generalization to new concepts has long remained a distinctive feature of human intelligence. However, recent progress in deep generative models has now led to neural architectures capable of synthesizing novel instances of unknown visual concepts from a single training example. Yet, a more precise comparison between these models and humans is not possible because existing performance metrics for generative models (i.e., FID, IS, likelihood) are not appropriate for the one-shot generation scenario. Here, we propose a new framework to evaluate one-shot generative models along two axes: sample \emph{recognizability} vs. \emph{diversity}  (i.e., intra-class variability). Using this framework, we perform a systematic evaluation of representative one-shot generative models on the Omniglot handwritten dataset. We first show that GAN-like and VAE-like models fall on opposite ends of the diversity-recognizability space. Extensive analyses of the effect of key model parameters further revealed that spatial attention and context integration have a linear contribution to the diversity-recognizability trade-off. In contrast, disentanglement transports the model along a parabolic curve that could be used to maximize recognizability. Using the diversity-recognizability framework, we were able to identify models and parameters that closely approximate human data.

\end{abstract}
\section{Introduction}

Our ability to learn and generalize from a limited number of samples is a hallmark of human cognition. In language, scientists have long highlighted how little training data children need in comparison to the richness and complexity of the language they learn so efficiently to master~\citep{chomsky1965aspects, piattelli1980language}. Similarly, children and adults alike are able to learn novel object categories from as little as a single training example~\citep{feldman1997structure, broedelet2022school}. From a computational point of view, such feats are remarkable because they suggest that learners must be relying on inductive biases to overcome such challenges~\citep{lake2017building} -- from an ability to detect suspicious coincidences or 'non-accidental' features~\citep{richards1992features, tiedemann2021one} to exploiting the principle of compositionality~\citep{lake2015human, lake2017building}. 

While a common criticism of modern AI approaches is their reliance on large training datasets, progress in one-shot categorization has been significant. One-shot categorization involves predicting an image category based on a unique training sample per class. Multiple algorithms have been proposed including meta-learning algorithms~\citep{finn2017model,santoro2016meta, chowdhury2022meta, mishra2017simple} or metric-learning algorithms~\citep{snell2017prototypical,koch2015siamese, sung2018learning} that are now starting to approach human accuracy. Perhaps a less studied problem is the one-shot generation problem -- aimed at creating new variations of a prototypical shape seen only once. Since the seminal work of ~\citet{lake2015human} who introduced the Bayesian Program Learning algorithm, only a handful of promising one-shot generative algorithms have been proposed~\citep{rezende2016one, edwards2016towards, antoniou2017data} (see section~\ref{MAIN:models} for a more exhaustive description of prior work). 

Why have so few algorithms for one-shot image generation vs. image categorization been proposed? We argue that one of the main reasons for this lack of progress is the absence of an adequate evaluation metric. As of today, one-shot generative models are evaluated using methods initially developed for models producing samples that belong to the training categories and trained on large datasets. Those metrics include the likelihood, the FID (Frechet Inception Distance), or the IS (Inception Score). In the one-shot image generation scenario in which training images are scarce and the generated samples represent new visual concepts, the likelihood, the FID, and the IS are biased~\citep{barratt2018note, chong2020effectively, nalisnick2018deep} (see~\ref{MAIN:metric} for more details). These limitations urge us to look for new metrics tailored for one-shot image generation.

\begin{wrapfigure}{r}{0.45\linewidth}
\vspace{-20pt}
	\centering
\begin{tikzpicture}
\draw [anchor=north west] (0\linewidth, 1\linewidth) node {\includegraphics[width=\linewidth]{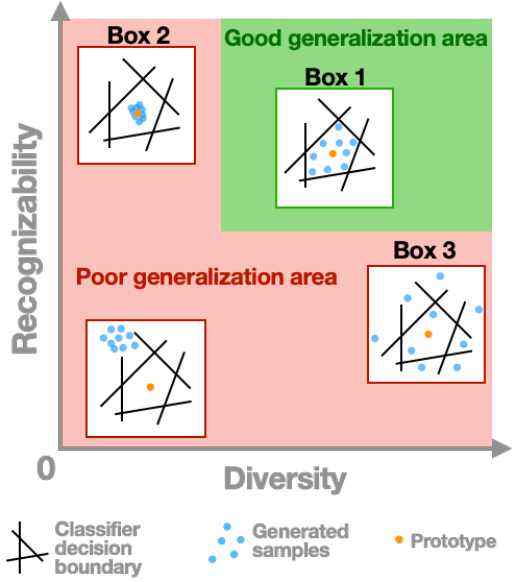}};
\end{tikzpicture}
\caption{{\bf The diversity vs. recognizability framework.} The best possible samples for good generalization (green area) are those that match the intra-class variations (i.e., remain within decision boundaries; Box~$1$). Bad samples associated with poor generalization (red area) include strategies that involve exact copies of the prototype (Box~$2$) and samples that exceed the intra-class variability (Box~$3$). }
\label{fig:fig1}
\vspace{0pt}
\end{wrapfigure}

Recent psychophysics work~\citep{tiedemann2021one} has characterized humans' ability for one-shot generation along two main axes: samples \emph{diversity} (i.e.,  intra-class variability) and samples \emph{recognizability} (i.e., how easy or hard they are to classify). According to this framework, ideal generalization corresponds to a combination of high recognizability and high diversity. As illustrated in Fig.~\ref{fig:fig1}, an ideal model should be able to generate samples that span the entire space within the decision boundary of a classifier (Box~$1$). In comparison, the model of Box~$2$ has learned to make identical copies of the prototype (i.e., low diversity but high accuracy). Such a model has failed to generalize the visual concept exemplified by the prototype. Similarly, if the model's samples are so diverse that they cannot be recognized accurately as shown in the Box~$3$ of Fig.~\ref{fig:fig1}, then the generated samples won't look like the prototype. 

Here, we borrow from this work and adapt it to create the first framework to evaluate and compare humans and one-shot generative models. 
Using this framework, we systematically evaluate an array of representative one-shot generative models on the Omniglot dataset~\citep{lake2015human}. We show that GAN-like and VAE-like one-shot generative models fall on opposite ends of the diversity-recognizability space: GAN-like models fall on the high recognizability --- low diversity end of the space while VAE-like models fall on the low recognizability --- high diversity end of the space. We further study some key model parameters that modulate spatial attention, context integration, and disentanglement. Our results suggest that spatial attention and context have an (almost) linear effect on the diversity vs. recognizability trade-off. In contrast, varying the disentanglement moves the models on a parabolic curve that could be used to maximize the recognizability. 
Last but not least, we have leveraged the diversity vs. recognizability space to identify models and parameters that best approximate the human data on the Omniglot handwritten dataset.

\section{Related work}
\subsection{Metrics to evaluate generative models and their limitations for one-shot generation tasks}
\label{MAIN:metric}
Different types of generative models are typically evaluated using different metrics. On the one hand, likelihood-based algorithms (e.g., VAE~\citep{kingma2013auto}, PixelCNN~\citep{van2016conditional}, GLOW~\citep{kingma2018glow}, etc.) are evaluated using their own objective function applied on a testing set. Likelihood provides a direct estimate of the KL divergence between the data points and the model's samples. On the other hand, implicit generative models such as Generative Adversarial Networks (GANs)~\citep{goodfellow2014generative} for which the loss function cannot be used, are typically evaluated using other scores such as the Inception Score (IS)~\citep{salimans2016improved} or Frechet Inception Distance (FID)~\citep{heusel2017gans}. IS and FID are heuristic measures used to aggregate both the sample quality and diversity in one single score. The IS scores a sample quality according to the confidence with which an Inception v3 Net~\citep{szegedy2016rethinking} assigns the correct class label to it. The FID score is the Wasserstein-2 distance between two Gaussian distributions: one fitted on the features of the data distribution and the other on the features of the model distribution (the features are also extracted from an Inception v3 Network). 

All these metrics are problematic for the one-shot generation scenario for $2$ main reasons that are intrinsically related to the task: the low number of samples per class, and the dissimilarity between training and testing visual concepts. IS and FID rely on statistical distances (either KL divergence for IS or Wasserstein-2 distance for FID) that require a high number of data points $N$ to produce an unbiased estimator of the distance. Even when used in the traditional settings (i.e., $N=50000$), it has been demonstrated that both scores are biased~\citep{chong2020effectively}. This is to be compared with the $N=20$ samples typically available in popular few-shot learning datasets such as Omniglot~\citep{lake2015human}. Another problem caused by the limited number of samples per class in the training set is the overfitting of the Inception Net used to extract the features to compute the IS and FID~\citep{brigato2021close}. To illustrate this phenomenon we have conducted a small control experiment in which we have trained a standard classifier to recognize images from the Omniglot datasets (see~\ref{SI:Overfitting}). In this experiment, we have used 18 samples per classes for training and 2 samples per classes for testing. In Fig. \ref{SI:Overfitting_lowdata}a, we observe an increase of the testing loss while the training loss is decreasing. This is a clear sign of overfitting. Note that this overfitting is not happening when the standard classifier is replaced by a one-shot (or few-shot) classifier. This control experiment show that standard classifier are not adapted to extract relevant features in the low-data regime. Consequently, IS and FID are not suitable in the low-data regime.

The second limitation of these metrics appears because the training and the testing samples are too dissimilar. Likelihood scores are known to yield higher scores for out-of-domain data compared to in-domain data~\citep{nalisnick2018deep}. Therefore, the evaluation of the novel visual concepts generated by one-shot generative models will be biased toward higher scores. In addition, both FID and IS rely on distance between features extracted by an Inception Net which comes with no guarantee that it will produce meaningful features for novel categories. For example, class misalignment has been reported when the Inception Net was trained on ImageNet and tested on CIFAR10~\citep{barratt2018note}. Because of all the aforementioned limitations, it is pretty clear that new procedures are needed to evaluate the performance of few-shot generative algorithms.

\subsection{One-shot generative models}
\label{MAIN:models}
One can distinguish between two broad classes of one-shot generative models: structured models and statistical models~\citep{feinman2020generating}. Structured models have strong inductive biases and rigid parametric assumptions based on a priori knowledge such as for example a given hierarchy of features, a known grammar or program~\citep{salakhutdinov2012one}. A prominent example of a structured model includes the very first algorithm for one-shot image generation, the Bayesian Program Learning (BPL) model~\citep{lake2015human}. Statistical models learn visual concepts by learning statistical regularities between observed patterns~\citep{rezende2016one, edwards2016towards, giannone2021hierarchical}. Here, we focus on representative architectures of one-shot generative statistical models, which we summarize below.
 
 \begin{itemize}[leftmargin=7mm]
    \item VAE with Spatial Transformer Network (\VAESTN)~\citep{rezende2016one}. The VAE-STN is a sequential and conditional Variational Auto-Encoder (VAE) constructing images iteratively. The VAE-STN algorithm uses a recurrent neural network (i.e., an LSTM) to encode the sequence of local patches extracted by an attentional module. A key ingredient of the  VAE-STN is an attention module composed of a Spatial Transformer Network (STN)~\citep{jaderberg2015spatial} to learn to shift attention to different locations of the input image. The STN is a trainable module to learn all possible affine transformations (i.e., translation, scaling, rotation, shearing) of an input image (see~\ref{SI:VAE_STN_details} for samples and details of the \VAESTN).
    \item Neural statistician (\textbf{\textcolor{BlueViolet}{VAE-NS}})~\citep{edwards2016towards}: The Neural Statistician is an extension of the conditional VAE model including contextual information. Therefore, in addition to learning an approximate inference network over latent variables for every image in the set (as done in a VAE), the approximate inference is also implemented over another latent variable, called the context variable, that is specific to the considered visual concept. The context inference network is fed with a small set of images representing variations of a given visual concept. The \VAENS~has been extended to include attention and hierarchical factorization of the generative process~\citep{giannone2021hierarchical} (see~\ref{SI:NS_details} for samples and details of the \VAENS).
    
    \item Data-Augmentation GAN  (DA-GAN)~\citep{antoniou2017data}: Data-Augmentation GAN is a generative adversarial network conditioned on a prototype image. The DA-GAN generator is fed with a concatenation of a vector drawn from a normal distribution and a compressed representation of the prototype. The discriminator is trained to differentiate images produced by the generator from images of the dataset, while the generator has to fool the discriminator. We have trained $2$ different DA-GAN, one is based on the U-Net architecture (\DAGANUN) and the other one on the ResNet architecture (\DAGANRN) (see~\ref{SI:DAGAN_details} for samples and details of the \DAGANUN~and ~\ref{SI:DAGANRN_details} for samples and details of the \DAGANRN).
\end{itemize}
All these models are generative models conditioned by an image prototype extracted from the training or the test set. The way we have selected the prototypes is detailed in Eq.~\ref{eq:eq1}. To the best of our knowledge, these models offer a representative set of one-shot generative models. We have reproduced all these models (sometimes with our own implementation when it was not available online). Our code could be found at \url{https://github.com/serre-lab/diversity_vs_recognizability}. 

\section{The diversity vs. accuracy framework}

Let $\{x_i^{j}\}$ be a dataset composed of K concepts (i.e., classes) with N samples each ($i\in[\![1,N ]\!]$ and $j\in[\![1,K]\!]$). The framework we propose aims at evaluating the performance of a generative model $p_{\theta}$, parameterized by $\theta$, that produces new images $v_i^j$ based on a single sample (or prototype) of a concept given to the generator $\tilde{x}^{j}$ (i.e., $v_{i}^{j} \sim p_{\theta}(\cdot|\tilde{x}^{j})$). For each concept $j$, we define a prototype as the sample closest to the center of mass for the concept $j$:
\begin{align}
\tilde{x}^j = x_{i^{*}}^{j} \quad \text{s.t.} \quad i^{*} = \argmin_{i} \norm{ f(x_{i}^{j}) - \frac{1}{N}\sum_{i=1}^{N} f(x_{i}^{j})}_2
\label{eq:eq1}
\end{align}
In Eq.~\ref{eq:eq1}, $f$ denotes a function that projects the input image from the pixel space to a feature space. We will detail the feature extractor $f$ shortly. Note that this definition of a prototype is not unique (one could also select the prototype randomly within individual classes $j$), nevertheless this selection mechanism is a guarantee that the selected sample will be representative of the concept.

\textbf{Dataset.}~In this article, we use the Omniglot dataset~\citep{lake2015human} with a weak generalization split~\citep{rezende2016one}. Omniglot is composed of binary images representing $1,623$ classes of handwritten letters and symbols (extracted from $50$ different alphabets) with just $20$ samples per class. We have downsampled the original dataset to be $50\times50$ pixels. The weak generalization split consists of a training set composed of all available symbols minus $3$ symbols per alphabet which are left aside for the test set. It is said to be \emph{weak} because all the alphabets were shown during the training process (albeit not all symbols in those alphabets). As the Omniglot dataset is hand-written by humans, we consider that these samples reflect a human generative process and we refer to this later as the \human~model.

\textbf{Diversity.}~In the proposed framework, \emph{diversity} refers to the intra-class variability of the samples produced by a generative model $p_{\theta}(\cdot|\tilde{x}^{j})$. For a given prototype $\tilde{x}^{j}$, we compute the diversity as the standard deviation of the generated samples in the feature space $f$:
\begin{align}
\sigma_{p_\theta}^{j} = \sqrt{\frac{1}{N-1}\sum_{i=1}^{N}\Big(f(v_{i}^{j}) -  \frac{1}{N}\sum_{i=1}^{N}f(v_{i}^{j})\Big)^2}\quad \text{s.t.} \quad v_{i}^{j} \sim p_{\theta}(\cdot|\tilde{x}^{j})
\label{eq:eq2}
\end{align}
 
We use the Bessel-corrected standard deviation to keep a good estimate of the data dispersion despite the relatively small number of samples (e.g., $N=20$ for the Omniglot dataset used here). To verify that this diversity measure is robust to the specific choice of the feature extractor $f$, we explored two different settings: features learned with class supervision by a Prototypical Net~\citep{snell2017prototypical} and features learned with self-supervision by a SimCLR network~\citep{chen2020simple}. In both cases, we extracted the features from the first fully-connected layer following the last convolutional layer. The Prototypical Net was optimized so that images that belong to the same category share similar latent representations as measured by the $\ell_2$-norm. Similarly, SimCLR leverages a contrastive loss to define a latent representation such that a sample is more similar to its augmented version than to other image samples. In SimCLR, this similarity is computed with cosine similarity. These two approaches represent two ends of a continuum of methods to learn suitable representational spaces without the need to explicitly learn to classify images and are thus more suitable for few-shot learning tasks~\citep{li2021deep}. For the sake of comparison, we have used the exact same network architecture for both feature extractors (see sections~\ref{SI:ProtoNet} and~\ref{SI:SimCLR} for more details on Prototypical Net and SimCLR, respectively). 
 
We computed the samples diversity for all $150$ categories of the Omniglot test set (i.e., $v_{i}^{j} = x_{i}^{j}$ in this experiment) using both the supervised and unsupervised settings. We found a high linear correlation ($\rho=0.86$, p-value~$<10^{-5}$) and a high rank-order Spearman correlation ($\rho=0.85$, p-value~$<10^{-5}$) between the two settings (see  section~\ref{SI:Diversity_supervised_vs_non_supervised}). Hence, the two feature extraction methods produce comparable diversity measures and henceforth, we will report results using the unsupervised setting.
 
As an additional control, we have also verified that the SimCLR metric is robust to changes to the augmentation method used (see section~\ref{SI:SimCLRAugmentation effect}) and to the specific choice of the dispersion metric (see section~\ref{SI:More_dispersion_measure} for more details on this comparison). We have compared the feature space of the Prototypical Net and SimCLR using a t-SNE analysis (see section~\ref{SI:SimCLR_latent_analysis}). We observed a strong clustering of samples belonging to the same category for both networks. It suggests that the augmentation methods used by the SimCLR contrastive loss are sufficient to disentangle the class information.

\begin{figure}[h!]
\begin{tikzpicture}
\draw [anchor=north west] (0\linewidth, 0.97\linewidth) node {\includegraphics[width=0.045\linewidth]{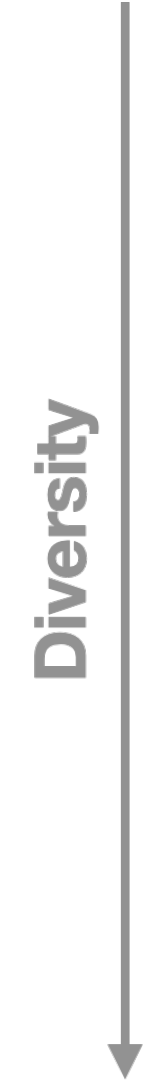}};
\draw [anchor=north west] (0.05\linewidth, 0.97\linewidth) node {\includegraphics[width=0.465\linewidth]{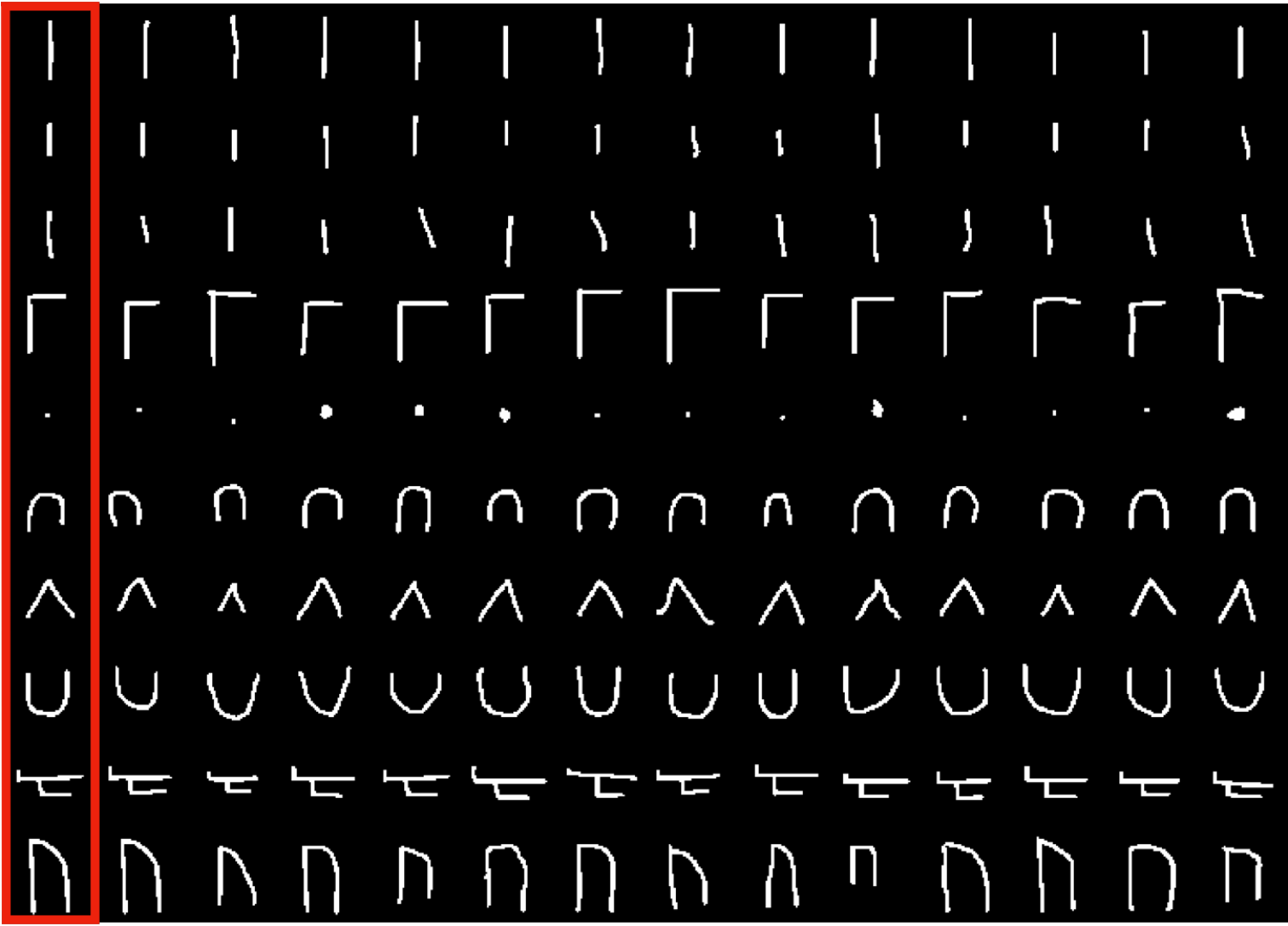}};
\draw [anchor=north west] (0.525\linewidth, 0.97\linewidth) node {\includegraphics[width=0.465\linewidth]{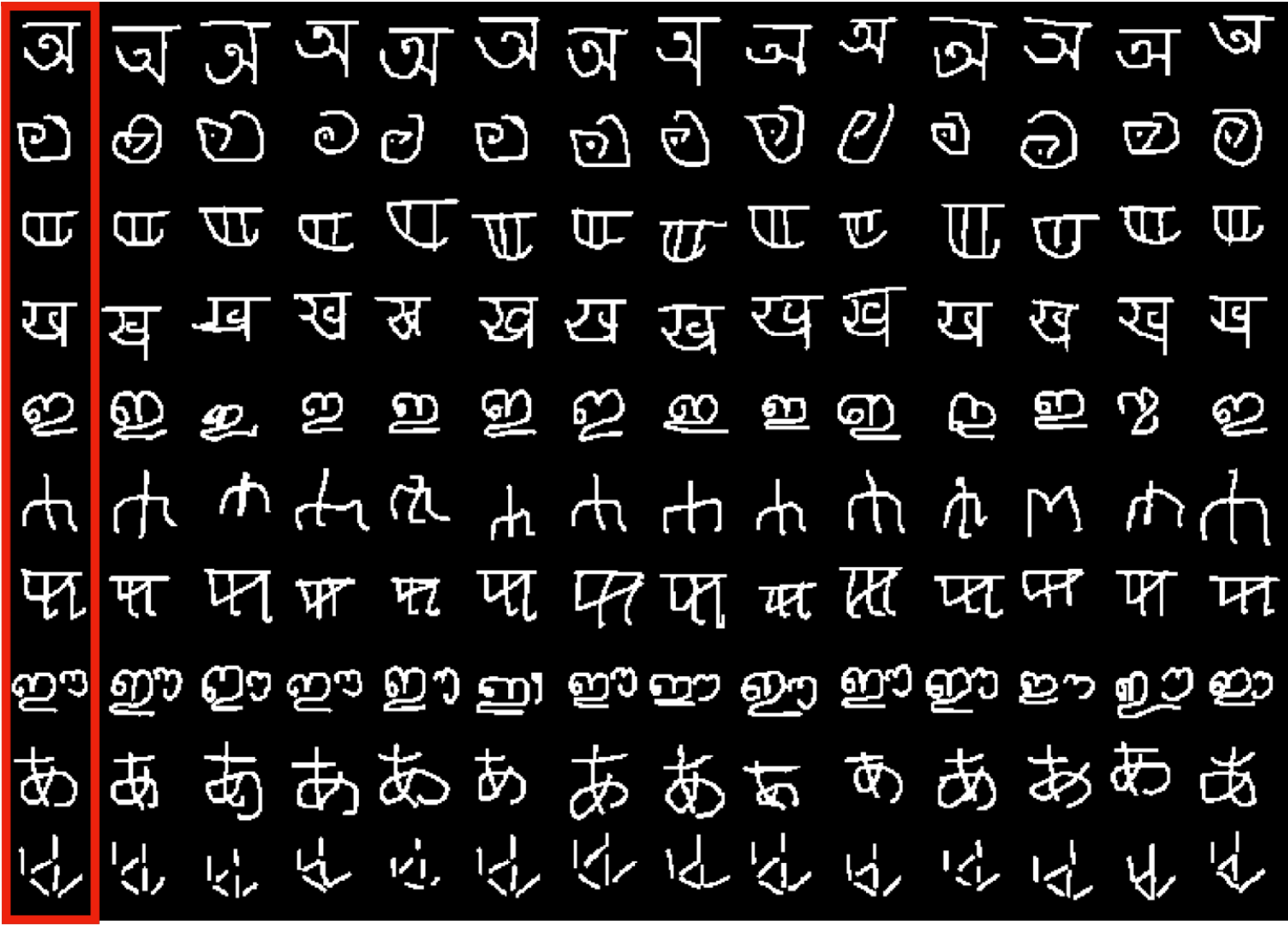}};
\begin{scope}
    \draw [anchor=north west,fill=white, align=left] (0.05\linewidth, 1\linewidth) node {\bf a) };
    \draw [anchor=north west,fill=white, align=left] (0.525\linewidth, 1\linewidth) node {\bf b)};
\end{scope}

\end{tikzpicture}
     \caption{Samples from the top 10 Omniglot concepts (test set) associated with the lowest (\textbf{a}) vs. highest diversity (\textbf{b}). The different concepts are ranked vertically from less diverse to more diverse. Prototypes for individual concepts are shown within a red box next to actual class samples.}
        \label{fig:fig2}
\end{figure}

Fig.~\ref{fig:fig2} shows the $10$ concepts from the Omniglot test set with the lowest and highest samples diversity, respectively (for more diversity-ranked concepts with unsupervised or supervised setting,  see sections~\ref{SI:ranked_concepts_unsupervised} and ~\ref{SI:ranked_concepts_supervised}, respectively). One can see that the proposed diversity metric is qualitatively similar to human judgment. Concepts with low diversity are composed of very few relatively basic strokes (e.g., lines, dots, etc) with little room for any kind of  ``creativity'' in the generation process while more diverse concepts are composed of more numerous and more complex stroke combinations with many more opportunities for creativity.

\textbf{Recognizability.}~We evaluate the recognizability of the samples produced by the one-shot generative models by leveraging one-shot classification models. As demonstrated in~\ref{SI:Overfitting}, it is also possible to use a few-shot classifier to evaluate the recognizability. We prefer one-shot classifier to match the settings proposed in~\citep{lake2015human}. In order to make sure our classification accuracy measure is robust to the choice of the model, we test different models which belong to the two main approaches used in machine learning for one-shot classification: metric learning and meta-learning~\citep{li2021deep}. We selected the Prototypical Net~\citep{snell2017prototypical} as a representative metric-learning approach and the Model-Agnostic Meta-Learning (MAML)~\citep{finn2017model} model as a representative meta-learning approach. Both models were trained and tested in a 1-shot 20-ways setting (see section~\ref{SI:MAML_details} for more details on the MAML architecture and training details). We report a high Pearson (negative) correlation between the logits produced by Prototypical Net and MAML ($\rho=-0.60$, p-value~$<10^{-5}$) as well as a strong Spearman rank-order correlation between the classification accuracy of both networks ($\rho=0.62$, p-value~$<10^{-5}$). See section~\ref{SI:ctrl_exp_accu} for more details about this control experiment. Hence, our recognizability metric is robust to the choice of the one-shot classification model (even when those models are leveraging different approaches) and henceforth, we will report results using the Prototypical Net model. 

\section{Results}
\subsection{GAN-like vs. VAE-like models}
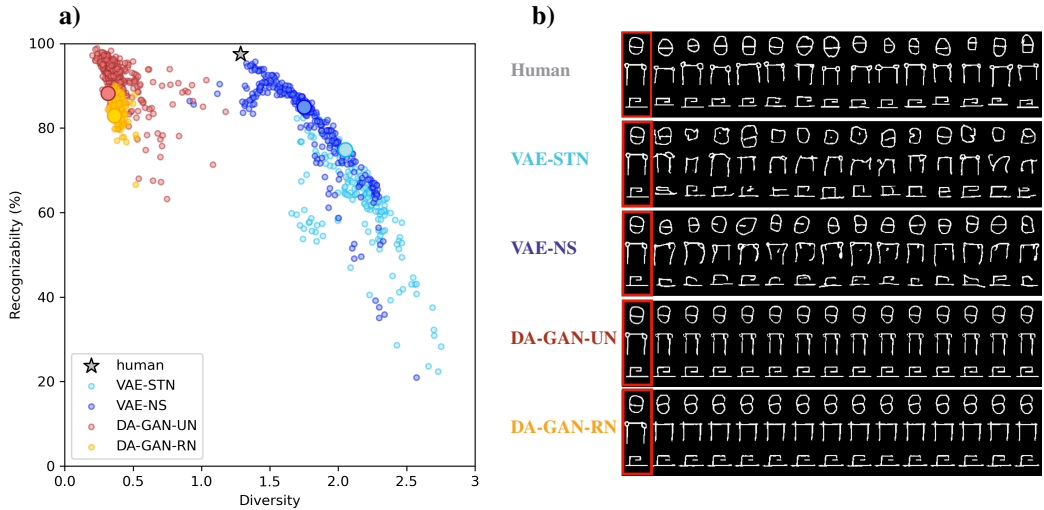
\begin{figure}[h!]
\begin{tikzpicture}
\draw [anchor=north west] (0\linewidth, 0.97\linewidth) node {\includegraphics[width=0.45\linewidth]{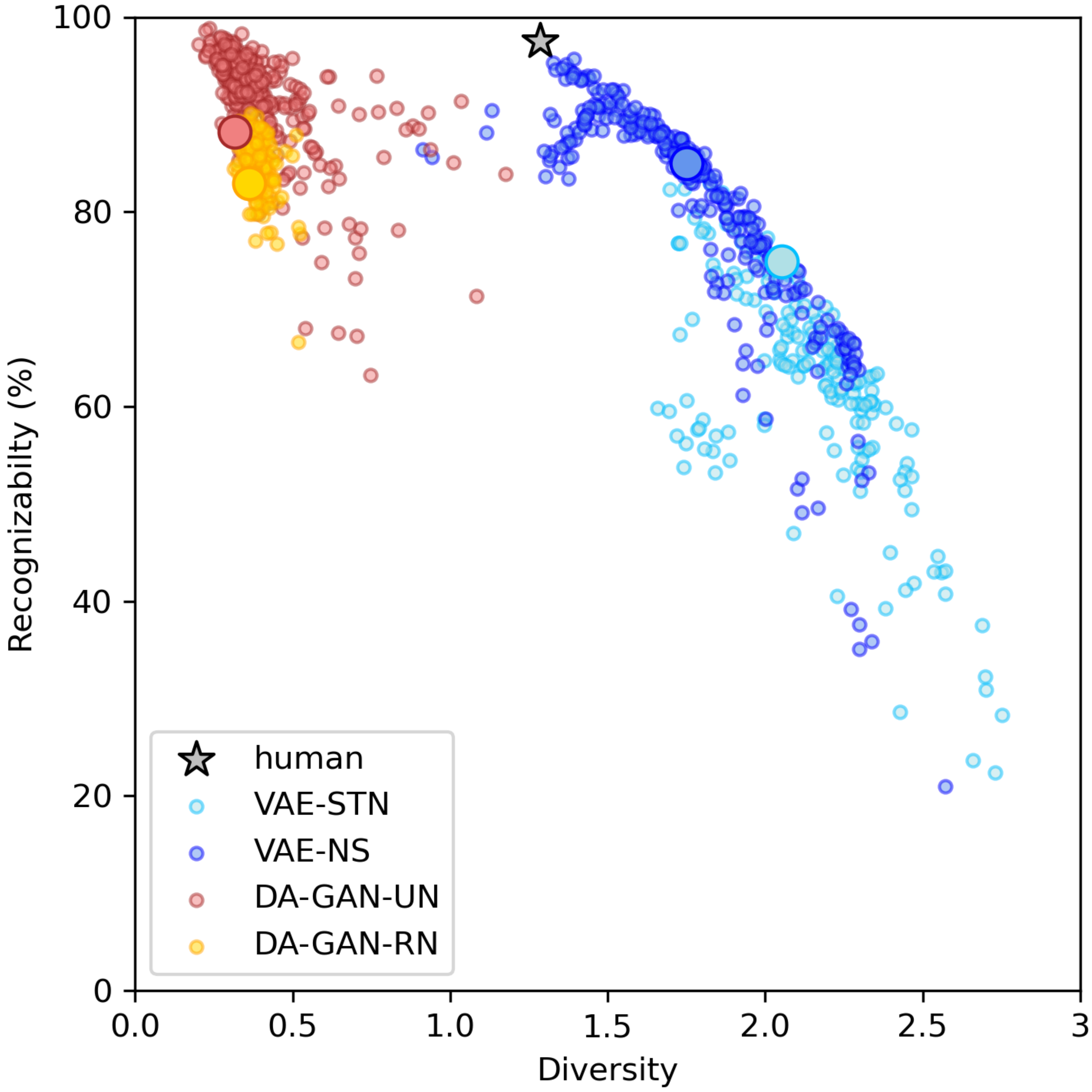}};

\draw [anchor=north west] (0.585\linewidth, 0.975\linewidth) node {\includegraphics[width=0.4\linewidth]{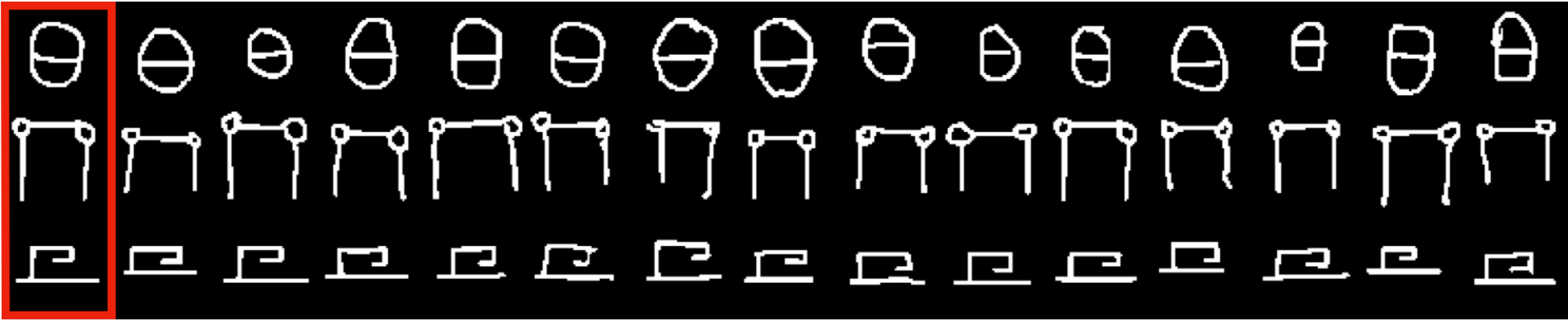}};

\draw [anchor=north west] (0.585\linewidth, 0.89\linewidth) node {\includegraphics[width=0.4\linewidth]{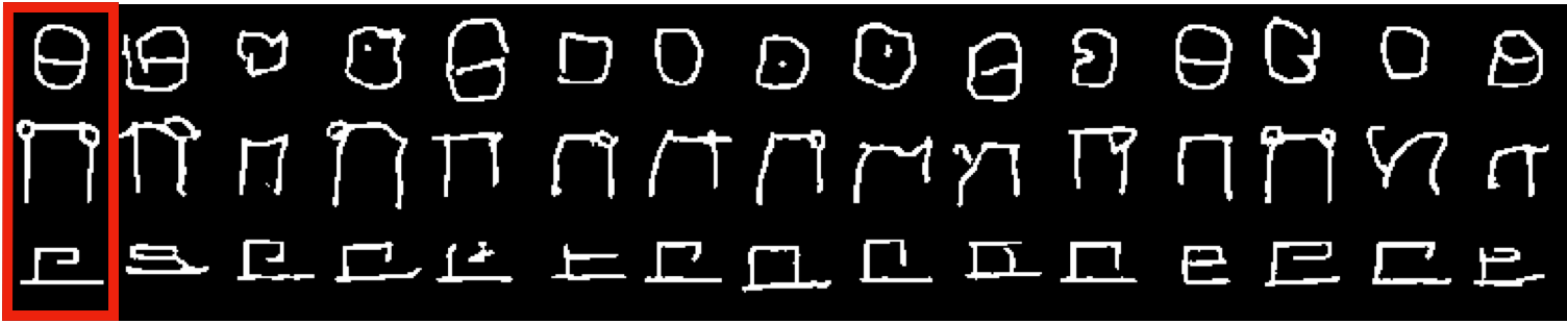}};

\draw [anchor=north west] (0.585\linewidth, 0.805\linewidth) node {\includegraphics[width=0.4\linewidth]{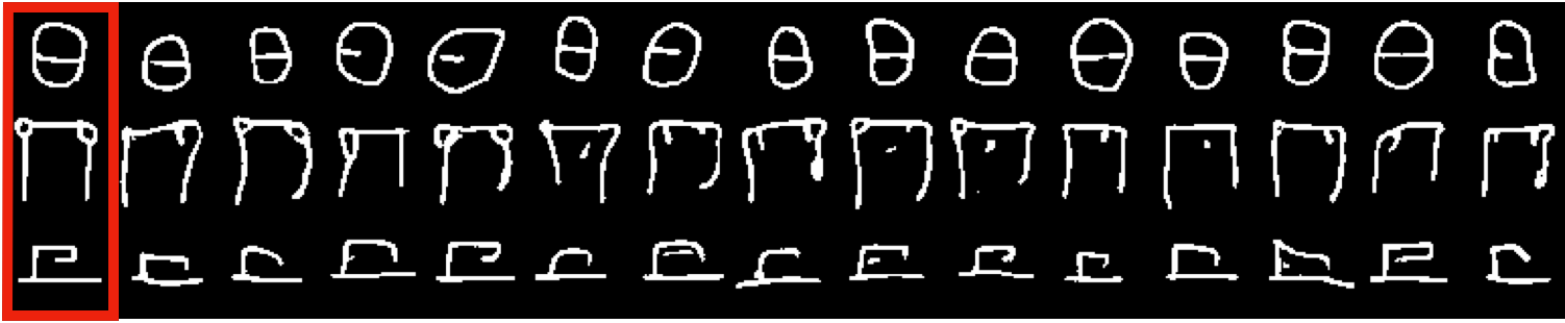}};

\draw [anchor=north west] (0.585\linewidth, 0.72\linewidth) node {\includegraphics[width=0.4\linewidth]{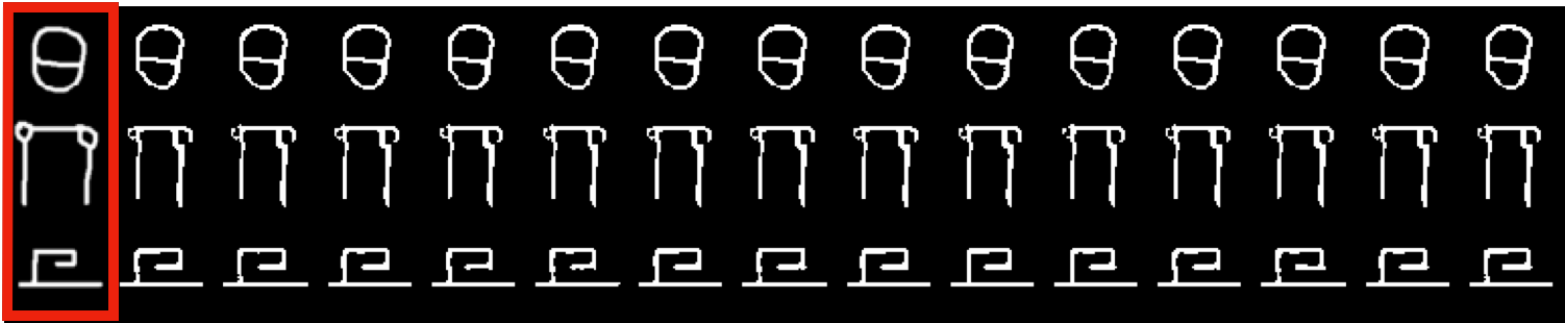}};

\draw [anchor=north west] (0.585\linewidth, 0.635\linewidth) node {\includegraphics[width=0.4\linewidth]{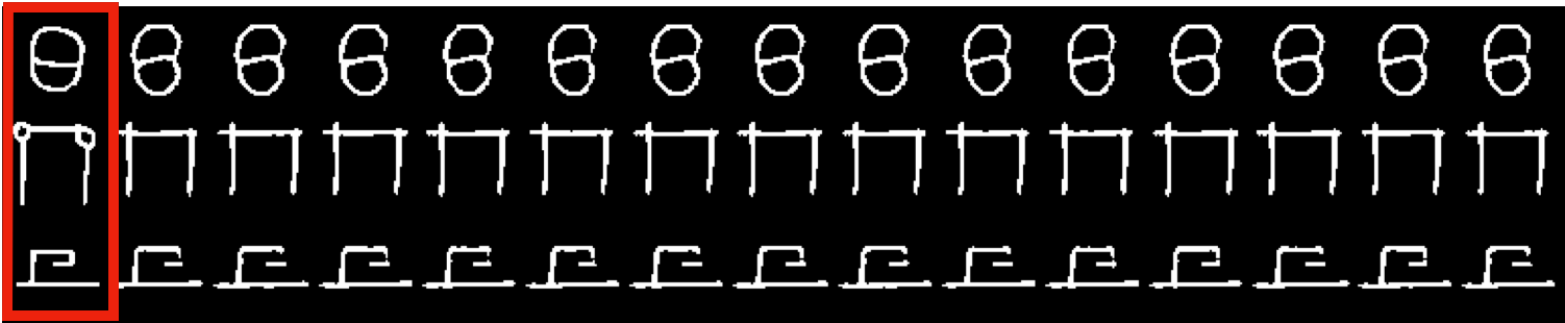}};

\begin{scope}
    \draw [anchor=north west,fill=white, align=left] (0.05\linewidth, 1\linewidth) node {\bf a) };
    
    \draw [anchor=north west,fill=white, align=left] (0.50\linewidth, 1\linewidth) node {\bf b)};
    
    \draw [anchor=north west,fill=white, align=left] (0.48\linewidth, 0.945\linewidth) node {\scriptsize \bf \textcolor{Gray}{Human}} ;
    
    \draw [anchor=north west,fill=white, align=left] (0.48\linewidth, 0.86\linewidth) node {\scriptsize \bf \textcolor{SkyBlue}{VAE-STN}};
    
    \draw [anchor=north west,fill=white, align=left] (0.48\linewidth, 0.775\linewidth) node {\scriptsize \bf \textcolor{BlueViolet}{VAE-NS}};
    
    \draw [anchor=north west,fill=white, align=left] (0.48\linewidth, 0.69\linewidth) node {\scriptsize  \bf \textcolor{Mahogany}{DA-GAN-UN}};
    
    \draw [anchor=north west,fill=white, align=left] (0.48\linewidth, 0.605\linewidth) node {\scriptsize  \bf \textcolor{YellowOrange}{DA-GAN-RN}};

\end{scope}

\end{tikzpicture}
     \caption{(\textbf{a}) Diversity vs. recognizability plot for all tested models (colored data points) and human (black star). Each data point corresponds to the mean diversity and recognizability over all classes of the Omniglot test set. Bigger circles correspond to the base architecture of each model that all have a comparable number of parameters ($\approx6-7$M). The \human~data-point is computed based on the testing samples of the Omniglot dataset. (\textbf{b}) Samples produced by the different models in their base architectures (corresponding to the bigger circles in Fig.~\ref{fig:fig3}a). Prototypes for individual concepts are shown within a red box next to actual class samples.}
        \label{fig:fig3}
\end{figure}

For all algorithms listed in section~\ref{MAIN:models} we have explored different hyper-parameters (see section~\ref{MAIN:hyper_parameters_effect} for more details), leading to various models represented in the diversity vs. recognizability plot in Fig.~\ref{fig:fig3}a. In this figure, we have reduced each model to a single point by averaging the diversity and recognizability over all classes of the Omniglot testing set. The black star corresponds to the \human~model, and colored data points are computed based on the samples generated by the \VAENS, \VAESTN, \DAGANUN~and the \DAGANRN. The base architectures for all algorithms (highlighted with bigger points in Fig.~\ref{fig:fig3}a) have a comparable number of parameters ($\approx$6-7 M, see \ref{SI:VAE_STN_details}, \ref{SI:NS_details} and \ref{SI:DAGAN_details} for more details on the base architectures). 

We observe that the GAN-like models (i.e., \DAGANUN~and \DAGANRN) tend to be located at the upper left side of the graph while VAE-like models (i.e., \VAENS~and \VAESTN) spread on the right side of the graph. Therefore, the GAN-like models produce very recognizable samples that are highly similar to each other (high recognizability and low diversity). In contrast, VAE-like models generate more diverse but less recognizable samples. The samples in Fig.~\ref{fig:fig3}b illustrate this observation. The difference between GAN and VAE-like samples could be explained by their loss functions~\citep{lucas2019adaptive}. The GANs' adversarial loss tends to drop some of the modes of the training distribution. In general, the distribution learned by GANs put excessive mass on the more likely modes but discards secondary modes~\citep{arjovsky2017wasserstein}. This phenomenon leads to sharp and recognizable generations at the cost of reduced samples diversity. On the other hand, VAEs (and likelihood-based models in general) are suffering from over-generalization: they cover all the modes of the training distribution and put mass in spurious regions~\citep{bishop2006pattern}. We refer the reader to Fig.~4 of~\citet{lucas2019adaptive} for an illustration of mode dropping in GANs and over-generalization in VAEs. Our diversity vs. recognizability plot in Fig.~\ref{fig:fig3}a shows that this phenomenon is holding even when the testing distribution is different from the training distribution as in the case of the one-shot generation scenario.

\subsection{VAE-NS vs. VAE-STN}
\label{MAIN:hyper_parameters_effect}
\begin{figure}[h!]
\begin{tikzpicture}
\draw [anchor=north west] (0\linewidth, 0.97\linewidth) node {\includegraphics[width=0.45\linewidth]{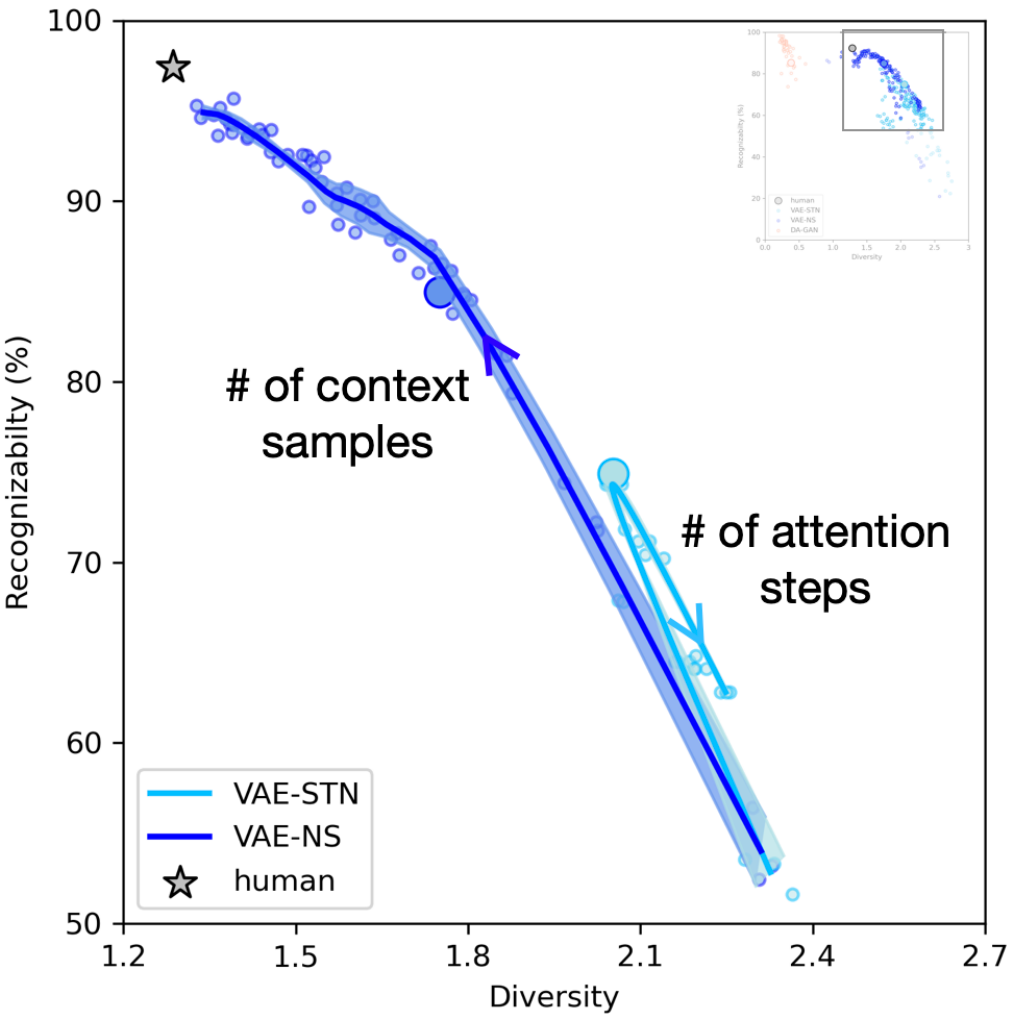}};
\draw [anchor=north west] (0.5\linewidth, 0.97\linewidth) node {\includegraphics[width=0.45\linewidth]{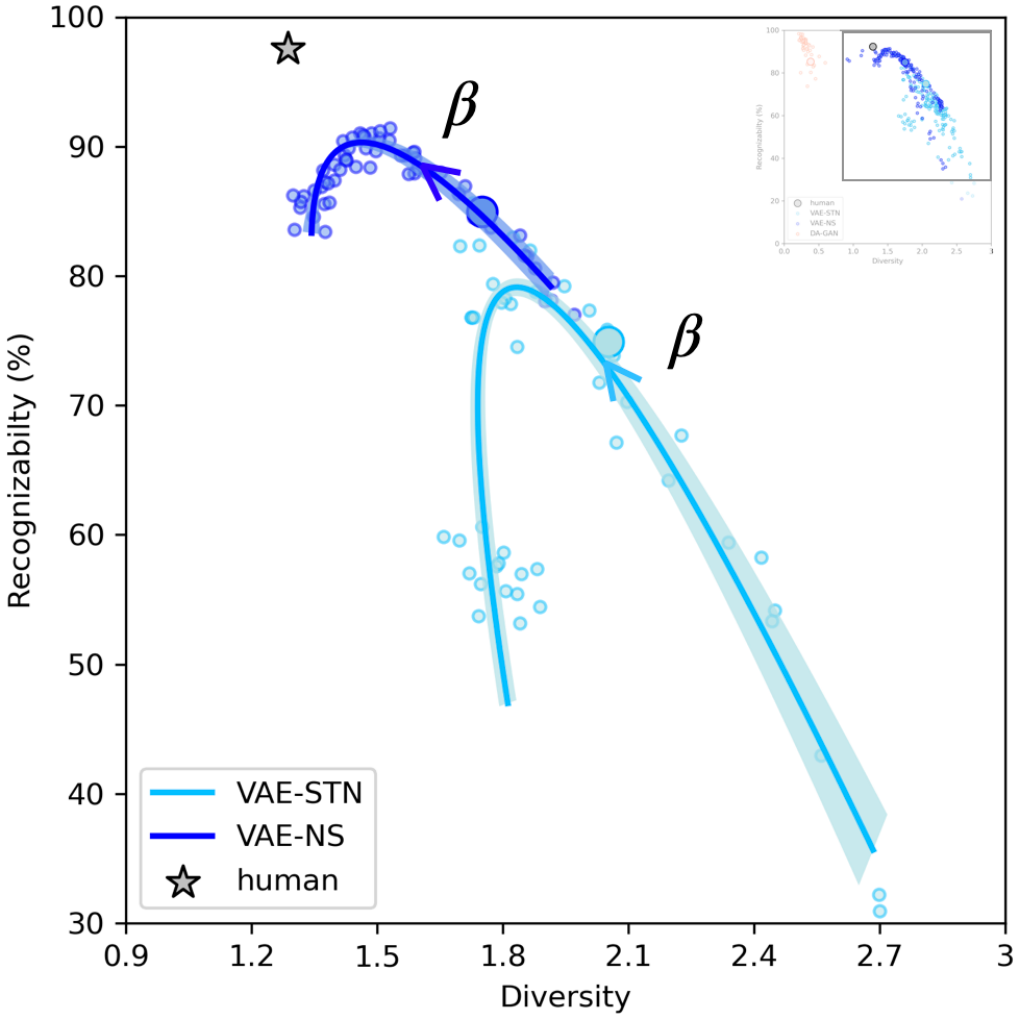}};
\begin{scope}
    \draw [anchor=north west,fill=white, align=left] (0.05\linewidth, 1\linewidth) node {\bf a) };
    
    \draw [anchor=north west,fill=white, align=left] (0.50\linewidth, 1\linewidth) node {\bf b)};
\end{scope}

\end{tikzpicture}
     \caption{(\textbf{a}) Effect of the number of context samples of the \VAENS~and attentional steps of the \VAESTN. Each data point represents a model with a different number of context samples for the \VAENS~(ranging from $2$ to $20$) or a different number of attentional steps for the \VAESTN~(ranging from $20$ to $90$). The base architectures, highlighted with a bigger circle, correspond to $10$ context samples and $60$ attentional steps for the \VAENS~and~\VAESTN~respectively. (\textbf{b}) Effect of $\beta$. The base architectures correspond to a $\beta=1$. In all curves, solid lines represent the mean of the parametric curves over $3$ different runs. Shaded areas are computing using the standard deviation over $3$ different runs. Arrows show the direction in which the tested variables (context samples, attention steps or $\beta$) are increased.}
        \label{fig:fig4}
\end{figure}

In this section we compare some key hyper-parameters of the \VAENS~and \VAESTN. The core idea of the \VAENS~is to integrate context information to the sample generation process. During training, the context is composed of several samples that all represent variations of the same visual concept. Those samples are passed in a separate encoder to extract a context statistics (denoted $c$ in ~\ref{SI:NS_details}) used to condition the generative process. During the testing phase, the \VAENS~infers the context statistics using a single image (i.e., the prototype). We evaluate the effect of the context on the position of the \VAENS~models on the diversity-recognizability space by varying the number of samples used to compute the context statistics during the training phase (from $2$ to $20 $ samples). Importantly, varying the number of context samples do not change the number of parameters of the network. For all tested runs, we observe a monotonic decrease of the samples diversity (see Fig.~\ref{SI:fig_effect_context}a) and a monotonic increase of the samples recognizability (see Fig.~\ref{SI:fig_effect_context}b) when the number of context samples is increased. In the diversity-recognizability space, the resulting curve is monotonically transporting models from the lower-right side to the upper-left side of the plot (see Fig.~\ref{fig:fig4}a, dark blue curve). The effect of the number of context samples is large: the diversity is almost divided by $2$ (from $2.4$ to $1.2$) and the classification accuracy is increased by $80\%$ (from $53\%$ to $96\%$). This result suggests that increasing the number of context samples for a given visual concept helps the generative model to identify the properties and the features that are crucial for good recognition, but hurts the diversity of the generated samples.

In contrast to the \VAENS, the \VAESTN~uses spatial attention to sequentially attend to sub-parts of the image to decompose it into simpler elements. These sub-parts are then easier to encode and synthesize. In the \VAESTN, one can vary the number of attentional steps (i.e., the number of attended locations) to modulate spatial attention. Importantly, varying the number of attentional steps does not change the number of parameters. We have varied the number of attentional steps from $20$ to $90$. The relationship between the number of attentional steps, the samples diversity, and the samples recognizability is non-monotonic. We have used a parametric curve fitting method (i.e., the least curve fitting method from~\citep{grossman1971parametric}) to parameterize the curve while maintaining the order of the data point (see ~\ref{SI:effect_attention} for more details on the fitting procedure). We report a convex parabolic relationship between the number of attentional steps and the samples diversity (see Fig.~\ref{SI:fig_effect_attention}a). This curve is minimal at $60$ steps. We observe a concave parabolic relationship between the number of attentional steps and the recognizability of samples. This curve is maximal at $60$ attentional steps (see Fig.~\ref{SI:fig_effect_attention}b). In Fig.~\ref{fig:fig4}b we have plotted the parametric fit illustrating the position of the \VAESTN~models in the diversity-recognizability space when one increases the number of attentional steps (the light blue curve). This curve follows a quasi-linear trend with a sharp turn-around (at $60$ attentional steps). The effect of the number of attentional steps on the diversity-recognizability is limited compared to the effect of the number of context samples.

Both \VAENS~and \VAESTN~are trained to maximize the Evidence Lower Bound (ELBO), it is then possible to tune the weight of the prior in the loss function. One can operate such a modulation by changing the $\beta$ coefficient in the ELBO loss function~\citep{higgins2016beta}. We refer the reader to \ref{SI:effect_beta} for more mathematical details about the ELBO. A high $\beta$ value enforces the latent variable to be closer to a normal distribution and increases the information bottleneck in the latent space. Increasing $\beta$ is known to force the disentanglement of the generative factors~\citep{burgess2018understanding}. We observe a monotonic decreasing relationship between the value of $\beta$ and the samples diversity for both the \VAESTN~and the \VAENS~(see Fig.~\ref{SI:fig_effect_beta_vae_stn}a and Fig.~\ref{SI:fig_effect_beta_ns}a, respectively). We report a concave parabolic relationship between $\beta$ and the samples recognizability. We use the least curve fitting method to find the optimal parabolic curves~\citep{grossman1971parametric}. This curve is maximal at $\beta=2.5$ for the \VAESTN~and at $\beta=3$ for the \VAENS~(see Fig.~\ref{SI:fig_effect_beta_vae_stn}b and Fig.~\ref{SI:fig_effect_beta_ns}b, respectively). The overall effect of $\beta$ on the position of the VAE-like models on the diversity-recognizability space is relatively similar for both the \VAESTN~and the \VAENS~and follows a clear parabolic trend. These curves demonstrate that one could modulate the value of $\beta$ to maximize the recognizability. In general, we observe that the variable controlling the context-size in the \VAENS~is the one having the biggest impact on the diversity-recognizability space.

We have also varied the architecture of the  \DAGANUN, \DAGANRN, \VAENS, and \VAESTN~by changing the size of the latent space. We did not find any common trend between the size of the latent variable, the diversity, and the recognizability (see~\ref{SI:effect_z} for more details). We observe that the \DAGANUN~tends to produce slightly more recognizable but less diverse samples than the \DAGANRN~while both architectures have the same number of parameters. It suggests that the extra skip connections included in the U-Net architecture, in between the encoder and the decoder of the \DAGANUN, allow to trade diversity for recognizability.

\subsection{Comparison with humans}
\begin{figure}[h!]
\begin{tikzpicture}
\draw [anchor=north west] (0\linewidth, 0.97\linewidth) node {\includegraphics[width=0.47\linewidth]{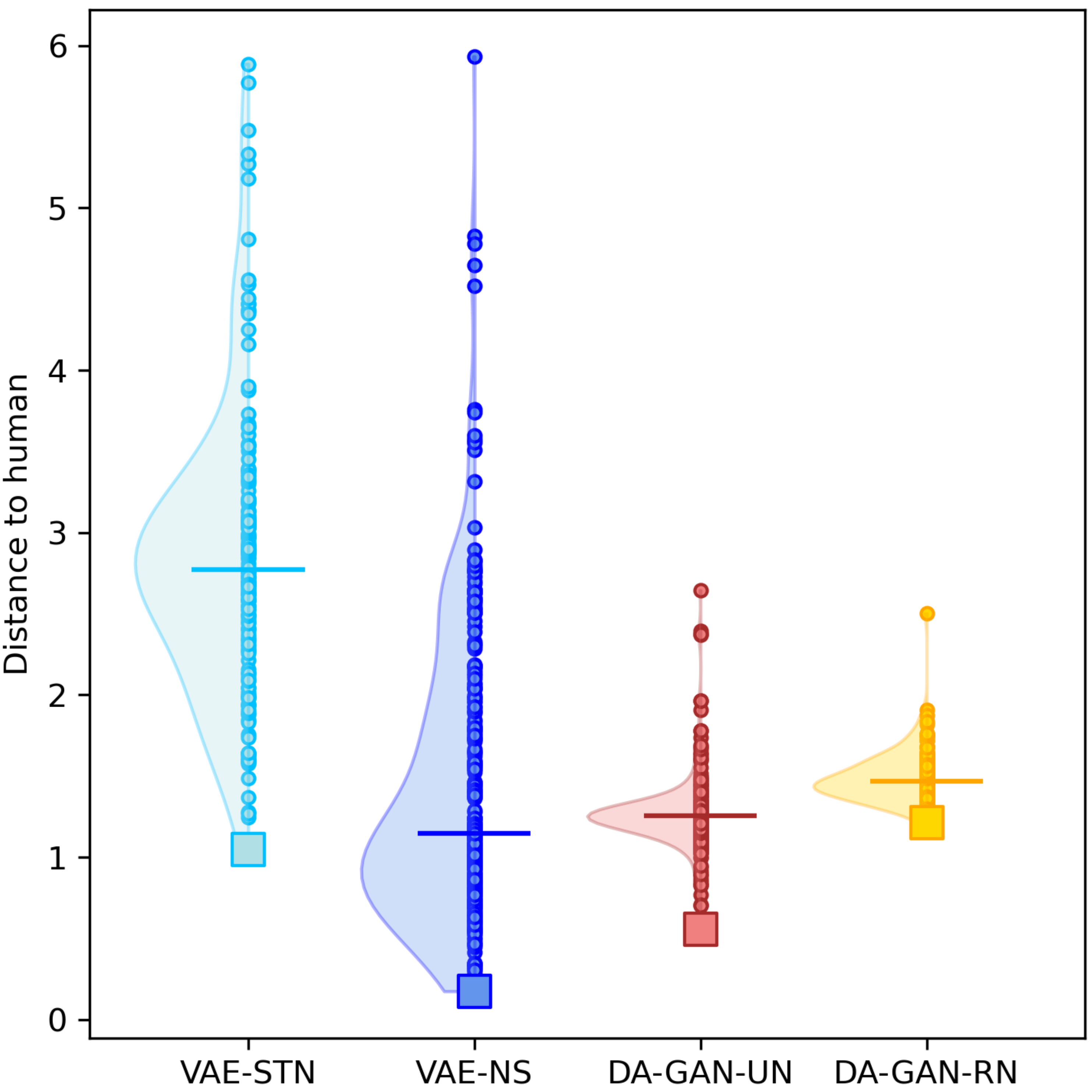}};
\draw [anchor=north west] (0.47\linewidth, 1\linewidth) node {\includegraphics[width=0.52\linewidth]{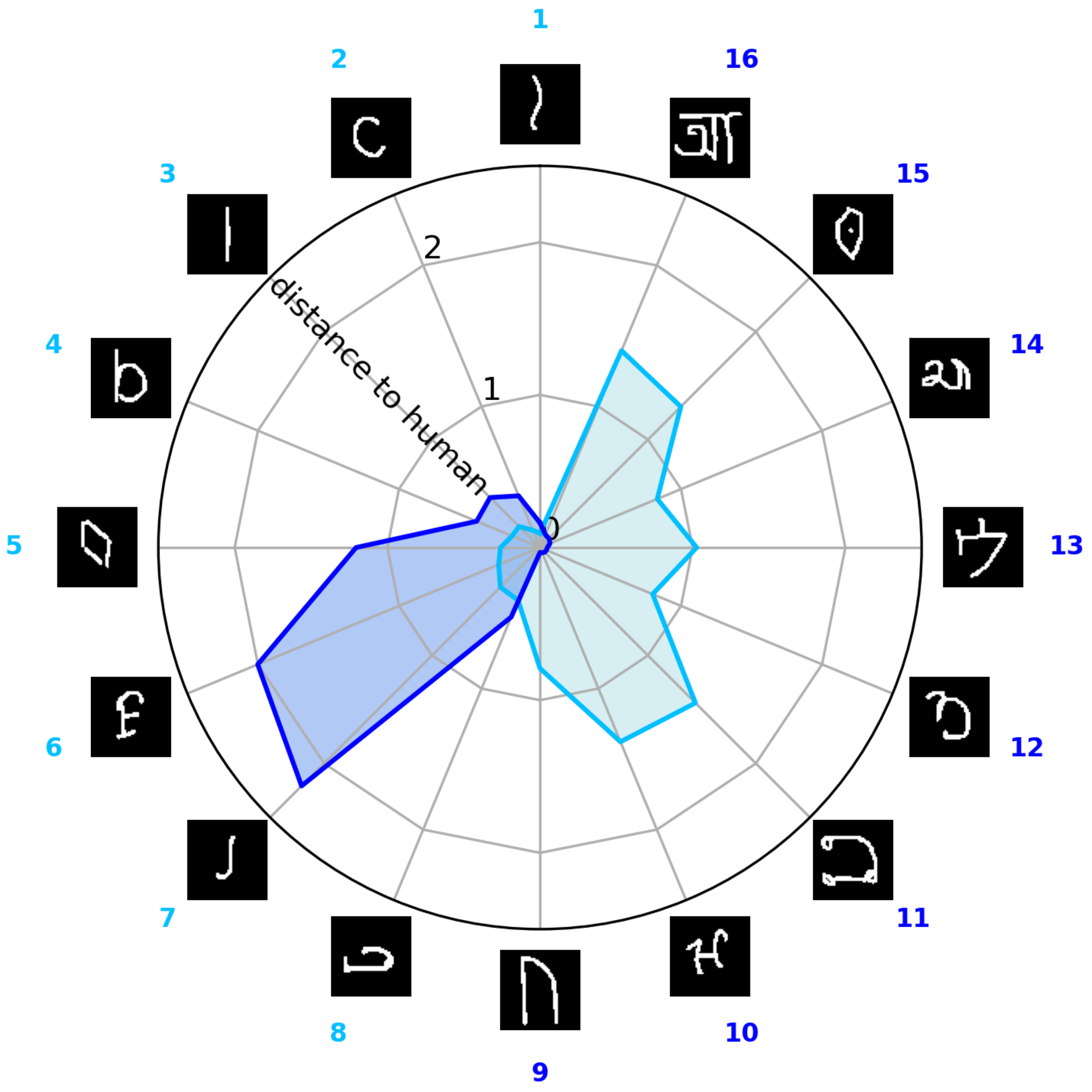}};
\draw [anchor=north west] (0.89\linewidth, 0.53\linewidth) node {\includegraphics[width=0.1\linewidth]{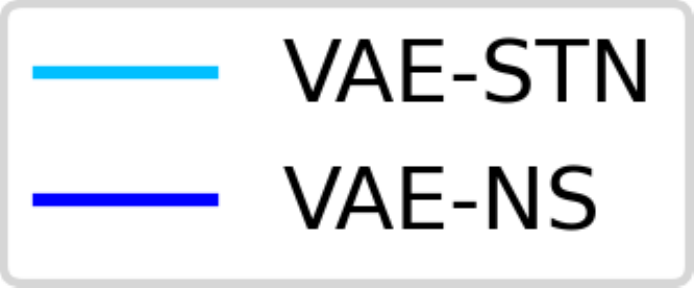}};

\begin{scope}
    \draw [anchor=north west,fill=white, align=left] (0.05\linewidth, 1\linewidth) node {\bf a) };
    
    \draw [anchor=north west,fill=white, align=left] (0.50\linewidth, 1\linewidth) node {\bf b)};
\end{scope}

\end{tikzpicture}
     \caption{(\textbf{a}) Distribution of average distance to humans for the \VAENS, \VAESTN, \DAGANUN~ and \DAGANRN. Each data point corresponds to the mean diversity and accuracy over all classes of the Omniglot test set. Squares correspond to the model showing the smallest distance to humans. The distance to humans is calculated with a $\ell_2$-norm on the diversity-recognizability space after z-score normalization. The horizontal line denotes the median of the model's distribution. (\textbf{b}) Distance to humans on $16$ different visual concepts for the \VAENS~and \VAESTN~models that best approximate human data (i.e., indicated by a square in Fig.~\ref{fig:fig5}a). The visual concepts $1$ to $8$ are selected to minimize the \VAESTN~distance to humans, and visual concepts $9$ to $16$ minimize the \VAENS~distance to humans. Images surrounding the radar plot are the prototypes of the visual concepts. }
    \label{fig:fig5}
\end{figure}

We now compare the tested models with the \human~data in the diversity-recognizability space. To perform such a comparison, we first normalize all the model's diversity and recognizability (including humans) using the z-score such that both axes are scaled and centered similarly. Then, for all models, we compute the $\ell_2$-distance between models and humans in the diversity-recognizability space. We remind that the \human~data point is computed using the samples of the Omniglot test set. Distances to humans as well as their distributions are reported for all models in Fig.~\ref{fig:fig5}a. The median of \VAENS~models is closer to humans, followed by \DAGANUN, \DAGANRN~and \VAESTN~(medians are indicated in Fig.~\ref{fig:fig5}a with horizontal bars). The \VAENS~model showing the smallest distance is almost at the human level (see dark blue square Fig.~\ref{fig:fig5}a). It has a context size of $20$ samples (the highest possible context size), and a $\beta=2.5$. The~\VAESTN~model that best approximates human has a $\beta=2.25$ and $60$ attentional steps (see light blue square in Fig.~\ref{fig:fig5}a).

So far, we have reduced all models to single points by averaging the diversity and recognizability values over all classes. We now study distances to humans for individual classes and for the \VAENS~and the \VAESTN~models showing the shortest distance to humans (indicated by blue squares in Fig.~\ref{fig:fig5}a). In Fig.~\ref{fig:fig5}b, we report distances to human for these $2$ models and for $16$ visual concepts. The visual concepts $1$ to $8$ and $9$ to $16$ are selected so that they minimize the distance to humans with the \VAESTN~and the \VAENS, respectively. We observe that these visual concepts are different for the \VAENS~and the \VAESTN~model. Therefore, both models are well approximating human data for some visual concepts but not for others. Interestingly, we qualitatively observe that the visual concepts $1$ to $8$ look simpler (i.e., made with fewer strokes) than the visual concepts $9$ to $16$. It suggests that the spatial attention mechanism used by \VAESTN~provides a better human approximation for simple visual concepts, while the context integration method leveraged by the \VAENS~is more relevant to mimic human data on more complex visual concepts.

\section{Discussion}
In this article, we have described a novel framework for comparing computational models with human participants on the one-shot generation task. The framework measures the diversity and the recognizability of the produced samples using metrics compatible with the one-shot scenario. To the best of our knowledge, this is the first and only framework specifically tailored to evaluate and compare one-shot image generation models. 

Among all tested algorithms, the \VAENS~is the best human approximator on Omniglot (see Fig.~\ref{fig:fig5}a). It suggests that the context integration mechanism of the \VAENS~is an important component to reach human-like generalization. Interestingly, motor learning experiments have demonstrated that human generalization performances are also strongly related to contextual information~\citep{taylor2013context}. Interestingly,~\citep{tenenbaum1998bayesian} have demonstrated that a bayesian observer tends to overestimate the intra-class variance when only a few context samples are accessible. Our results are in-line with this finding: Fig.~\ref{fig:fig4}a shows an high diversity when the number of context samples is low while the diversity is decreasing when more context samples are available. It suggests that the \VAENS~is acting as a bayesian observer: it overestimates intra-class variance when the context is scarse.

In addition, we demonstrate that one can tune $\beta$ so that the model becomes closer to human data (see Fig.~\ref{fig:fig4}b). This is consistent with a prior computational neuroscience study that has shown that disentangled VAEs (with $\beta>1$) provide a good model of face-tuned neurons in the inferotemporal cortex~\citep{higgins2021unsupervised}. Our comparison between the \VAENS~and the \VAESTN~suggests that a model which uses a spatial attention mechanism better fits human data for simple visual concepts. In contrast, the context integration mechanism of the \VAENS~appears to be a better human approximator for more complex visual concepts. One could thus try to combine both mechanisms towards  improving the similarity with human data independent of the complexity of the visual concept. We have also found that GAN-like models (\DAGANRN~and \DAGANUN) better account for human recognizability but do not approximate well the diversity of the human samples. In contrast, VAE-like models (\VAENS~and \VAESTN) better account for human diversity. An interesting approach would be to leverage a hybrid architecture (such as  the VAE-GAN~\cite{larsen2016autoencoding}) to try to better match human data. 

Other candidate ingredients include the ability to harness compositionality~\citep{lake2015human} or the recurrent processes thought to be crucial for human generalization~\citep{wyatte2012limits}. Compositionality could be introduced in one-shot generative algorithms by quantizing the latent space (as in the VQVAE~\citep{van2017neural}). As a result, each coordinate of the latent variable represents an address in a codebook, and the role of the prior is then to combine simpler concepts to generate more complex samples. One promising way to include recurrent processing into generative models is through the predictive coding framework~\citep{rao1999predictive}. Predictive Coding suggests that each processing step is part of an inference scheme that minimizes the prediction error~\citep{boutin2020iterative}. Previous work has demonstrated that such networks are more robust and exhibit improved generalization abilities~\citep{boutin2020effect, choksi2021predify}. All these ingredients could be tested and compared against human abilities using the diversity/recognizability framework we have proposed in this paper.

In the current version of the Omniglot dataset, the intra-class variability does not reflect the human level of creativity. It is mainly due to the experimental protocol in which one asks human participants to copy a given visual concept. The Omniglot dataset could be enriched with more diverse samples, by explicitly asking human participants to be as creative as possible. Other drawing databases with more complex symbols such as \emph{Quick Draw!}~\citep{jongejan2016quick} could also be considered to strengthen the comparison with humans.

By decomposing the performance of the one-shot generation task along the recognizability vs. diversity axes we wanted to shed light on the relationship between generalization and creativity (quantified by the samples diversity in our framework). We hope one can make use of our framework to validate key hypotheses about human generalization abilities so that we can better understand the brain. We argue that the best way to reach human-like generalization abilities is to unleash the algorithms' creativity.

\section*{Acknowledgement}
This work was funded by ANITI (Artificial and Natural Intelligence Toulouse Institute) and the French National Research Agency,  under the grant agreement number : ANR-19-PI3A-0004. Additional funding to TS was provided by ONR (N00014-19-1-2029) and NSF (IIS-1912280 and EAR-1925481).  Computing hardware supported by NIH Office of the Director grant S10OD025181 via the Center for Computation and Visualization (CCV). We thanks Roland W. Fleming and his team for the insightful feedback and discussion about the diversity vs. recognizability framework.

\newpage
\bibliography{biblio}

\begin{thebibliography}{54}
\providecommand{\natexlab}[1]{#1}
\providecommand{\EM}{\em}
\providecommand{\RNtxt}{\relax}
\RNtxt{}

\bibitem[Antoniou et~al.(2017)A.~Antoniou, A.~Storkey,
  H.~Edwards]{antoniou2017data}
{\EM Antoniou Antreas, Storkey Amos, Edwards Harrison}.
\newblock Data augmentation generative adversarial networks
  \allowbreak\newblock// arXiv preprint arXiv:1711.04340. 2017.

\bibitem[Arjovsky et~al.(2017)M.~Arjovsky, S.~Chintala,
  L.~Bottou]{arjovsky2017wasserstein}
{\EM Arjovsky Martin, Chintala Soumith, Bottou L{\'e}on}.
\newblock Wasserstein generative adversarial networks \allowbreak\newblock//
  International conference on machine learning. 2017.  214--223.

\bibitem[Barratt, Sharma(2018)S.~Barratt, R.~Sharma]{barratt2018note}
{\EM Barratt Shane, Sharma Rishi}.
\newblock A note on the inception score \allowbreak\newblock// arXiv preprint
  arXiv:1801.01973. 2018.

\bibitem[Bishop, Nasrabadi(2006)C.~M. Bishop, N.~M.
  Nasrabadi]{bishop2006pattern}
{\EM Bishop Christopher~M, Nasrabadi Nasser~M}.
\newblock Pattern recognition and machine learning.  4, 4. 2006.

\bibitem[Boutin et~al.(2020{\natexlab{a}})V.~Boutin, A.~Franciosini,
  F.~Ruffier, L.~Perrinet]{boutin2020effect}
{\EM Boutin Victor, Franciosini Angelo, Ruffier Franck, Perrinet Laurent}.
\newblock Effect of top-down connections in Hierarchical Sparse Coding
  \allowbreak\newblock// Neural Computation. 2020{\natexlab{a}}. 32, 11.
  2279--2309.

\bibitem[Boutin et~al.(2020{\natexlab{b}})V.~Boutin, A.~Zerroug, M.~Jung,
  T.~Serre]{boutin2020iterative}
{\EM Boutin Victor, Zerroug Aimen, Jung Minju, Serre Thomas}.
\newblock Iterative VAE as a predictive brain model for out-of-distribution
  generalization \allowbreak\newblock// arXiv preprint arXiv:2012.00557.
  2020{\natexlab{b}}.

\bibitem[Brigato, Iocchi(2021)L.~Brigato, L.~Iocchi]{brigato2021close}
{\EM Brigato Lorenzo, Iocchi Luca}.
\newblock A close look at deep learning with small data \allowbreak\newblock//
  2020 25th International Conference on Pattern Recognition (ICPR). 2021.
  2490--2497.

\bibitem[Broedelet et~al.(2022)I.~Broedelet, P.~Boersma, J.~Rispens,
  et~al.]{broedelet2022school}
{\EM Broedelet Iris, Boersma Paul, Rispens Judith, others }.
\newblock School-Aged Children Learn Novel Categories on the Basis of
  Distributional Information \allowbreak\newblock// Frontiers in Psychology.
  2022. 12, 799241.

\bibitem[Burgess et~al.(2018)C.~P. Burgess, I.~Higgins, A.~Pal, L.~Matthey,
  N.~Watters, G.~Desjardins, A.~Lerchner]{burgess2018understanding}
{\EM Burgess Christopher~P, Higgins Irina, Pal Arka, Matthey Loic, Watters
  Nick, Desjardins Guillaume, Lerchner Alexander}.
\newblock Understanding disentangling in beta-VAE \allowbreak\newblock// arXiv
  preprint arXiv:1804.03599. 2018.

\bibitem[Chen et~al.(2020)T.~Chen, S.~Kornblith, M.~Norouzi,
  G.~Hinton]{chen2020simple}
{\EM Chen Ting, Kornblith Simon, Norouzi Mohammad, Hinton Geoffrey}.
\newblock A simple framework for contrastive learning of visual representations
  \allowbreak\newblock// International conference on machine learning. 2020.
  1597--1607.

\bibitem[Choksi et~al.(2021)B.~Choksi, M.~Mozafari, C.~Biggs~O'May, B.~Ador,
  A.~Alamia, R.~VanRullen]{choksi2021predify}
{\EM Choksi Bhavin, Mozafari Milad, Biggs~O'May Callum, Ador Benjamin, Alamia
  Andrea, VanRullen Rufin}.
\newblock Predify: Augmenting deep neural networks with brain-inspired
  predictive coding dynamics \allowbreak\newblock// Advances in Neural
  Information Processing Systems. 2021. 34.

\bibitem[Chomsky(1965)N.~Chomsky]{chomsky1965aspects}
{\EM Chomsky Noam}.
\newblock Aspects of the theory of syntax Special technical report no. 11.
  1965.

\bibitem[Chong, Forsyth(2020)M.~J. Chong, D.~Forsyth]{chong2020effectively}
{\EM Chong Min~Jin, Forsyth David}.
\newblock Effectively unbiased fid and inception score and where to find them
  \allowbreak\newblock// Proceedings of the IEEE/CVF conference on computer
  vision and pattern recognition. 2020.  6070--6079.

\bibitem[Chowdhury et~al.(2022)A.~Chowdhury, D.~Chaudhari, S.~Chaudhuri,
  C.~Jermaine]{chowdhury2022meta}
{\EM Chowdhury Arkabandhu, Chaudhari Dipak, Chaudhuri Swarat, Jermaine Chris}.
\newblock Meta-Meta Classification for One-Shot Learning \allowbreak\newblock//
  Proceedings of the IEEE/CVF Winter Conference on Applications of Computer
  Vision. 2022.  177--186.

\bibitem[Edwards, Storkey(2016)H.~Edwards, A.~Storkey]{edwards2016towards}
{\EM Edwards Harrison, Storkey Amos}.
\newblock Towards a neural statistician \allowbreak\newblock// arXiv preprint
  arXiv:1606.02185. 2016.

\bibitem[Feinman, Lake(2020)R.~Feinman, B.~M. Lake]{feinman2020generating}
{\EM Feinman Reuben, Lake Brenden~M}.
\newblock Generating new concepts with hybrid neuro-symbolic models
  \allowbreak\newblock// arXiv preprint arXiv:2003.08978. 2020.

\bibitem[Feldman(1997)J.~Feldman]{feldman1997structure}
{\EM Feldman Jacob}.
\newblock The structure of perceptual categories \allowbreak\newblock// Journal
  of mathematical psychology. 1997. 41, 2. 145--170.

\bibitem[Finn et~al.(2017)C.~Finn, P.~Abbeel, S.~Levine]{finn2017model}
{\EM Finn Chelsea, Abbeel Pieter, Levine Sergey}.
\newblock Model-agnostic meta-learning for fast adaptation of deep networks
  \allowbreak\newblock// International conference on machine learning. 2017.
  1126--1135.

\bibitem[Giannone, Winther(2021)G.~Giannone,
  O.~Winther]{giannone2021hierarchical}
{\EM Giannone Giorgio, Winther Ole}.
\newblock Hierarchical Few-Shot Generative Models \allowbreak\newblock// Fifth
  Workshop on Meta-Learning at the Conference on Neural Information Processing
  Systems. 2021.

\bibitem[Goodfellow et~al.(2014)I.~Goodfellow, J.~Pouget-Abadie, M.~Mirza,
  B.~Xu, D.~Warde-Farley, S.~Ozair, A.~Courville,
  Y.~Bengio]{goodfellow2014generative}
{\EM Goodfellow Ian, Pouget-Abadie Jean, Mirza Mehdi, Xu~Bing, Warde-Farley
  David, Ozair Sherjil, Courville Aaron, Bengio Yoshua}.
\newblock Generative adversarial nets \allowbreak\newblock// Advances in neural
  information processing systems. 2014. 27.

\bibitem[Grossman(1971)M.~Grossman]{grossman1971parametric}
{\EM Grossman M}.
\newblock Parametric curve fitting \allowbreak\newblock// The Computer Journal.
  1971. 14, 2. 169--172.

\bibitem[Heusel et~al.(2017)M.~Heusel, H.~Ramsauer, T.~Unterthiner, B.~Nessler,
  S.~Hochreiter]{heusel2017gans}
{\EM Heusel Martin, Ramsauer Hubert, Unterthiner Thomas, Nessler Bernhard,
  Hochreiter Sepp}.
\newblock Gans trained by a two time-scale update rule converge to a local nash
  equilibrium \allowbreak\newblock// Advances in neural information processing
  systems. 2017. 30.

\bibitem[Higgins et~al.(2021)I.~Higgins, L.~Chang, V.~Langston, D.~Hassabis,
  C.~Summerfield, D.~Tsao, M.~Botvinick]{higgins2021unsupervised}
{\EM Higgins Irina, Chang Le, Langston Victoria, Hassabis Demis, Summerfield
  Christopher, Tsao Doris, Botvinick Matthew}.
\newblock Unsupervised deep learning identifies semantic disentanglement in
  single inferotemporal face patch neurons \allowbreak\newblock// Nature
  communications. 2021. 12, 1. 1--14.

\bibitem[Higgins et~al.(2016)I.~Higgins, L.~Matthey, A.~Pal, C.~Burgess,
  X.~Glorot, M.~Botvinick, S.~Mohamed, A.~Lerchner]{higgins2016beta}
{\EM Higgins Irina, Matthey Loic, Pal Arka, Burgess Christopher, Glorot Xavier,
  Botvinick Matthew, Mohamed Shakir, Lerchner Alexander}.
\newblock beta-vae: Learning basic visual concepts with a constrained
  variational framework \allowbreak\newblock// arXiv preprint arXiv:1804.03599.
  2016.

\bibitem[Hinton et~al.(2012)G.~Hinton, N.~Srivastava,
  K.~Swersky]{hinton2012neural}
{\EM Hinton Geoffrey, Srivastava Nitish, Swersky Kevin}.
\newblock Neural networks for machine learning lecture 6a overview of
  mini-batch gradient descent \allowbreak\newblock// Cited on. 2012. 14, 8. 2.

\bibitem[Jaderberg et~al.(2015)M.~Jaderberg, K.~Simonyan, A.~Zisserman,
  et~al.]{jaderberg2015spatial}
{\EM Jaderberg Max, Simonyan Karen, Zisserman Andrew, others }.
\newblock Spatial transformer networks \allowbreak\newblock// Advances in
  neural information processing systems. 2015. 28.

\bibitem[Jongejan et~al.(2016)J.~Jongejan, H.~Rowley, T.~Kawashima, J.~Kim,
  N.~Fox-Gieg]{jongejan2016quick}
{\EM Jongejan Jonas, Rowley Henry, Kawashima Takashi, Kim Jongmin, Fox-Gieg
  Nick}.
\newblock The quick, draw!-ai experiment \allowbreak\newblock// Mount View, CA,
  accessed Feb. 2016. 17, 2018. 4.

\bibitem[Kingma, Ba(2014)D.~P. Kingma, J.~Ba]{kingma2014adam}
{\EM Kingma Diederik~P, Ba~Jimmy}.
\newblock Adam: A method for stochastic optimization \allowbreak\newblock//
  arXiv preprint arXiv:1412.6980. 2014.

\bibitem[Kingma, Welling(2013)D.~P. Kingma, M.~Welling]{kingma2013auto}
{\EM Kingma Diederik~P, Welling Max}.
\newblock Auto-encoding variational bayes \allowbreak\newblock// arXiv preprint
  arXiv:1312.6114. 2013.

\bibitem[Kingma, Dhariwal(2018)D.~P. Kingma, P.~Dhariwal]{kingma2018glow}
{\EM Kingma Durk~P, Dhariwal Prafulla}.
\newblock Glow: Generative flow with invertible 1x1 convolutions
  \allowbreak\newblock// Advances in neural information processing systems.
  2018. 31.

\bibitem[Koch et~al.(2015)G.~Koch, R.~Zemel, R.~Salakhutdinov,
  et~al.]{koch2015siamese}
{\EM Koch Gregory, Zemel Richard, Salakhutdinov Ruslan, others }.
\newblock Siamese neural networks for one-shot image recognition
  \allowbreak\newblock// ICML deep learning workshop.  2. 2015. ~0.

\bibitem[Lake et~al.(2015)B.~M. Lake, R.~Salakhutdinov, J.~B.
  Tenenbaum]{lake2015human}
{\EM Lake Brenden~M, Salakhutdinov Ruslan, Tenenbaum Joshua~B}.
\newblock Human-level concept learning through probabilistic program induction
  \allowbreak\newblock// Science. 2015. 350, 6266. 1332--1338.

\bibitem[Lake et~al.(2017)B.~M. Lake, T.~D. Ullman, J.~B. Tenenbaum, S.~J.
  Gershman]{lake2017building}
{\EM Lake Brenden~M, Ullman Tomer~D, Tenenbaum Joshua~B, Gershman Samuel~J}.
\newblock Building machines that learn and think like people
  \allowbreak\newblock// Behavioral and brain sciences. 2017. 40.

\bibitem[Larsen et~al.(2016)A.~B.~L. Larsen, S.~K. S{\o}nderby, H.~Larochelle,
  O.~Winther]{larsen2016autoencoding}
{\EM Larsen Anders Boesen~Lindbo, S{\o}nderby S{\o}ren~Kaae, Larochelle Hugo,
  Winther Ole}.
\newblock Autoencoding beyond pixels using a learned similarity metric
  \allowbreak\newblock// International conference on machine learning. 2016.
  1558--1566.

\bibitem[Li et~al.(2021)X.~Li, X.~Yang, Z.~Ma, J.-H. Xue]{li2021deep}
{\EM Li~Xiaoxu, Yang Xiaochen, Ma~Zhanyu, Xue Jing-Hao}.
\newblock Deep metric learning for few-shot image classification: A selective
  review \allowbreak\newblock// arXiv preprint arXiv:2105.08149. 2021.

\bibitem[Lucas et~al.(2019)T.~Lucas, K.~Shmelkov, K.~Alahari, C.~Schmid,
  J.~Verbeek]{lucas2019adaptive}
{\EM Lucas Thomas, Shmelkov Konstantin, Alahari Karteek, Schmid Cordelia,
  Verbeek Jakob}.
\newblock Adaptive density estimation for generative models
  \allowbreak\newblock// Advances in Neural Information Processing Systems.
  2019. 32.

\bibitem[Mishra et~al.(2017)N.~Mishra, M.~Rohaninejad, X.~Chen,
  P.~Abbeel]{mishra2017simple}
{\EM Mishra Nikhil, Rohaninejad Mostafa, Chen Xi, Abbeel Pieter}.
\newblock A simple neural attentive meta-learner \allowbreak\newblock// arXiv
  preprint arXiv:1707.03141. 2017.

\bibitem[Nalisnick et~al.(2018)E.~Nalisnick, A.~Matsukawa, Y.~W. Teh, D.~Gorur,
  B.~Lakshminarayanan]{nalisnick2018deep}
{\EM Nalisnick Eric, Matsukawa Akihiro, Teh Yee~Whye, Gorur Dilan,
  Lakshminarayanan Balaji}.
\newblock Do deep generative models know what they don't know?
  \allowbreak\newblock// arXiv preprint arXiv:1810.09136. 2018.

\bibitem[Oord Van~den et~al.(2016)A.~Van~den Oord, N.~Kalchbrenner,
  L.~Espeholt, O.~Vinyals, A.~Graves, et~al.]{van2016conditional}
{\EM Oord Aaron Van~den, Kalchbrenner Nal, Espeholt Lasse, Vinyals Oriol,
  Graves Alex, others }.
\newblock Conditional image generation with pixelcnn decoders
  \allowbreak\newblock// Advances in neural information processing systems.
  2016. 29.

\bibitem[Piattelli-Palmarini(1980)M.~Piattelli-Palmarini]{piattelli1980language}
{\EM Piattelli-Palmarini Massimo}.
\newblock Language and learning: the debate between Jean Piaget and Noam
  Chomsky \allowbreak\newblock// Harvard Univ Press, Cambridge, MA. 1980.

\bibitem[Rao, Ballard(1999)R.~P. Rao, D.~H. Ballard]{rao1999predictive}
{\EM Rao Rajesh~PN, Ballard Dana~H}.
\newblock Predictive coding in the visual cortex: a functional interpretation
  of some extra-classical receptive-field effects \allowbreak\newblock// Nature
  neuroscience. 1999. 2, 1. 79--87.

\bibitem[Rezende et~al.(2016)D.~Rezende, I.~Danihelka, K.~Gregor, D.~Wierstra,
  et~al.]{rezende2016one}
{\EM Rezende Danilo, Danihelka Ivo, Gregor Karol, Wierstra Daan, others }.
\newblock One-shot generalization in deep generative models
  \allowbreak\newblock// International conference on machine learning. 2016.
  1521--1529.

\bibitem[Richards et~al.(1992)W.~Richards, J.~Feldman,
  A.~Jepson]{richards1992features}
{\EM Richards Whitman, Feldman Jacob, Jepson A}.
\newblock From features to perceptual categories \allowbreak\newblock// BMVC92.
  1992.  99--108.

\bibitem[Salakhutdinov et~al.(2012)R.~Salakhutdinov, J.~Tenenbaum,
  A.~Torralba]{salakhutdinov2012one}
{\EM Salakhutdinov Ruslan, Tenenbaum Joshua, Torralba Antonio}.
\newblock One-shot learning with a hierarchical nonparametric bayesian model
  \allowbreak\newblock// Proceedings of ICML Workshop on Unsupervised and
  Transfer Learning. 2012.  195--206.

\bibitem[Salimans et~al.(2016)T.~Salimans, I.~Goodfellow, W.~Zaremba,
  V.~Cheung, A.~Radford, X.~Chen]{salimans2016improved}
{\EM Salimans Tim, Goodfellow Ian, Zaremba Wojciech, Cheung Vicki, Radford
  Alec, Chen Xi}.
\newblock Improved techniques for training gans \allowbreak\newblock// Advances
  in neural information processing systems. 2016. 29.

\bibitem[Santoro et~al.(2016)A.~Santoro, S.~Bartunov, M.~Botvinick,
  D.~Wierstra, T.~Lillicrap]{santoro2016meta}
{\EM Santoro Adam, Bartunov Sergey, Botvinick Matthew, Wierstra Daan, Lillicrap
  Timothy}.
\newblock Meta-learning with memory-augmented neural networks
  \allowbreak\newblock// International conference on machine learning. 2016.
  1842--1850.

\bibitem[Snell et~al.(2017)J.~Snell, K.~Swersky,
  R.~Zemel]{snell2017prototypical}
{\EM Snell Jake, Swersky Kevin, Zemel Richard}.
\newblock Prototypical networks for few-shot learning \allowbreak\newblock//
  Advances in neural information processing systems. 2017. 30.

\bibitem[Sung et~al.(2018)F.~Sung, Y.~Yang, L.~Zhang, T.~Xiang, P.~H. Torr,
  T.~M. Hospedales]{sung2018learning}
{\EM Sung Flood, Yang Yongxin, Zhang Li, Xiang Tao, Torr Philip~HS, Hospedales
  Timothy~M}.
\newblock Learning to compare: Relation network for few-shot learning
  \allowbreak\newblock// Proceedings of the IEEE conference on computer vision
  and pattern recognition. 2018.  1199--1208.

\bibitem[Szegedy et~al.(2016)C.~Szegedy, V.~Vanhoucke, S.~Ioffe, J.~Shlens,
  Z.~Wojna]{szegedy2016rethinking}
{\EM Szegedy Christian, Vanhoucke Vincent, Ioffe Sergey, Shlens Jon, Wojna
  Zbigniew}.
\newblock Rethinking the inception architecture for computer vision
  \allowbreak\newblock// Proceedings of the IEEE conference on computer vision
  and pattern recognition. 2016.  2818--2826.

\bibitem[Taylor, Ivry(2013)J.~A. Taylor, R.~B. Ivry]{taylor2013context}
{\EM Taylor Jordan~A, Ivry Richard~B}.
\newblock Context-dependent generalization \allowbreak\newblock// Frontiers in
  Human Neuroscience. 2013. 7. 171.

\bibitem[Tenenbaum(1998)J.~Tenenbaum]{tenenbaum1998bayesian}
{\EM Tenenbaum Joshua}.
\newblock Bayesian modeling of human concept learning \allowbreak\newblock//
  Advances in neural information processing systems. 1998. 11.

\bibitem[Tiedemann et~al.(2021)H.~Tiedemann, Y.~Morgenstern, F.~Schmidt, R.~W.
  Fleming]{tiedemann2021one}
{\EM Tiedemann Henning, Morgenstern Yaniv, Schmidt Filipp, Fleming Roland~W}.
\newblock One shot generalization in humans revealed through a drawing task
  \allowbreak\newblock// bioRxiv. 2021.

\bibitem[Van Den~Oord et~al.(2017)A.~Van Den~Oord, O.~Vinyals,
  et~al.]{van2017neural}
{\EM Van Den~Oord Aaron, Vinyals Oriol, others }.
\newblock Neural discrete representation learning \allowbreak\newblock//
  Advances in neural information processing systems. 2017. 30.

\bibitem[Wyatte et~al.(2012)D.~Wyatte, T.~Curran,
  R.~O'Reilly]{wyatte2012limits}
{\EM Wyatte Dean, Curran Tim, O'Reilly Randall}.
\newblock The limits of feedforward vision: Recurrent processing promotes
  robust object recognition when objects are degraded \allowbreak\newblock//
  Journal of Cognitive Neuroscience. 2012. 24, 11. 2248--2261.

\end{thebibliography}

\if \checklist1
	\newpage
	\section*{Checklist}


	\begin{enumerate}

	\item For all authors...
	\begin{enumerate}
	  \item Do the main claims made in the abstract and introduction accurately reflect the paper's contributions and scope?
	    \answerYes{}
	  \item Did you describe the limitations of your work?
	    \answerYes{Described in the conclusion.}
	  \item Did you discuss any potential negative societal impacts of your work?
	    \answerYes{We have briefly discussed societal impact in~\ref{SI:computational_resources}.}
	  \item Have you read the ethics review guidelines and ensured that your paper conforms to them?
	    \answerYes{}
	\end{enumerate}

	\item If you are including theoretical results...
	\begin{enumerate}
	  \item Did you state the full set of assumptions of all theoretical results?
	    \answerNA{No theoretical results}
		\item Did you include complete proofs of all theoretical results?
	    \answerNA{No theoretical results}
	\end{enumerate}

	\item If you ran experiments...
	\begin{enumerate}
	  \item Did you include the code, data, and instructions needed to reproduce the main experimental results (either in the supplemental material or as a URL)?
	    \answerYes{All the codes are available on github ( link given in section 2.2)}
	  \item Did you specify all the training details (e.g., data splits, hyper-parameters, how they were chosen)?
	    \answerYes{All the training details, as well as the architectural details are given in the supplementary materials.}
		\item Did you report error bars (e.g., with respect to the random seed after running experiments multiple times)?
	    \answerYes{Error bars have been reported on $3$ different runs}
		\item Did you include the total amount of compute and the type of resources used (e.g., type of GPUs, internal cluster, or cloud provider)?
	    \answerYes{In the Supplementary Information at section ~\ref{SI:computational_resources}}
	\end{enumerate}

	\item If you are using existing assets (e.g., code, data, models) or curating/releasing new assets...
	\begin{enumerate}
	  \item If your work uses existing assets, did you cite the creators?
	    \answerYes{All the github links are disclosed in the Supplementary information, and the contributing authors are cited in the main article.}
	  \item Did you mention the license of the assets?
	    \answerNA{}
	  \item Did you include any new assets either in the supplemental material or as a URL?
	    \answerYes{The github link of our code is included in the article}
	  \item Did you discuss whether and how consent was obtained from people whose data you're using/curating?
	    \answerNA{All assets we are using are publicly available, and do not require any consent}
	  \item Did you discuss whether the data you are using/curating contains personally identifiable information or offensive content?
	    \answerNA{No personal information in Omniglot}
	\end{enumerate}
	\item If you used crowdsourcing or conducted research with human subjects...
	\begin{enumerate}
	  \item Did you include the full text of instructions given to participants and screenshots, if applicable?
	    \answerNA{}
	  \item Did you describe any potential participant risks, with links to Institutional Review Board (IRB) approvals, if applicable?
	    \answerNA{}
	  \item Did you include the estimated hourly wage paid to participants and the total amount spent on participant compensation?
	    \answerNA{}
	\end{enumerate}

	\end{enumerate}
\fi
\newpage

\section*{Supplementary Information}
\beginsupplement
\subsection{More details on Prototypical Net}
\label{SI:ProtoNet}
\subsubsection*{Architecture}
Table~\ref{SI:table1} describes the architecture of the Propotypical Net~\citep{snell2017prototypical} we are using in this article. We use the Pytorch convention to describe the layers of the network.

\begin{table}[h!]
  \caption{Description of the Prototypical Net Architecture}
  \centering
  \begin{tabular}{ccc}
    \toprule
    Network & Layer & \# params \\
    \midrule
    \multirow{4}{*}{ConvBlock(In$_{c}$, Out$_{c}$)} & Conv2d(In$_{c}$, Out$_{c}$, 3, padding=1)    &    In$_{c}$ $\times$ Out$_{c}$ $\times$ 3 $\times$ 3 +  Out$_{c}$ \\
    & BatchNorm2d(Out$_{c}$)  & 2 x Out$_{c}$ \\
    & ReLU & - \\
    & MaxPool2d(2, 2)  & - \\
    \midrule
    \multirow{9}{*}{Prototypical Net} & ConvBlock(1, 64) & 0.7 K \\
    & ConvBlock(64, 64) &  37 K\\ 
    & ConvBlock(64, 64) & 37 K\\
    & ConvBlock(64, 64) & 37 K\\
    & Flatten & - \\
    & ReLU & - \\
    & Linear(576, 256) & 147 K\\
    & ReLU & \\
    & Linear(256, 128) & 32 K\\
    \bottomrule
  \end{tabular}
\label{SI:table1}
\end{table}

The overall number of parameters of the Prototypical Net we are using is around 292 K parameters. The loss of the Prototypical Net is applied on the output of the last fully connected layers (of size $128$). For the computation of the samples diversity, we extract the features on the first fully-connected layer after the last convolutional layer (i.e., of size $256$).

\subsubsection*{Training details}
The Prototypical Net is trained in a 1-shot 60-ways setting and tested on a $1$-shot $20$-ways setting. The size of the query set is always $1$ for both training and testing phase. The model is trained during $80$ epochs, with a batch size of $128$. For training, we are using an Adam optimizer~\citep{kingma2014adam} with a learning rate of $1\times10^{-3}$ (all other parameters of the Adam optimizer are the default ones). We are scheduling the learning rate such that it is divided by $2$ every $20$ epochs.

At the end of the training, the training accuracy (evaluated on 1000 episodes) has reached $100\%$ and the testing accuracy reaches a plateau at $96.55\%$.

\subsection{More details on SimCLR}
\label{SI:SimCLR}
\subsubsection{Architecture and Data Augmentation}
The architecture we are using for SimCLR~\citep{chen2020simple} is the exact same than the one used for Prototypical Net (see Table~\ref{SI:table1}). In SimCLR, we also extract the features on the first fully-connected layer after the last convolutional layer (i.e., of size $256$). The augmentations we use are randomly chosen among the $3$ following transformations
\begin{itemize}
    \item \textbf{Random resized crop: } it crops random portion of the image and resizes it to a given size. $2$ sets of parameters are used for this transformation: the scale and the ratio. The scale parameter specifies the lower and upper bounds for the random area of the crop. The ratio parameter specifies the lower and upper bounds for the random aspect ratio of the crop. Our scale range is ($0.1$, $0.9$) and our ratio range is ($0.8$, $1.2$).
    \item \textbf{Random affine transformation: } it applies a random affine transformation of the image while keeping the center invariant. The affine transformation is a combination of a rotation (from $-15^{\circ}$ to $15^{\circ}$), a translation (from $-5$ pixels to $5$ pixels), a zoom (with a ratio from $0.75$ to $1.25$) and a shearing (from $-10^{\circ}$ to $10^{\circ}$).
    \item \textbf{Random perspective transformation: } apply a scale distortion with a certain probability to simulate 3D transformations. The scale distortion we have chosen is $0.5$, and it is applied to the image with a probability of $50\%$
\end{itemize}
Please see the site \url{https://pytorch.org/vision/main/auto_examples/plot_transforms.html} for illustration of the transformations.
Note that we have tried different settings for the augmentations (varying the parameters of the augmentations), and we have observed a very limited impact of those settings on the computation of the samples diversity (see \ref{SI:SimCLRAugmentation effect} for more details).

\subsubsection{Training details}

Our SimCLR network is trained for $100$ epochs with a batch size of $128$. We used an RMSprop optimizer~\citep{hinton2012neural}, with a learning rate of $10^{-3}$ (all other parameters of the RMSprop are the default ones).

\subsection{Control experiments for the samples diversity computation}

\subsubsection{Comparing the supervised and the unsupervised settings for the computation of the samples diversity}
\label{SI:Diversity_supervised_vs_non_supervised}
\begin{figure}[h!]
\centering
\includegraphics[width=0.5\textwidth]{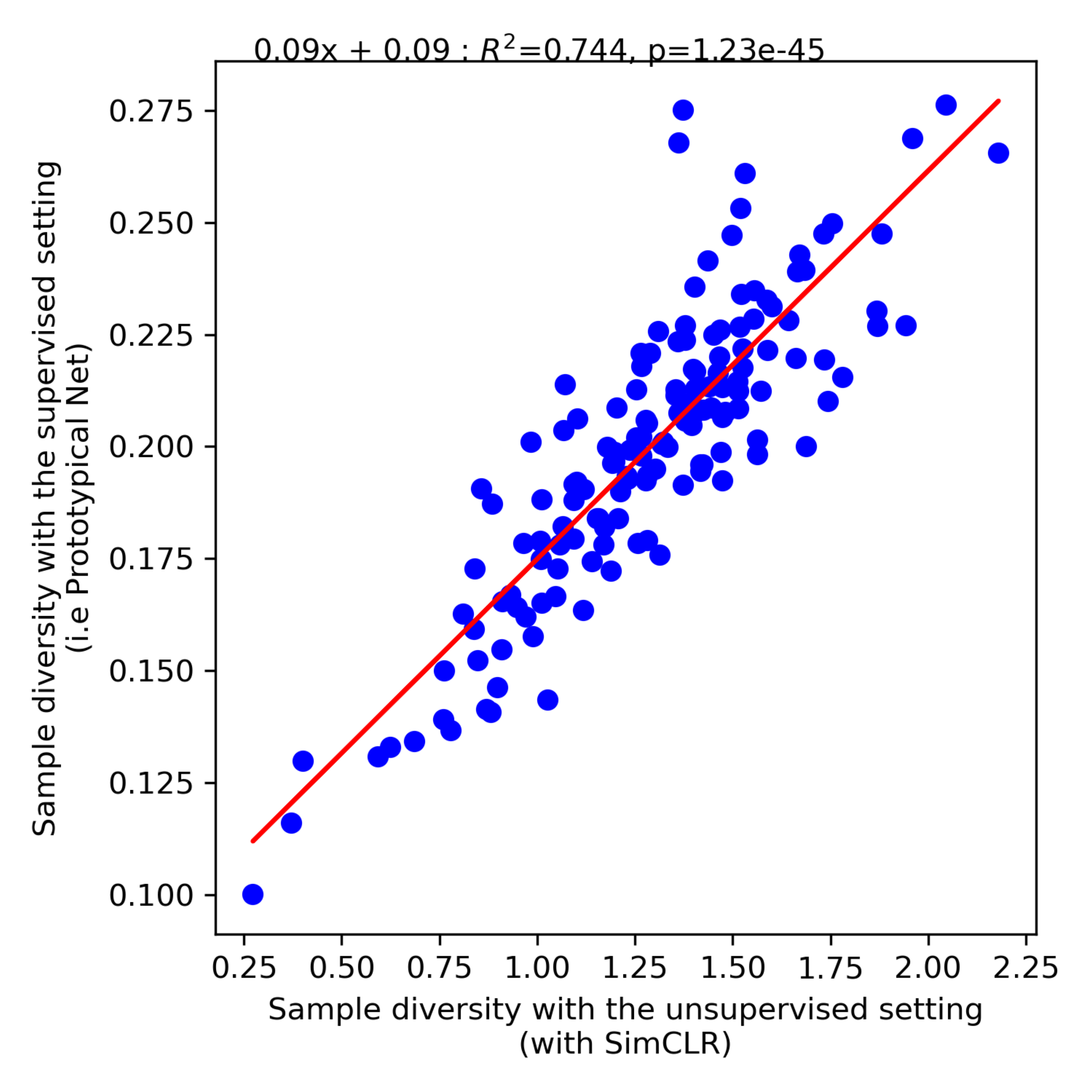}
\caption{Comparison of the samples diversity computed by the supervised and the unsupervised settings. Each data point corresponds to a specific class in the Omniglot test set. Here, the samples diversity is computed applying the standard deviation (see Eq.~\ref{eq:eq2}) on SimCLR features (for x-axis) or on the features of Prototypical Net (for y-axis)}
\label{SI:fig1}
\end{figure}

To compare the unsupervised with the supervised setting, we have computed for all of the $150$ classes of the Omniglot testing set the samples diversity. We plot the samples diversity values for each category and for both settings in Fig.~\ref{SI:fig1}. We report a linear correlation coefficient $R^2=0.74$ and a Spearman rank order correlation $\rho=0.85$ (see Table~\ref{SI:table2}, first line). It does mean that the samples diversity, as computed with one of the setting, is strongly correlated both in terms of rank order and explained variance, with the samples diversity as computed with the other setting.

\begin{table}[h!]
  \caption{Spearman rank order correlation for different settings}
  \label{SI:table2}
  \centering
  \begin{tabular}{cccc}
    \toprule
    Setting 1     & Setting 2     & Spearman correlation & p value \\
    \midrule
     Proto. Net + Eq.~\ref{eq:eq2} & SimCLR + Eq.~\ref{eq:eq2} & $0.85$ & $8.99\times10^{-43}$\\
     Proto. Net + Eq.~\ref{SI:eq1} &  SimCLR + Eq.~\ref{SI:eq1} & $0.71$ & $1.47\times10^{-24}$ \\
     Proto. Net + Eq.~\ref{eq:eq2} &  Proto. Net + Eq.~\ref{SI:eq1} & $0.73$ & $1.19\times10^{-26}$ \\
     SimCLR + Eq.~\ref{eq:eq2} &  SimCLR + Eq.~\ref{SI:eq1} & $0.63$ & $5.21\times10^{-18}$ \\
    \bottomrule
  \end{tabular}
\end{table}

\subsubsection{More control experiments on the effect of the dispersion measure}
\label{SI:More_dispersion_measure}
\begin{figure}[h!]
     \centering
     \captionsetup[subfigure]{justification=centering}
     \begin{subfigure}[b]{0.325\textwidth}
         \centering
         \includegraphics[width=\textwidth]{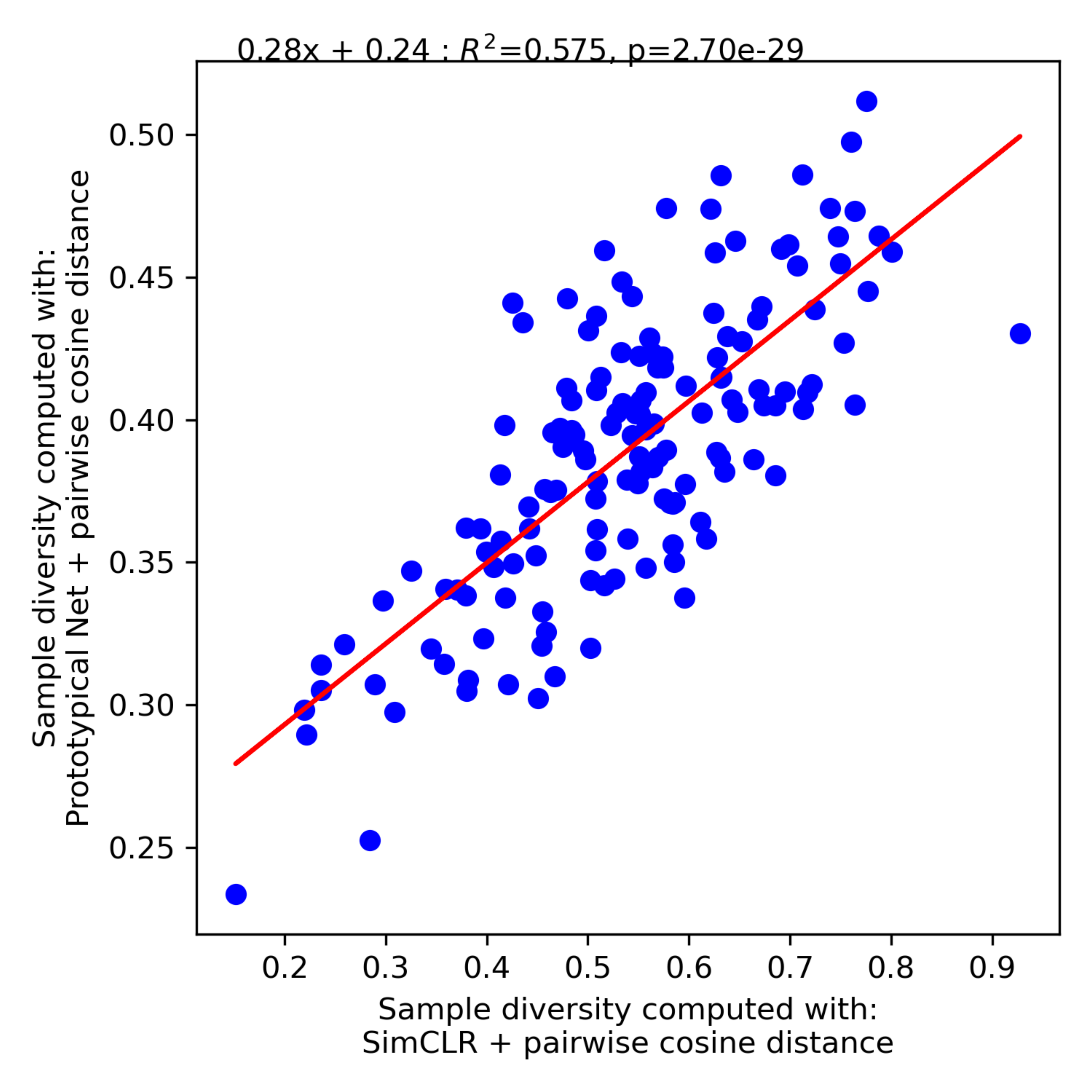}
         \caption{SimCLR + cosine distance vs. Proto Net + cosine distance}
         \label{SI:fig2a}
         
     \end{subfigure}
     \hfill
     \begin{subfigure}[b]{0.325\textwidth}
         \centering
         \includegraphics[width=\textwidth]{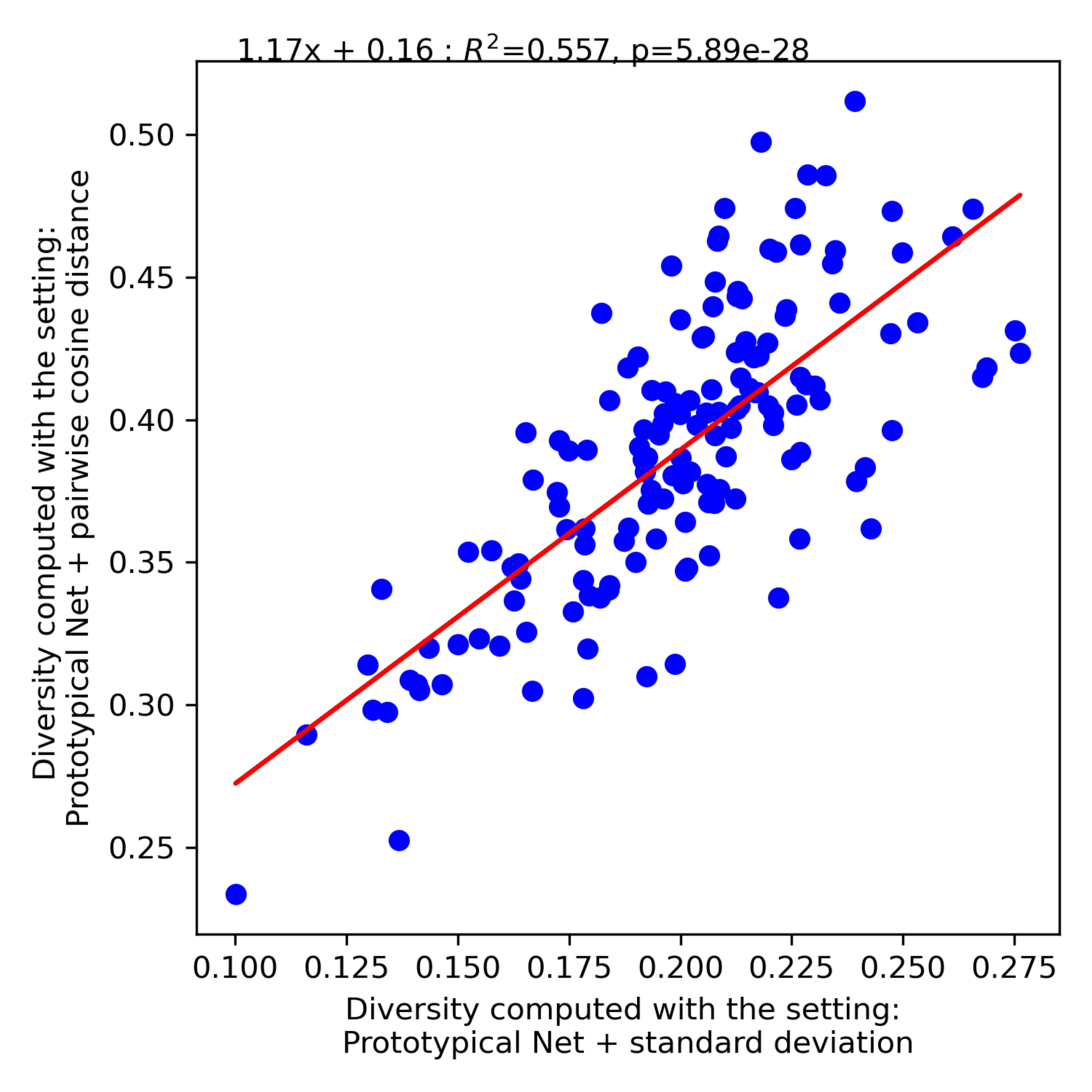}
         \caption{Proto Net + standard deviation vs. Proto Net + cosine distance}
         \label{SI:fig2b}
     \end{subfigure}
     \hfill
     \begin{subfigure}[b]{0.325\textwidth}
         \centering
         \includegraphics[width=\textwidth]{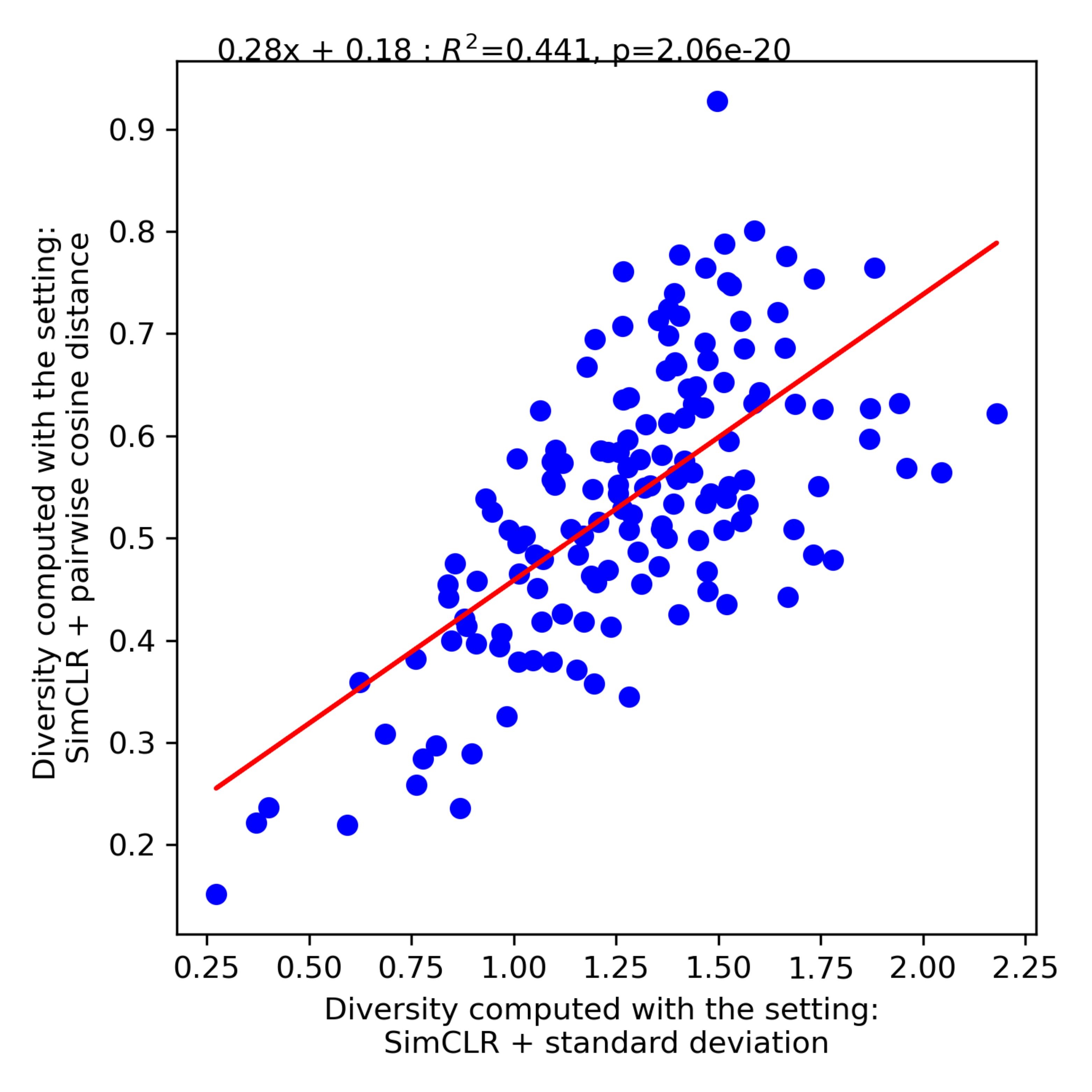}
         \caption{SimCLR + cosine distance vs. Proto Net + standard deviation}
         \label{SI:fig2c}
     \end{subfigure}
     
        \caption{Control experiments when the dispersion metric is the pairwise cosine distance (as defined in Eq.~\ref{SI:eq1}). Each point corresponds to a specific class of the Omniglot testing set. In a) we vary the feature extraction network, while keeping the same dispersion metric (i.e., the pairwise cosine distance). In b) and c), we fix the feature extraction network (Prototypical Net for b) and SimCLR for c)) and we vary the dispersion metric (standard deviation for the x-axis or the pairwise cosine distance for the y-axis).}
        \label{SI:fig2}
\end{figure}

To make our analysis more robust we have conducted additional control experiments with different measures of dispersion. In Eq.~\ref{eq:eq2} we have presented a classical measure of dispersion that is the standard deviation. Another measure of data dispersion is the pair-wise cosine distance among the samples belonging to the same class:
\begin{align}
\sigma_{p_\theta}^j = \sum_{i=1}^{N}\sum_{\substack{k=1 \\ k>i}}^{N}\sqrt{2-2C(f(v_{i}^{j}), f(v_{k}^{j}))} \quad \text{s.t.} \quad v_{i}^{j} \sim p_{\theta}(\cdot|\tilde{x}^{j}) \quad \text{and} \quad C(u,v) = \frac{u \cdot v}{\norm{u}\norm{v}} 
\label{SI:eq1}
\end{align}
In Eq.~\ref{SI:eq1}, $C$ denotes the cosine similarity. In Fig.~\ref{SI:fig2a}, we plot the samples diversity for both feature extraction networks but with a dispersion measure based on the pairwise cosine distance as formulated in Eq.~\ref{SI:eq1}. We report a linear correlation of $R^2=0.57$ and a Spearman rank order correlation of $\rho=0.71$ (see second line of Table~\ref{SI:table2}). This control experiment suggests that even by using a different dispersion metric (i.e., the pairwise cosine distance), the $2$ feature extraction networks produce samples diversity values that are heavily correlated. This strengthen our observation made in~\ref{SI:Diversity_supervised_vs_non_supervised}: the representations produced by the SimCLR and Prototypical Net are similar. Another interesting control experiment is to compare the impact of the dispersion measure on the samples diversity metric. To do so, we have compared the samples diversity computed with one feature extractor (either Prototypical Net in Fig.~\ref{SI:fig2b} or SimCLR in Fig.~\ref{SI:fig2c}) but for $2$ different dispersion metrics (i.e., the standard deviation as formulated in Eq.~\ref{eq:eq2} and the pairwise cosine distance as defined in Eq.~\ref{SI:eq1}). In both cases, we have a non negligible linear correlation (i.e., $R^2>0.44$) and a strong Spearman rank order correlation (i.e., $\rho>0.63$, see third and fourth lines of Table~\ref{SI:table2}). All these control experiments confirm that our computation of the samples diversity is robust to 1) the type of approach we used to extract the features and 2) the measure of dispersion we are using to compute the intraclass variability.

\subsubsection{Impact of the image augmentation on the diversity measure}
\label{SI:SimCLRAugmentation effect}
\begin{figure}[h!]
     \centering
     \captionsetup[subfigure]{justification=centering}
     \begin{subfigure}[b]{0.49\textwidth}
         \centering
        \includegraphics[width=\textwidth]{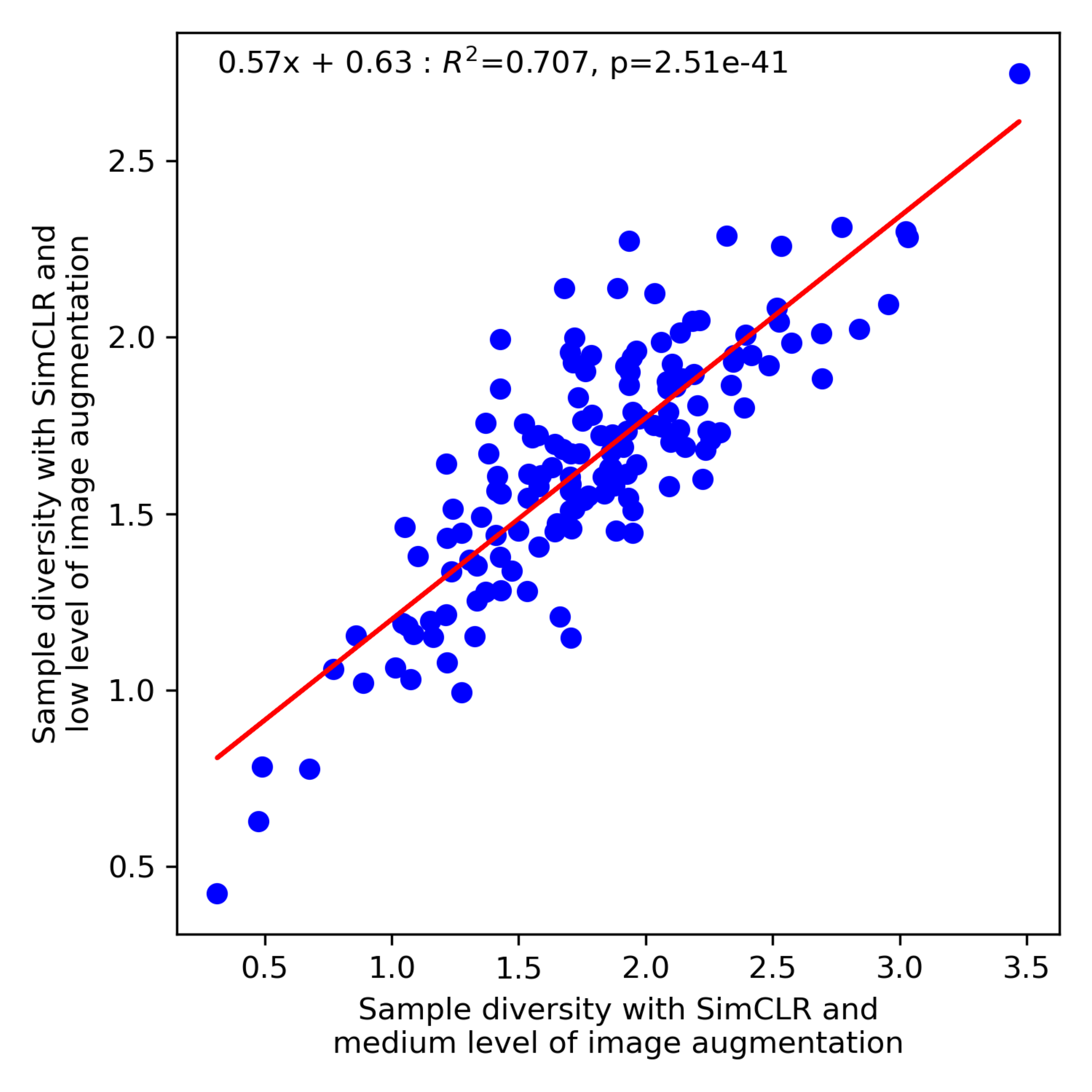}
        \caption{SimCLR with moderate level of image augmentation versus SimCLR with low level of image augmentation}
         \label{SI:fig3a}
     \end{subfigure}
     \hfill
     \begin{subfigure}[b]{0.49\textwidth}
         \centering
         \includegraphics[width=\textwidth]{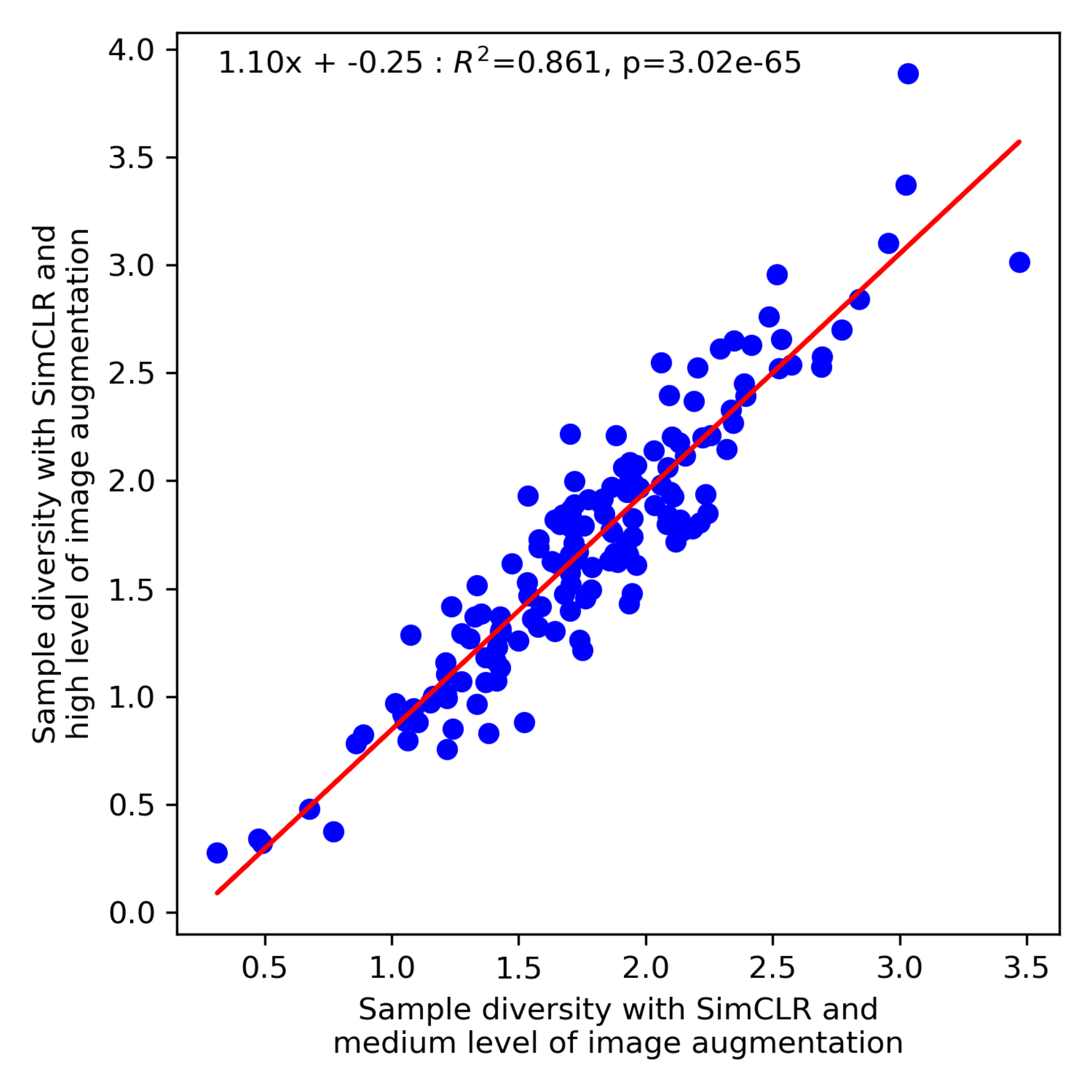}
         \caption{SimCLR with moderate level of image augmentation versus SimCLR with high level of image augmentation}
         \label{SI:fig3b}
     \end{subfigure}
     \caption{Control experiment to assess the impact of the level of image augmentation on the sample metric as evaluated as a standard deviation in the SimCLR feature space.}
    \label{SI:fig3}
\end{figure}

To test the impact of the image augmentations on the SimCLR network we have trained $3$ SimCLR networks with different augmentation levels.
\begin{itemize}
    \item With moderate level of image augmentation. All the augmentations here are those described in section~\ref{SI:SimCLR}.
    \item With a low level of image augmentation. Here the scale of the random resized crop is varied from $0.05$ to $0.95$ and the crop ratio is ranging from $0.9$ to $1.1$. The rotation of the affine transformation is ranging from $-7\deg$ to $7\deg$, the translation from $-3$ pixels to $3$ pixels, the zoom from $0.9$ to $1.1$ and the shearing from $-5\deg$ to $5\deg$. The scale distortion applied to the image is $0.25$ (with a probability of $50$\%).  
    \item With a high level of image augmentation. In this setting, the scale of the random resized crop is varied from $0.2$ to $0.8$ and the crop ratio is ranging from $0.6$ to $1.4$. The rotation of the affine transformation is ranging from $-30\deg$ to $30\deg$, the translation from $-10$ pixels to $10$ pixels, the zoom from $0.5$ to $1.5$ and the shearing from $-20\deg$ to $20\deg$. The scale distortion applied to the image is $0.75$ (with a probability of $50$\%).
\end{itemize}
In Fig.~\ref{SI:fig3}, we compare the samples diversity obtains for each category of the Omniglot testing set when we train the SimCLR network with moderate level of image augmentation and with a low level of image augmentation (see Fig.~\ref{SI:fig3a}), or with a high level of image augmentation (see Fig.~\ref{SI:fig3b}). We also report the Spearman correlation in Table~\ref{SI:table0}.
\begin{table}[h!]
  \caption{Spearman rank order correlation for different settings}
  \label{SI:table0}
  \centering
  \begin{tabular}{cccc}
    \toprule
    Setting 1     & Setting 2     & Spearman correlation & p value \\
    \midrule
     moderate augmentation & light augmentation & $0.79$ & $1.7\times10^{-33}$\\
     moderate augmentation & strong augmentation & $0.90$ &$1.45\times10^{-55}$ \\
    \bottomrule
  \end{tabular}
\end{table}

We observe a high linear correlation as well as a high Spearman rank order correlation between the tested settings. It suggests that the samples diversity is relatively independent to the level of image augmentations used during the SimCLR training.

\subsubsection{T-SNE of the SimCLR and Prototypical Net latent space}
\label{SI:SimCLR_latent_analysis}
\begin{figure}[h!]
     \centering
     \captionsetup[subfigure]{justification=centering}
     \begin{subfigure}[b]{0.49\textwidth}
         \centering
        \includegraphics[width=\textwidth]{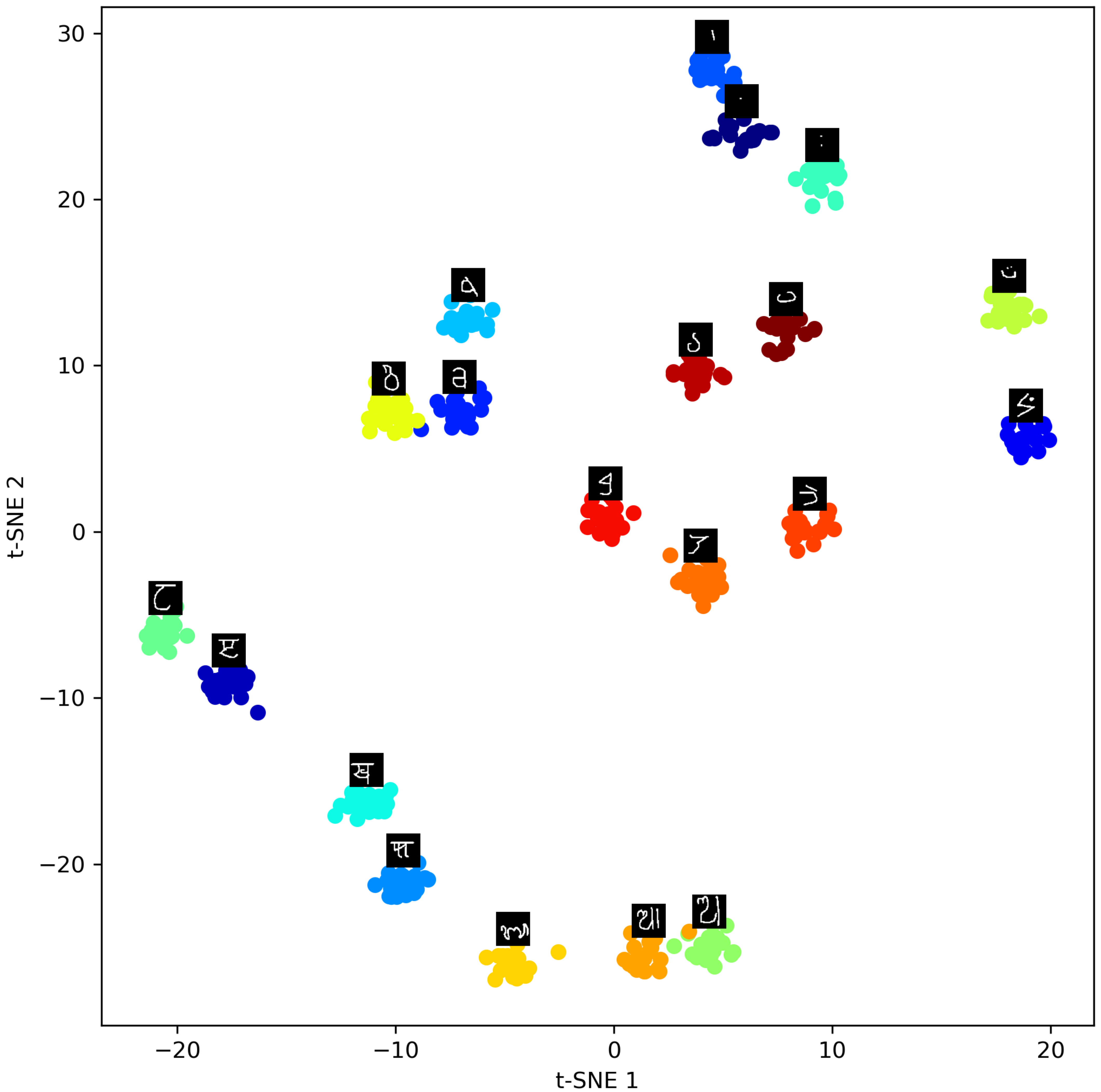}
        \caption{t-SNE of the Prototypical Net feature space}
         \label{SI:fig4a}
     \end{subfigure}
     \hfill
     \begin{subfigure}[b]{0.49\textwidth}
         \centering
         \includegraphics[width=\textwidth]{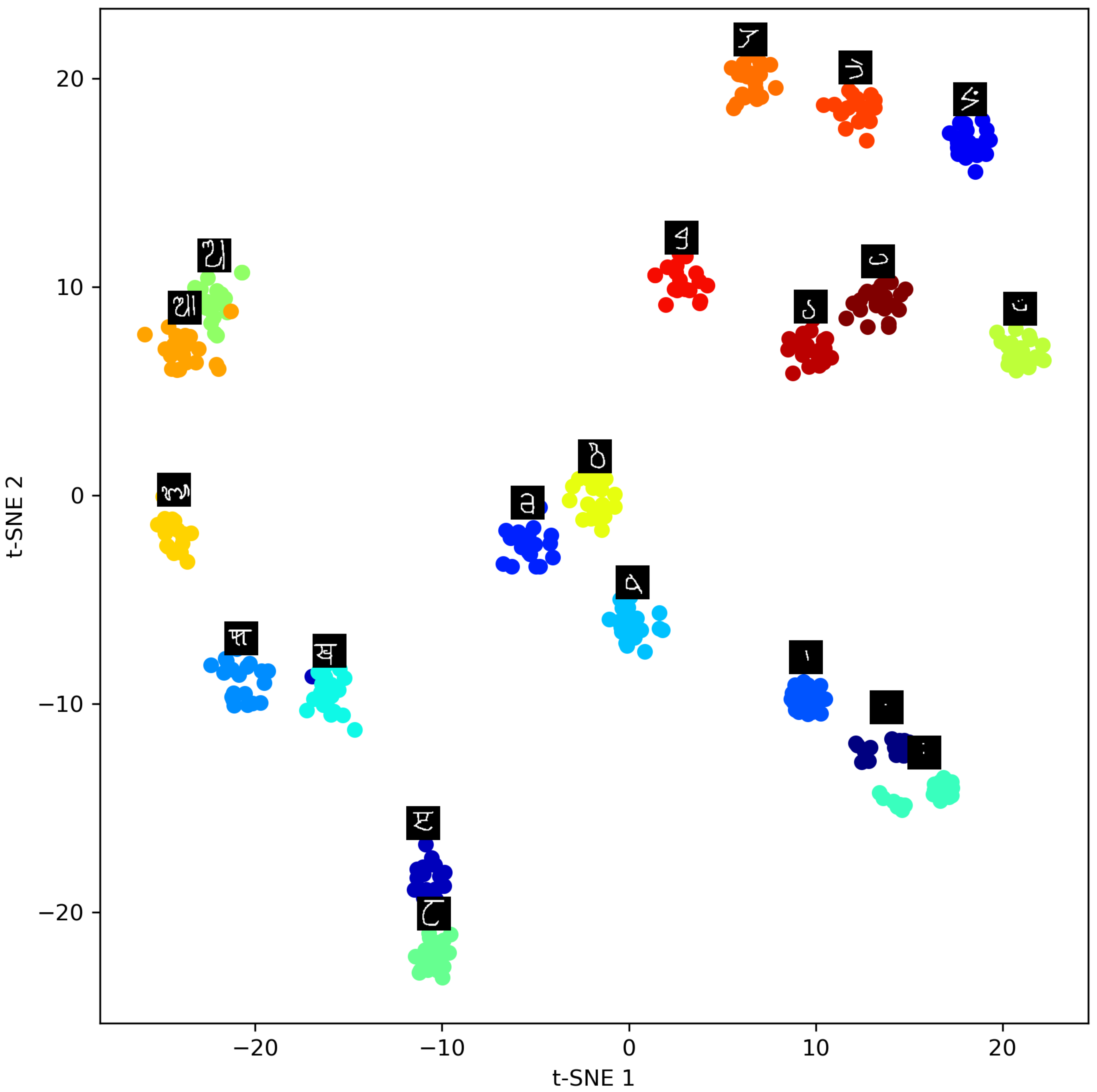}
         \caption{t-SNE of the SimCLR feature space}
         \label{SI:fig4b}
     \end{subfigure}
     \caption{In these $2$ figures the t-SNE analysis has been conducted on the $150$ classes of the testing set of Omniglot. For the sake of clarity we show here a randomly selected subset of those classes (i.e., $20$ classes).}
    \label{SI:fig4}
\end{figure}

In Fig.~\ref{SI:fig4a} and Fig.~\ref{SI:fig4b}, we show a t-SNE analysis of the feature space of Prototypical Net and SimCLR respectively.
In Fig.~\ref{SI:fig4a}, the t-SNE analysis of the Prototypical Net feature space reveals a strong clustering of the samples belonging to the same class. Note that this phenomenon is not surprising as the loss of the Prototypical Net forces the samples belonging to the same class to be close in the feature space. More surprisingly, we also observe a clustering effect in the SimCLR t-SNE analysis (see Fig.~\ref{SI:fig4b}). Note that SimCLR is a fully unsupervised algorithm: there is no class information given to the algorithm. Consequently, the strong clustering effect we observe suggests that forcing the proximity between a sample and its augmented version is enough to retrieve the class information. This observation might explain why contrastive learning algorithms are in general so efficient in semi-supervised (or even unsupervised) classification tasks.

\newpage
\subsection{Concepts ranked by diversity for the unsupervised setting}
\label{SI:ranked_concepts_unsupervised}

\begin{figure}[h!]
\centering
\includegraphics[width=\textwidth]{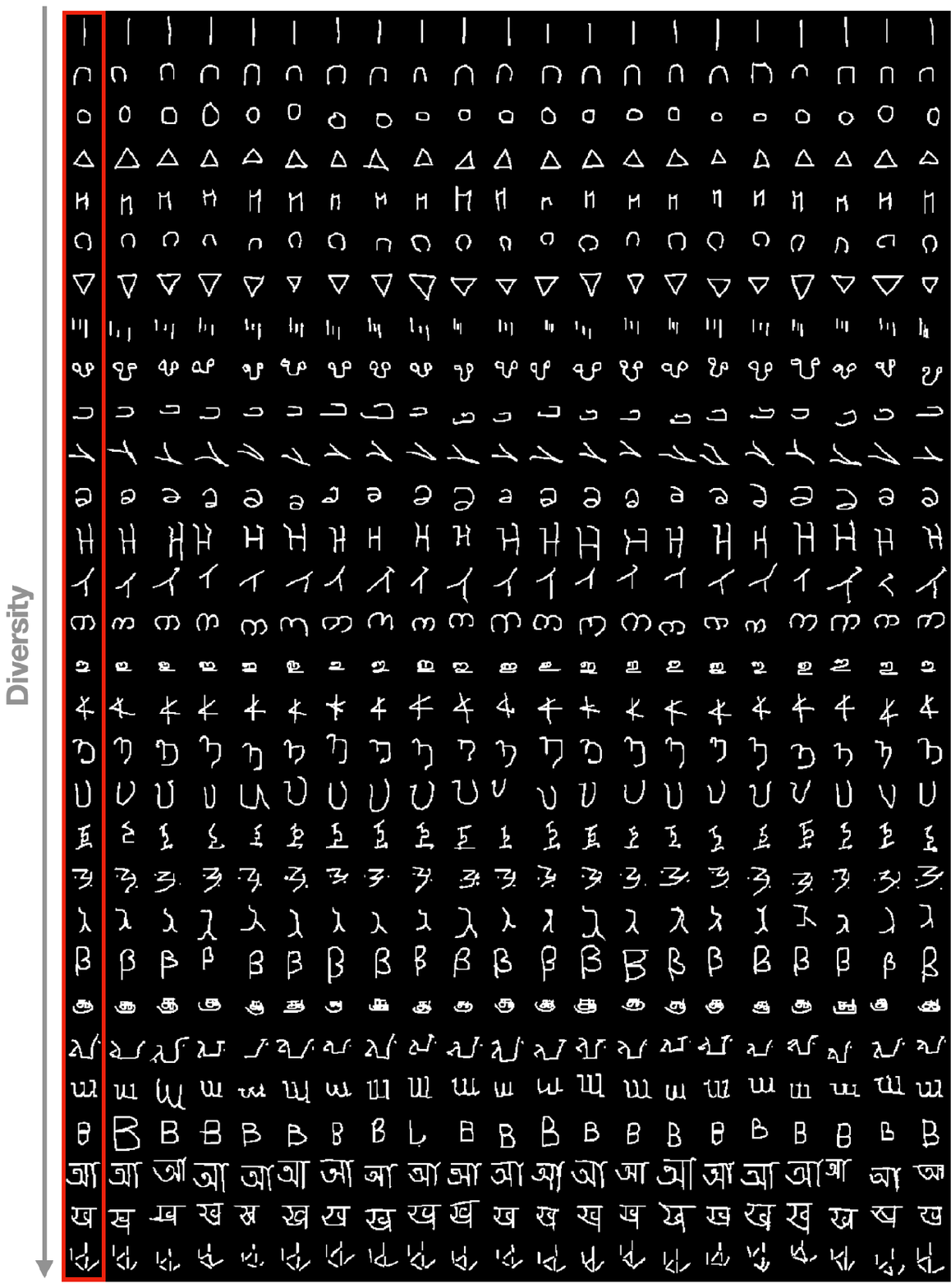}
        \vspace{-20pt}
         \caption{Concepts of the Omniglot test set, ranked by their diversity as computed with the unsupervised setting (i.e., SimCLR as a feature extractor and standard deviation for the dispersion measure). Here we linearly sub-sampled $30$ of out of $150$ concepts of the test set. Concepts are ranked in a increasing order (from low to high diversity). The samples in the red box are the prototypes, the rest of the line is composed with samples belonging to the same category.}
         \label{SI:fig5}
\end{figure}

\newpage
\subsection{Concepts ranked by diversity for the supervised setting}
\label{SI:ranked_concepts_supervised}
\begin{figure}[h!]
\centering
\includegraphics[width=\textwidth]{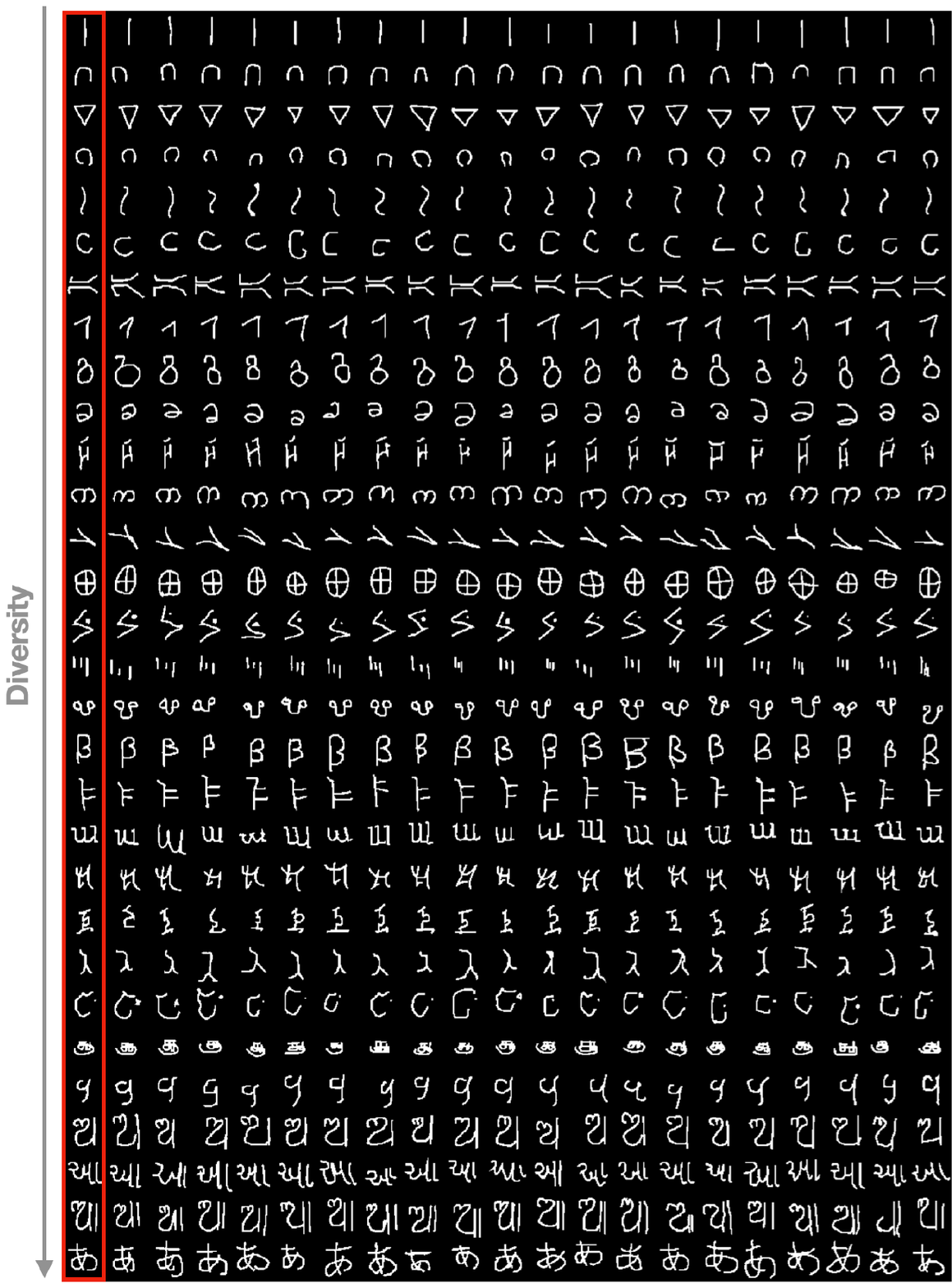}
\vspace{-20pt}
         \caption{Concepts of the Omniglot test set, ranked by their diversity as computed with the supervised setting (i.e., Prototypical Net as a feature extractor and standard deviation for the dispersion measure). Here we linearly sub-sampled $30$ of out of $150$ concepts of the test set. Concepts are ranked in a increasing order (from low to high diversity). The samples in the red box are the prototypes, the rest of the line is composed with samples belonging to the same category.}
         \label{SI:fig6}
\end{figure}

\newpage
\subsection{MAML architecture and training details}
\label{SI:MAML_details}
The architecture we have used for the MAML classifier is exactly the same used for the Prototypical Net (see Table~\ref{SI:table1}). The only difference is the last fully-connected layer that is : Linear(256, 20). Indeed, as the MAML network is directly predicting the logits (and not a distance metric), the last layer needs to have the same dimension than the number of class of the experiment. In a $1$-shot $20$-way classification experiment, the number of classes is $20$.

We have used a $2^{nd}$ order meta-learning scheme~\citep{finn2017model}. The outer-loop optimizer is an Adam optimizer with a learning rate of $10^{-3}$, and the inner-loop optimizer is a simple Stochastic Gradient Decent with a learning rate of $10^{-2}$. The number of inner loops is set to $5$ during the training and to $10$ during the testing. The number of tasks for each outer-loop is set to $4$.

\subsection{Control experiments: Comparing Prototypical Net and MAML}
\label{SI:ctrl_exp_accu}
To rigorously compare MAML and Prototypical Net, we have conducted $2$ types of control experiments. First we have verified whether the classification accuracy obtained for each class were ranked in the same order for both MAML and Prototypical Net. To do so, we have presented the same series of categorization tasks to both algorithms. The high Spearman rank coefficient ($\rho=0.60$) indicates that both classifiers rank each category' classification accuracy similarly (see section~\ref{SI:table3}).

To confirm this result, we have computed the correlation between the logits generated by both models. In the case of the MAML model, extracting the logits is straightforward. For Prototypical Net, we use the distance to prototypes as logits. This explain why both model's logits are anti-correlated: the MAML logits are the (un-normalized) probability of belonging to a given classes whereas the Prototypical Net logits correspond to the distance to the category (so the lower the distance, the higher the probability). We report a strong negative correlation ($r=-0.62$) between the logits of the MAML network and those of Prototypical Net (see section~\ref{SI:table3}).

\begin{table}[h!]
  \caption{Spearman rank order correlation for different settings}
  \label{SI:table3}
  \centering
  \begin{tabular}{cccc}
    \toprule
     Comparison & correlation type &  correlation value & p value \\
    \midrule
     MAML vs. Proto. Net (accuracy) & Spearman & 0.60 & $4.24\times10^{-15}$\\
     MAML vs. Proto. Net (logits) & Pearson & -0.62 & $2.63\times10^{-19}$\\
    
    \bottomrule
  \end{tabular}
\end{table}

\newpage
\subsection{Architecture and training details of the VAE-STN}
\label{SI:VAE_STN_details}
\subsubsection{Architecture of the VAE-STN}
The VAE-STN is a sequential VAE that allows for the iterative construction of a complex image~\citep{rezende2016one}. A pseudo-code of the algorithm is described in Algo~\ref{alg:SI_algo_vae_stn}. At each iteration, the algorithm focuses its attention on a specific part of the image ($\boldsymbol{x}$), the prototype ($\boldsymbol{\tilde{x}}$) and the residual image ($\boldsymbol{\hat{x}}$) using the Reading Spatial Transformer Network (STN${_r}$). Then the extracted patch is passed to an encoding network (EncBlock) to transform it into a latent variable. This latent variable is concatenated to a patch extracted from the prototype and then passed to the RecBlock network. The produced hidden state is first passed to DecBlock to recover the original patch, and then to the STN$_{w}$ to replace and rescale the patch into the original image. The LocNet network is used to learn the parameter of the affine transformation we used in the STN. Note that the affine parameters used in STN$_{w}$ are simply the inverse of those used in STN$_{r}$. 

\begin{algorithm}
\caption{Pseudo-code of the VAE-STN}\label{alg:SI_algo_vae_stn}
\textbf{Input:} \text{image:} \textbf{x}, \text{prototype:} $\tilde{\textbf{x}}$
\begin{algorithmic}
\State $c \gets \textbf{0}$
\State $\boldsymbol{\theta_1} \gets [[1, 0, 0],[0, 1, 0]]$
\State $\boldsymbol{h_1}  \gets \boldsymbol{0}$

\For{$i=1 \; \; \text{to} \; \; N_{steps}$}
\State $\boldsymbol{\hat{x}} = \boldsymbol{x} - \text{sigmoid}(\boldsymbol{c})$
\State $\boldsymbol{r}, \;\boldsymbol{\hat{r}},\; \boldsymbol{\tilde{r}}= STN_r(\boldsymbol{\theta_t}, \boldsymbol{x}), \;STN_r(\boldsymbol{\theta_t}, \boldsymbol{\hat{x}}), \;STN_r(\boldsymbol{\theta_t}, \boldsymbol{\tilde{x}})$
\State $\boldsymbol{r} \gets [\boldsymbol{r}, \boldsymbol{\hat{r}}, \boldsymbol{\tilde{r}},\boldsymbol{h_t} ]$
\State $\boldsymbol{\mu}, \boldsymbol{\sigma} = EncBlock(\boldsymbol{r})$
\State $\boldsymbol{z} = \boldsymbol{\mu} + \boldsymbol{\epsilon}\boldsymbol{\sigma} \;\; \textrm{with} \;\;
\boldsymbol{\epsilon}\sim\mathcal{N}(0,1)$
\State $\boldsymbol{z} \gets [\boldsymbol{z}, \boldsymbol{\tilde{r}}]$
\State $\boldsymbol{p}=DecBlock(\boldsymbol{h_t})$
\State $\boldsymbol{c} \gets \boldsymbol{c} + STN_{w}(\boldsymbol{\theta_{t}^{-1}}, \boldsymbol{p})$
\State $\boldsymbol{h_{t+1}} \gets RecBlock(\boldsymbol{z}, \boldsymbol{h_{t}})$
\State $\boldsymbol{\theta_t+1} \gets LocNet(\boldsymbol{h_{t+1}})$
\EndFor

\end{algorithmic}
\end{algorithm}

The STN modules take $2$ variables in input: an image (or a patch in the case to the STN$_{w}$) and a matrix (3$\times$2) describing the parameters of the affine transformation to apply to the input image~\citep{jaderberg2015spatial}. All other modules are made with MLPs networks, and are described in Table~\ref{SI:table_param_VAE_STN}. In the Table~\ref{SI:table_param_VAE_STN} we use the following notations:
\begin{itemize}
    \item $s_z$: This the size of the latent space. In the base architecture, we set $s_z=80$.
    \item $s_{LSTM}$: This is the size of the output of the Long-Short Term Memory (LSTM) unit. In the base architecture, we set $s_{LSTM}=400$
    \item $s_r$: This is the resolution of the patches extracted by the Spatial Transformer Net (STN) during the reading operation. In the base architecture we set $s_r=15$.
    \item $s_{loc}$: This is the number of neurons used at the input of the localization network. In the base architecture, we set $s_{loc}=100$
    \item $s_{w}$: This is the resolution of the patch passed to the the STN network for the writing operation. In the base architecture $s_w=15$.
\end{itemize}

For the base architecture we used $N_{steps}=60$. The base architecture of the VAE-STN has $6.2$ millions parameters. For more details on the loss function, please refer to \cite{rezende2016one}.
\begin{table}[h!]
  \caption{Description of the VAE-STN architecture}
  \centering
  \begin{tabular}{ccc}
    \toprule
    Network & Layer & \# params \\
    \midrule
    \multirow{9}{*}{EncBlock(s$_r$, s$_{LSTM}$, s$_z$)} & Linear($3$ $\times$ s$_{r}^2$ + s$_{LSTM}$ , $1024$) &   \thead{($3$ $\times$ s$_{r}^2$ + s$_{LSTM}$) $\times$ $1024$)\\ + $1024$}\\
    & ReLU \\
    & Linear($1024$, $1024$) & $1050$ K \\
    & ReLU \\
    & Linear($1024$, $512$) & $524$ K \\
    & ReLU  & - \\
    & Linear($512$, $128$) & $65$ K \\
    & ReLU  & - \\
    & Linear($128$, $2\times$ s$_{z}$) & $256\times$s$_{z}$ + 2$\times$s$_{z}$ \\
    \midrule
    \multirow{4}{*}{LocNet(s$_{loc}$)} & Linear(s$_{loc}$, $64$) &   s$_{loc}\times64$ + $64$\\
    & ReLU & - \\
    & Linear($64$, $32$) & $2$ K \\
    & ReLU  & - \\
    & Linear($32$, $6$) & $0.2$ K \\
    \midrule
    \multirow{7}{*}{DecBlock(s$_{LSTM}$, s$_{loc}$, s$_{w}$)} & Linear(s$_{LSTM}$ - s$_{loc}$, $1024$) & (s$_{LSTM}$ - s$_{loc}$)$\times1024$ + $1024$ \\
    & ReLU & - \\
    & Linear($1024$, $512$) & $525$ K \\ 
    & ReLU & - \\
    & Linear($512$, $256$) & $131$ K \\ 
    & ReLU & - \\
    & Linear($256$, s$_{w}^2$) & $256\times$s$_{w}^2 + $s$_{w}^2$ \\ 
    \midrule
    \multirow{1}{*}{RecBlock(s$_z$, s$_r$, s$_{LSTM}$)} & LSTMCell(s$_z$ + s$_{r}^2$, s$_{LSTM}$) & \thead{$4\times\big($s$_z$ + s$_{r}^2$)$\times$s$_{LSTM}$ \\ + s$_{LSTM}^2$ + s$_{LSTM}$\big)} \\
    \midrule
    \multirow{4}{*}{VAE-STN} & EncBlock($15$ , $800$, $80$) & $3,172$ K \\
    & RecBlock($80$, $15$, $800$) & $1,600$ K \\
    & DecBlock($400$, $100$, $15$) & $1,431$ K \\ 
    & LocNet($100$) & $8.7 K$ \\
    \bottomrule
  \end{tabular}
\label{SI:table_param_VAE_STN}
\end{table}

\subsubsection{Training details of the VAE-STN}
The VAE-STN is trained for $500$ epochs, with batches of size $128$. We use an Adam optimizer with a learning rate of $1\times10^{-3}$ and $\beta_1=0.9$. All other parameters are the default Pytorch parameters. To avoid training instabilities we clip the norm of the gradient to $5$. The learning rate was divided by $2$ when the evaluation loss has not decreased for $10$ epochs (reduce on plateau strategy).

\newpage
\subsubsection{VAE-STN samples}
\label{SI:VAE_STN_samples}
\begin{figure}[h!]
\centering
\includegraphics[width=\textwidth]{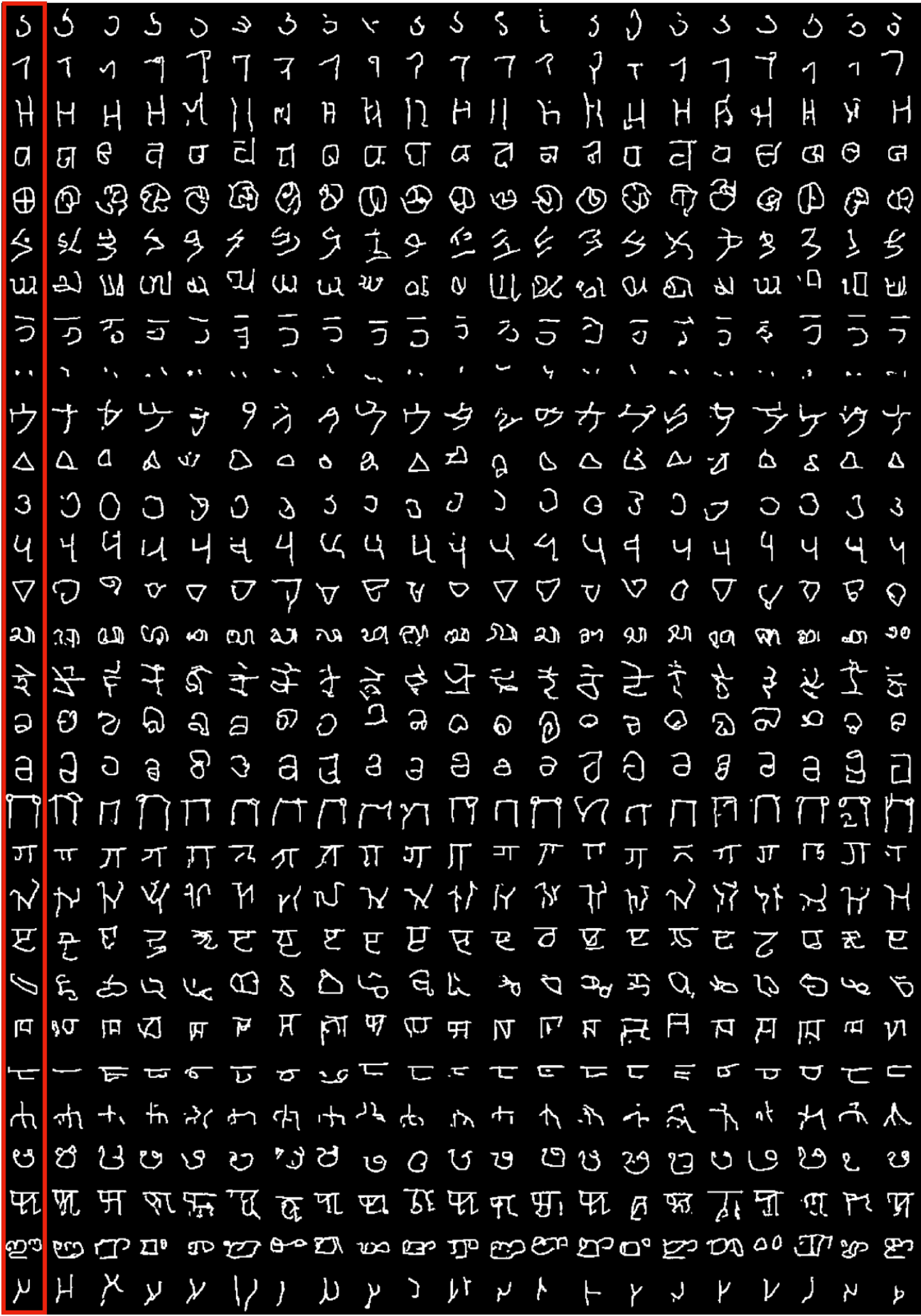}
         \caption{Sampled generated by the \VAESTN. All the prototypes used to condition the generative model are in the red frame. The $30$ concepts has been randomly sampled (out of $150$ concepts) from the Omniglot test set. The lines are composed with $20$ samples that has been generated by the VAE-STN.}
         \label{SI:fig7}
\end{figure}

\newpage
\subsection{Architecture and training details of the Neural Statistician}
\label{SI:NS_details}
\subsubsection*{Architecture}

\begin{figure}[h!]
\centering
\includegraphics[width=\textwidth]{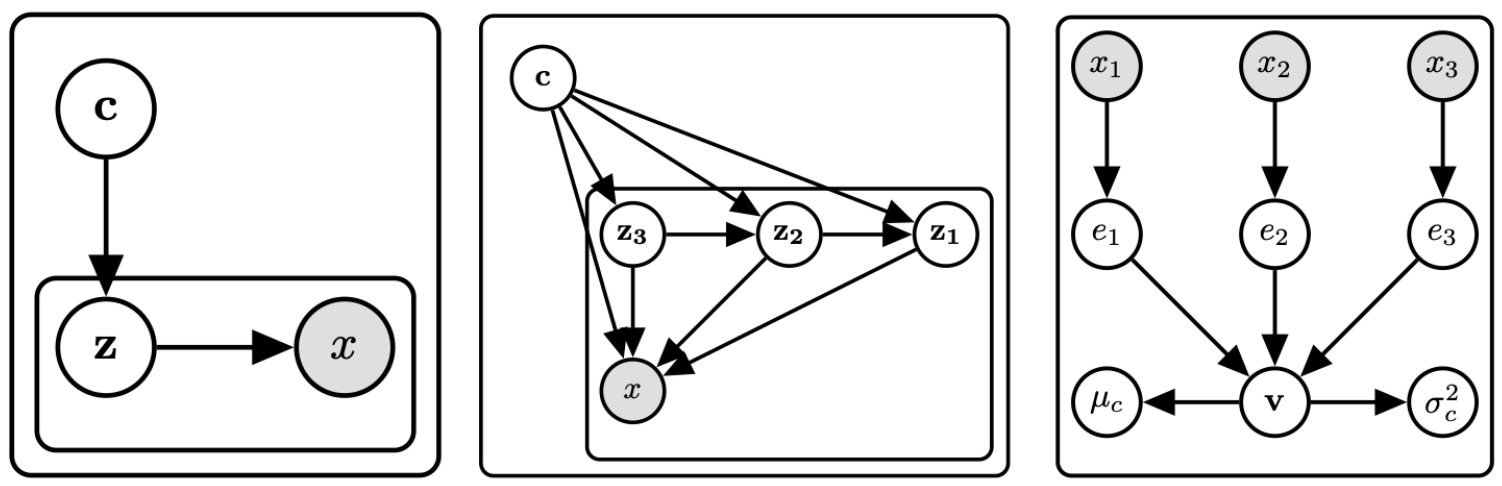}
         \caption{\textit{Left}: basic hierarchical model, where the plate encodes the fact that the context variable $c$ is shared across each item in a given dataset. \textit{Center}: full neural statistician model with three latent layers $z_1,z_2,z_3$. Each collection of incoming edges to a node is implemented as a neural network, the input of which is the concatenation of the edges’ sources, the output of which is a parameterization of a distribution over the random variable represented by that node. \textit{Right}: The statistic network, which combines the data via an exchangeable statistic layer. The above figures were obtained from ~\citep{edwards2016towards} }
         \label{SI:fig8bis}
\end{figure}

Table~\ref{SI:table_param_NS} describes the base architecture of the Neural Statistician model adopted from ~\citep{giannone2021hierarchical} which is a close approximation of  ~\citep{edwards2016towards}. We make minor changes in the network architecture to accommodate the higher input image size of $50\times50$ of the Omniglot dataset. The Neural Statistician model is composed of the following sub-networks:

\begin{itemize}

\item Shared encoder $x \mapsto h$: An instance encoder $E$ that takes each individual datapoint $x_i$ to a feature representation $h_i$ = $E(x_i)$.

\item Statistic network $q(c|D,\phi):h_1,...,h_k \mapsto \mu_c, {\sigma^2}_c$: A pooling layer that aggregates the matrix $(h_1,...,h_k)$ to a single pre-statistic vector $v$. ~\citep{edwards2016towards} uses sample mean for their experiments. Which is followed by a post-pooling network that takes $v$ to a parametrization of a Gaussian.

\item Inference network $q(z|x, c, \phi) : h, c \mapsto \mu_z , {\sigma^2}_z$: Inference network gives an approximate posterior over latent variables.

\item Latent decoder network $p(z|c; \theta) : c \mapsto \mu_z , {\sigma^2}_z$

\item Observation decoder network $p(x|c, z; \theta) : c, z \mapsto \mu_x$

\end{itemize}

\begin{table}[h!]
  \caption{Description of the Neural Statistician Architecture}
  \centering
  \begin{tabular}{ccc}
    \toprule
    Network & Layer & \# params \\
    
    \midrule
    \multirow{2}{*}{ConvBlock(In$_{c}$, Out$_{c}$, stride)}  & 
    Conv2d(In$_{c}$, Out$_{c}$, stride, 3, padding=1)   &    \thead{In$_{c}$ $\times$ Out$_{c}$ $\times$ 3 $\times$ 3 \\+  Out$_{c}$} \\
    & BatchNorm2d(Out$_{c}$), ELU  & 2 x Out$_{c}$ \\
    
    \midrule
    \multirow{2}{*}{FcBlock(In, Out)}  & 
    Linear(In, Out)   &    In $\times$ Out \\
    & BatchNorm1d(Out), ELU  & - \\
    
    \midrule
    \multirow{2}{*}{DeConvBlock(In$_{c}$, Out$_{c}$)}  & 
    ConvTranspose2d(In$_{c}$, Out$_{c}$, 2, 2)   &    \thead{In$_{c}$ $\times$ Out$_{c}$ $\times$ 3 $\times$ 3 \\+  Out$_{c}$} \\
    & BatchNorm2d(Out$_{c}$), ELU  & 2 x Out$_{c}$ \\


    \midrule
    \multirow{12}{*}{Shared encoder}  & 
    ConvBlock(1, 32, 1) &\multirow{12}{*}{1,958,400} \\
    & ConvBlock(32, 32, 1) \\
    & ConvBlock(32, 32, 2) \\
    & ConvBlock(32, 64, 1) \\
    & ConvBlock(64, 64, 1) \\
    & ConvBlock(64, 64, 2) \\
    & ConvBlock(64, 128, 1) \\
    & ConvBlock(128, 128, 1) \\
    & ConvBlock(128, 128, 2) \\
    & ConvBlock(128, 256, 1) \\
    & ConvBlock(256, 256, 1) \\
    & ConvBlock(256, 256, 2) \\

    \midrule
    \multirow{5}{*}{Statistic network} & 
    FcBlock(256*4*4, 256) &\multirow{4}{*}{1,445,122} \\
    & average pooling within each dataset\\ 
    & 2× FcBlock(256, 256) \\
    & Linear(256, 512), BatchNorm1d(1) to $\mu_c$, ${\log {\sigma^2}_c}$\\ 
    
    \midrule
    \multirow{4}{*}{Inference network} & 
    FcBlock(256, 256) $\mapsto$ $h$ & \multirow{4}{*}{408,610} \\
    & FcBlock(512, 256) $\mapsto$ $c$\\
    & combine c and h, ELU\\ 
    & Residual Block\{3× FcBlock(256, 256)\} \\
    & Linear(256, 32), BatchNorm1d(1) to $\mu_z$, ${\log {\sigma^2}_z}$ \\ 
    
    \midrule
    \multirow{3}{*}{Latent decoder network} & 
    Linear(512, 256) $\mapsto$ $c$, ELU & \multirow{3}{*}{342,818} \\
    & Residual Block\{3× FcBlock(256, 256)\}  \\
    & Linear(256, 32), BatchNorm1d(1) to $\mu_z$, ${\log {\sigma^2}_z}$ \\ 
    
    \midrule
    \multirow{16}{*}{Observation decoder network} & 
    FcBlock(512, 256)  $\mapsto$ $z$ & \multirow{16}{*}{3,324,673} \\
    & FcBlock(512, 256)  $\mapsto$ $c$  \\
    & combine z and c, ELU  \\
    & FcBlock(256, 256*4*4) \\
    & ConvBlock(256, 256, 1) \\
    & ConvBlock(256, 256, 1) \\
    & DeConvBlock(256, 256) \\
    & ConvBlock(256, 128, 1) \\
    & ConvBlock(128, 128, 1) \\
    & DeConvBlock(128, 128) \\
    & ConvBlock(128, 64, 1) \\
    & Conv2d(64, 64, 4, 1, 0) \\
    & DeConvBlock(64, 64) \\
    & ConvBlock(64, 32, 1) \\
    & Conv2d(32, 32, 2, 1, 0) \\
    & DeConvBlock(32, 32) \\
    & Conv2d(32, 1, 1) \\
    \bottomrule
  \end{tabular}
\label{SI:table_param_NS}
\end{table}

The overall number of parameters of the base model (which has the same architecture as used in ~\citep{edwards2016towards}) for the Neural Statistician we are using is around 7.48M parameters.

\subsubsection*{Training details}
The Neural Statistician is trained for 300 epochs, with batch size of 32 and learning rate of $1\times10^{-3}$. We adopt the same setting of the Neural Statistician as used in  ~\citep{edwards2016towards} for the omniglot dataset. We constructed context sets by splitting each class into datasets of size 5 while training, and use a single out-of-distribution exemplar while testing. As discussed in the paper, we create new classes by reflecting and rotating characters. We based our implementation from \url{https://github.com/georgosgeorgos/hierarchical-few-shot-generative-models} and \url{https://github.com/comRamona/Neural-Statistician}.

\subsubsection*{Intuition about context integration in the Neural-Statistician}
In the Neural Statistician, the context correspond to the samples used during training, to evaluate the statistics of a specific category (i.e. a concept). In practice, we pass to the network different samples representing the same concept and we vary the number of these samples (from $2$ to $20$ in the experiment described in section~\ref{MAIN:hyper_parameters_effect}). Intuitively, with more context samples for a given category, it becomes easier for the network to identify the properties and features that are crucial to define a given handwritten letter (which results in a higher recognizability but leaves less room for diversity).

\newpage
\subsubsection{Neural statistician samples}
\label{SI:NS_samples}
\begin{figure}[h!]
\centering
\includegraphics[width=\textwidth]{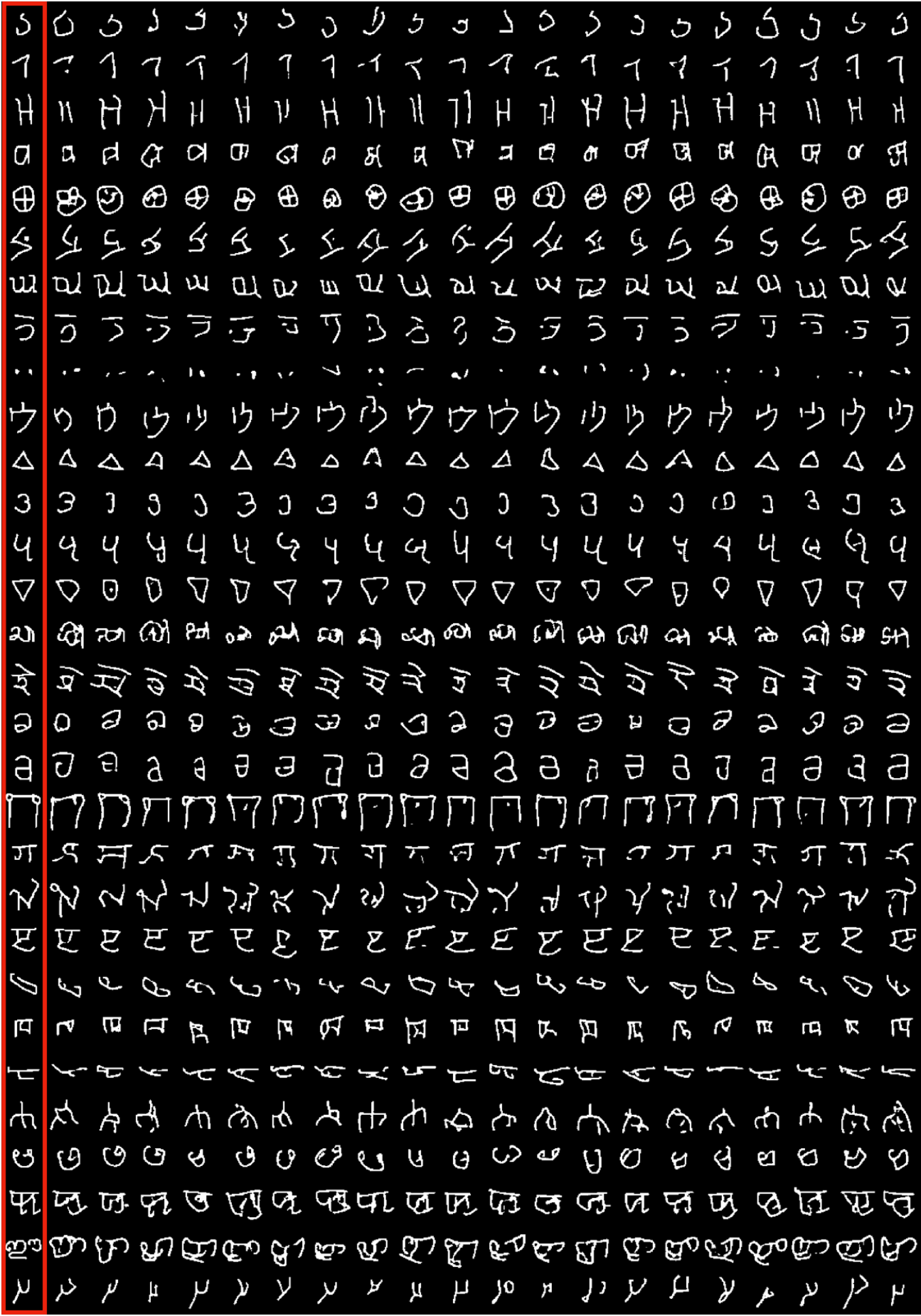}
         \caption{Sampled generated by the neural statistician network (\VAENS). All the prototypes used to condition the generative model are in the red frame. The $30$ concepts has been randomly sampled (out of $150$ concepts) from the Omniglot test set. The lines are composed with $20$ samples that has been generated by the VAE-NS.}
         \label{SI:fig8}
\end{figure}

\newpage
\subsection{Architecture and training details of the DA-GAN based on U-Net (\DAGANUN)}
\label{SI:DAGAN_details}
\subsection*{Architecture}

\begin{figure}[h!]
\centering
\includegraphics[width=0.5\textwidth]{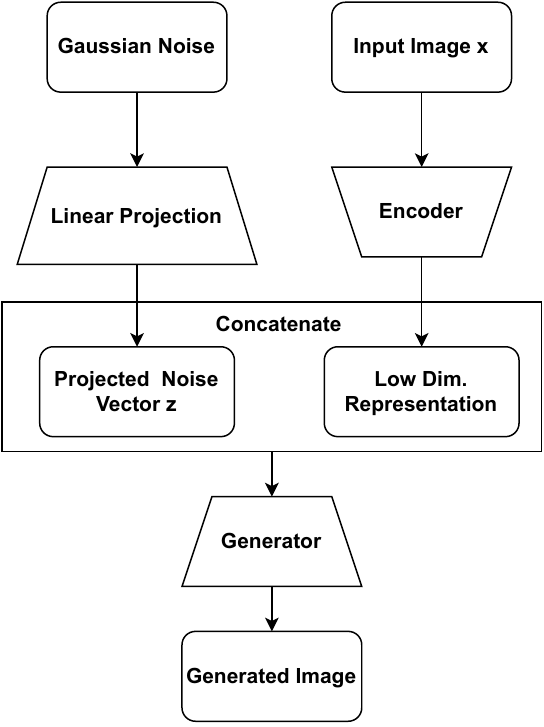}
         \caption{DAGAN Generator: The generator is composed of an encoder projecting the input image to a lower dimensional manifold. A random gaussian noise vector is transformed and concatenated with the bottleneck vector. The resulting vector is passed through the decoder (generator), which outputs the augmented image. }
         \label{SI:fig9bis}
\end{figure}

Table~\ref{SI:table_param_DAGAN-UN} describes the base architecture of the DA-GAN-UN's Generator model adopted from ~\citep{antoniou2017data}. We have modified the architecture of the DA-GAN-UN model such that it can accommodate a higher input image size  $ 50\times50$. Also, we reduced the number of trainable parameters in the original DA-GAN-UN architecture to have a fair comparison with other few-shot models. Following are the notations used in Table~\ref{SI:table_param_DAGAN-UN}:

\begin{itemize}

\item s$_z$: This is the size of the latent space. In the base architecture, we set s$_z$ = 128
\item Generator $G(x,z)$: A generator network that takes data points and Gaussian noise as input, and generate new samples.

\end{itemize}

The base architecture of the DAGAN model we are using in our experiments has around 6.8 million parameters. 

\subsubsection*{Training details}

The DA-GAN-UN model was trained for 30 epochs, with batches of size 32. We use an Adam optimizer with a learning rate of 1 $\times 10^{-4}$ and $\beta_{1}$ = 0.9. We update our generator after every 5 updates of discriminator. We based our implementation from \url{https://github.com/amurthy1/dagan_torch}

\begin{table}[h!]
 \caption{Description of the Data Augmentation GAN Architecture}
 \centering
 \begin{tabular}{ccc}
    \toprule
    Network & Layer & \# params \\
    
    \midrule
    \multirow{2}{*}{ConvBlock(In$_{c}$, Out$_{c}$, s$_{l}$)}  & 
    Conv2d(In$_{c}$, Out$_{c}$, 3, stride=s$_{l}$, padding=1)   &    Out$_{c}$ $\times$ (In$_{c}$ $\times$ 3 $\times$ 3 + 1)    \\
    & LeakyReLU(0.2), BatchNorm2d(Out$_{c}$)   & 2 x Out$_{c}$ \\
    
\midrule

    \multirow{2}{*}{DeConvBlock(In$_{c}$, Out$_{c}$, s$_{l}$)}  & 
    ConvTranspose2d(In$_{c}$, Out$_{c}$, 3, stride=s$_{l}$, padding=1)   &    Out$_{c}$ $\times$ (In$_{c}$ $\times$ 3 $\times$ 3 + 1)    \\
    & LeakyReLU(0.2), BatchNorm2d(Out$_{c}$)   & 2 x Out$_{c}$ \\
    
\midrule
    
    \multirow{4}{*}{EncoderBlock(In$_{p}$, In$_{c}$, Out$_{c}$)}  & 
    ConvBlock(In$_{p}$, In$_{p}$)   &      \\
    & ConvBlock(In$_{c} $ + In$_{p}$, Out$_{c}$)   &    \\
    & Conv2d(In$_{c} $+ Out$_{c}$, Out$_{c}$)   &    \\
    & Conv2d(In$_{c} $+ 2 $\times$ Out$_{c}$, Out$_{c}$)   &  \\

\midrule

\multirow{8}{*}{DecoderBlock(In$_{p}$, In$_{c}$, Out$_{c}$)}  & 
    DeConvBlock(In$_{p}$, In$_{p}$, 1)   & \\
    & ConvBlock(In$_{c}$+In$_{p}$, Out$_{c}$, 1)   &    \\
    & DeConvBlock(In$_{p}$, In$_{p}$, 1)   &  \\
    & ConvBlock(In$_{c}$ + In$_{p}$ + Out$_{c}$, Out$_{c}$, 1)   & \\
    & DeConvBlock(In$_{c}$ + 2 $\times$ Out$_{c}$, Out$_{c}$, 1)   &     \\
    
\midrule

\multirow{18}{*}{Generator(s$_z$)}  & 
    ConvBlock(1, 64, 2)   &  \multirow{17}{*}{6,813,857}  \\
    & EncoderBlock(1, 64, 64)  \\
    & EncoderBlock(64, 64, 128)  \\
    & EncoderBlock(128, 128, 128)  \\
    & Linear(s$_z$, 4$\times$4$\times$8) \\
    & DecoderBlock(0, 136, 64)\\
    & Linear(s$_z$, 7$\times$7$\times$4)  \\
    & DecoderBlock(128, 260, 64) &\\
    & Linear(s$_z$, 13$\times$13$\times$2) \\
    & DecoderBlock(128, 194, 64) \\
    & DecoderBlock(64, 128, 64)  \\
    & DecoderBlock(64, 65, 64) \\
    & ConvBlock(64, 64, 1)   & \\
    & ConvBlock(64, 64, 1)   & \\
    & Conv2d(64, 1, 3, stride=1, padding=1)    \\
    
    \bottomrule
\end{tabular}
\label{SI:table_param_DAGAN-UN}
\end{table}

\newpage

\subsubsection{DA-GAN-UN samples}
\label{SI:DAGAN_samples}
\begin{figure}[h!]
\centering
\includegraphics[width=\textwidth]{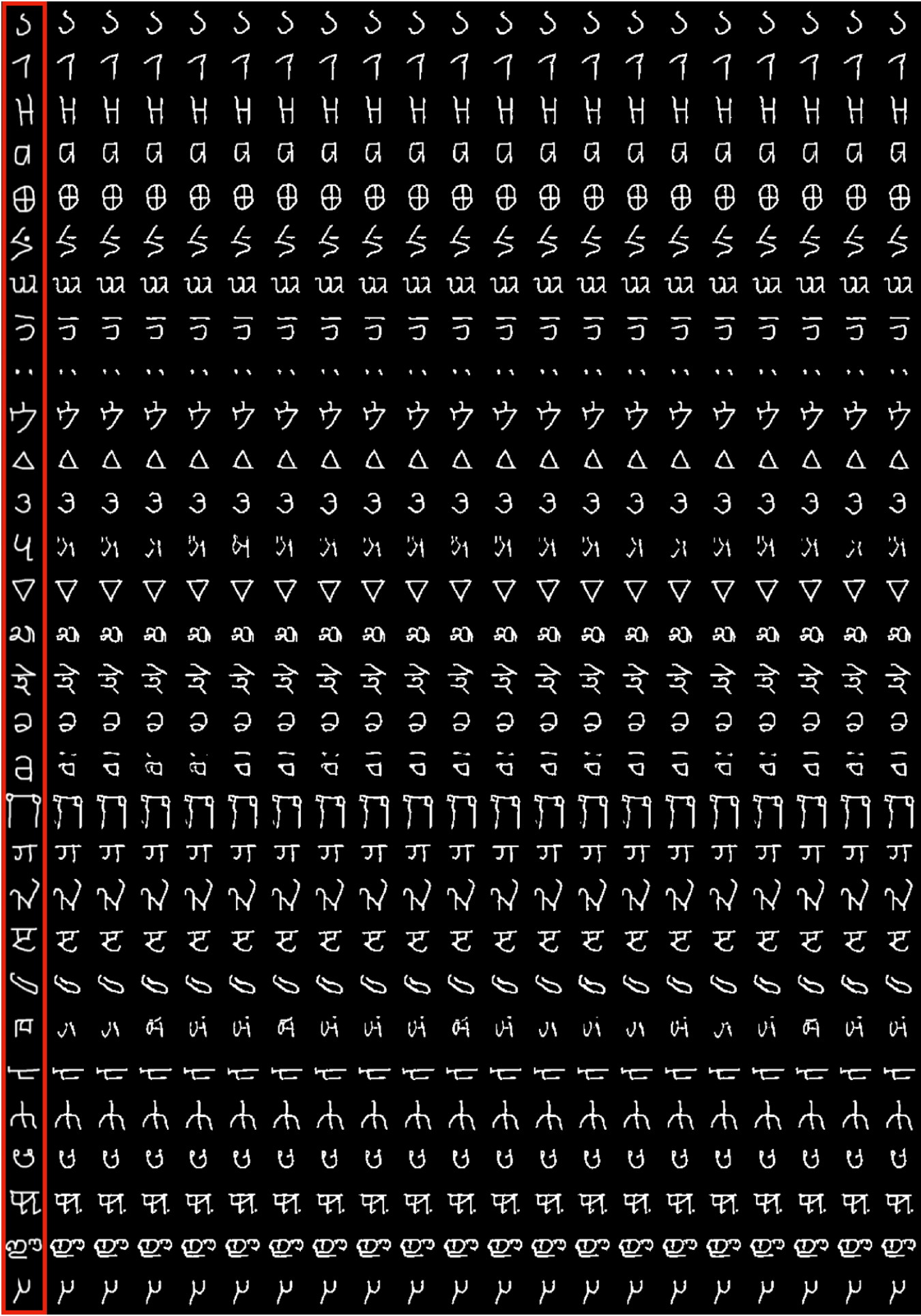}
         \caption{Sampled generated by the Data Augmentation GAN with U-Net architecture (\DAGANUN). All the prototypes used to condition the generative model are in the red frame. The $30$ concepts has been randomly sampled (out of $150$ concepts) from the Omniglot test set. The lines are composed with $20$ samples that has been generated by the DA-GAN-UN.}
         \label{SI:fig9}
\end{figure}

\newpage

\subsection{Architecture and training details of the DA-GAN based on ResNet (\DAGANRN)}
\label{SI:DAGANRN_details}

\subsubsection*{Architecture}

We use the same base architecture of \DAGANUN, except we remove the skip connections between the contracting path (encoder) and the expansive path (decoder). ~\citep{antoniou2017data} used a combination of UNet and ResNet in their results, in \DAGANRN~we consider only a ResNet type architecture.

\subsubsection*{Training details}
Refer ~\ref{SI:DAGAN_details} for training details.

\newpage
\subsubsection{DA-GAN-RN samples}
\label{SI:DAGAN_RN_samples}
\begin{figure}[h!]
\centering
\includegraphics[width=\textwidth]{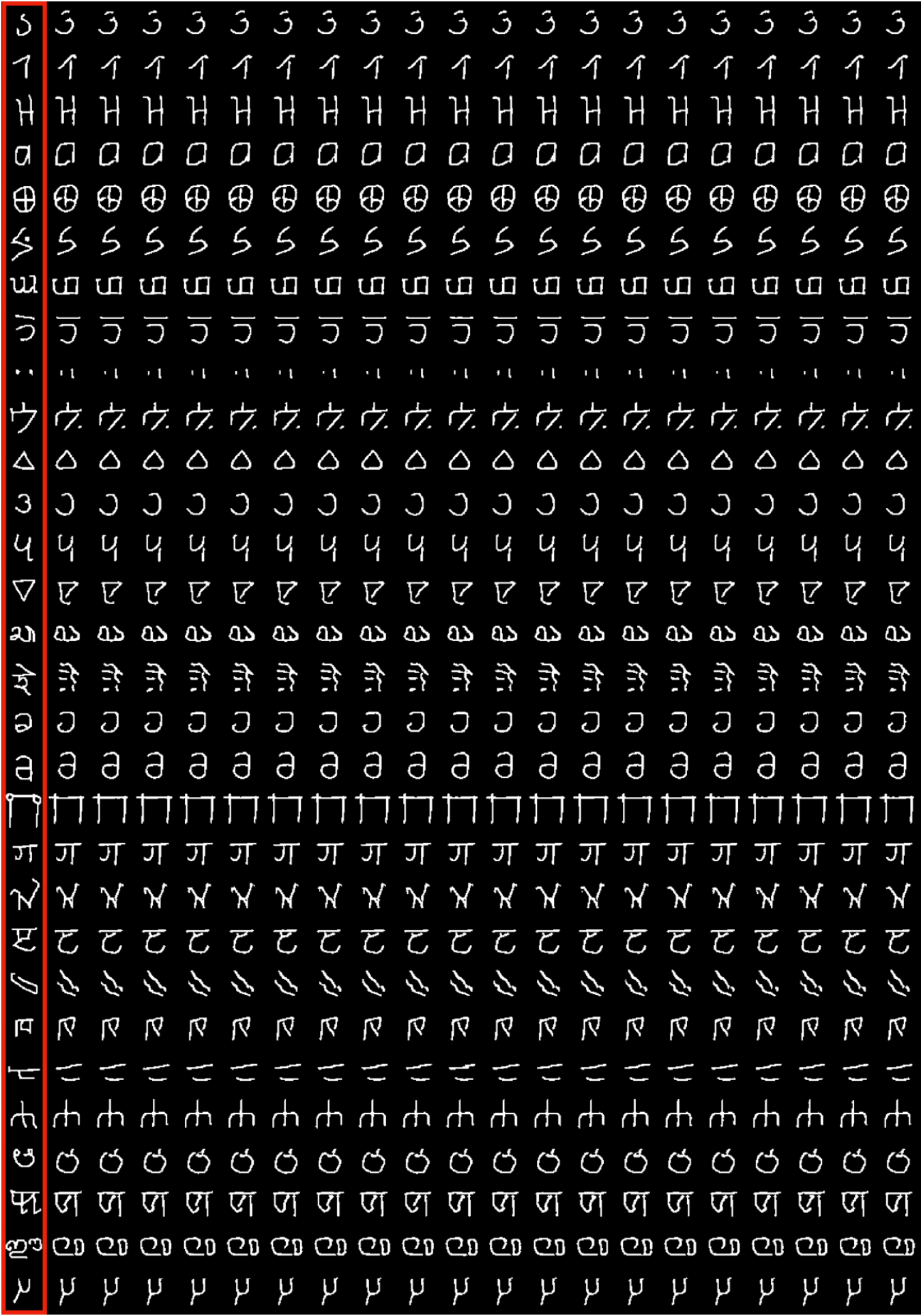}
         \caption{Sampled generated by the Data Augmentation GAN with ResNet architecture (\DAGANRN). All the prototypes used to condition the generative model are in the red frame. The $30$ concepts has been randomly sampled (out of $150$ concepts) from the Omniglot test set. The lines are composed with $20$ samples that has been generated by the DA-GAN-RN.}
         \label{SI:fig9b}
\end{figure}

\newpage
\subsection{Effect of the number of context samples on the diversity/recognizability framework}
\label{SI:effect_context}

\begin{figure}[h!]
\begin{tikzpicture}

\draw [anchor=north west] (0\linewidth, 0.97\linewidth) node {\includegraphics[width=1\linewidth]{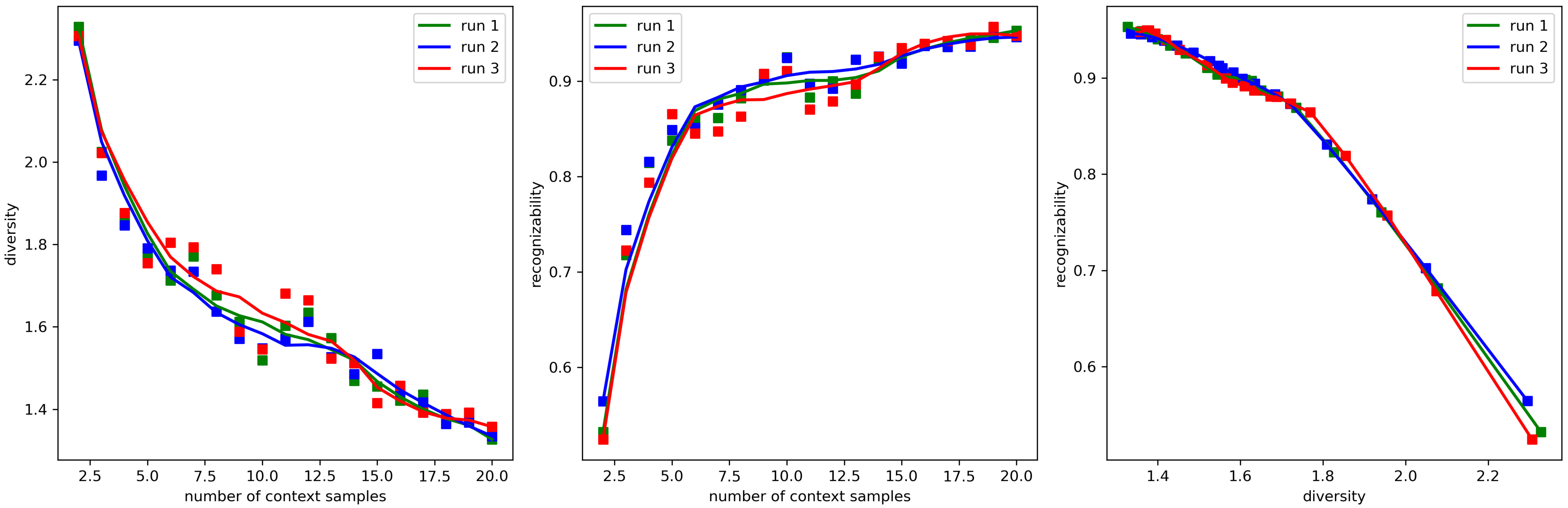}};
\begin{scope}
    \draw [anchor=north west,fill=white, align=left] (0.0\linewidth, 1\linewidth) node {\bf a) };
    \draw [anchor=north west,fill=white, align=left] (0.33\linewidth, 1\linewidth) node {\bf b)};
    \draw [anchor=north west,fill=white, align=left] (0.66\linewidth, 1\linewidth) node {\bf c)};
\end{scope}

\end{tikzpicture}
     \caption{Effect of the number of context samples on the diversity/recognizability framework for $3$ different runs. (\textbf{a}) Effect of the number of context samples on the diversity. (\textbf{b}) Effect of the number of context samples on the recognizability. (\textbf{c}) Simultaneous evolution of diversity and recognizability when ones varies the number of context samples from $2$ to $20$.}
\label{SI:fig_effect_context}
\end{figure}

We observe a monotonic decrease of the diversity and a monotonic increase of the recognizability when the number of context samples increases. We vary the number of context samples from $2$ to $20$. This experiment has been conducted with $3$ different seeds (i.e., different network initialization), represented with red, green and blue data points, respectively. For each seed, we report $19$ data points. To highlight the trend in the diversity-recognizability space, we have smoothed the curves in Fig.~\ref{SI:fig_effect_context}a and Fig.~\ref{SI:fig_effect_context}b, using a Savitzky-Golay filter (second order, window size of 7). 

\subsection{Effect of the number of attentional steps on the diversity/recognizability framework}
\label{SI:effect_attention}
\begin{figure}[h!]
\begin{tikzpicture}

\draw [anchor=north west] (0\linewidth, 0.97\linewidth) node {\includegraphics[width=1\linewidth]{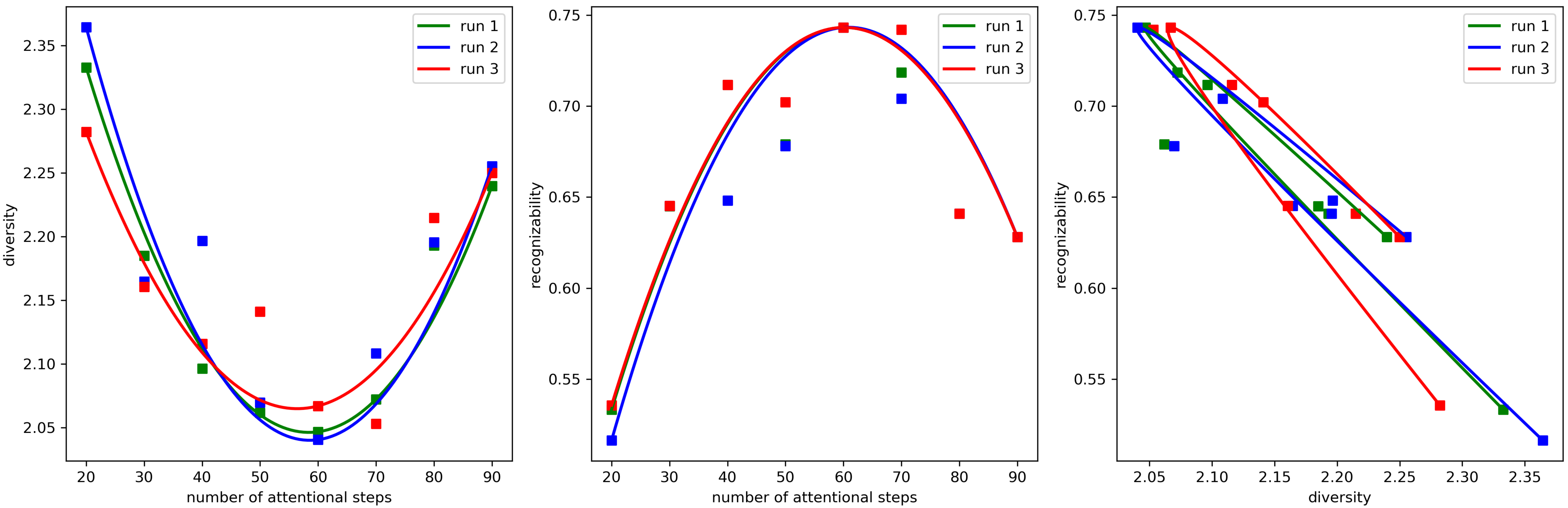}};
\begin{scope}
    \draw [anchor=north west,fill=white, align=left] (0.0\linewidth, 1\linewidth) node {\bf a) };
    \draw [anchor=north west,fill=white, align=left] (0.33\linewidth, 1\linewidth) node {\bf b)};
    \draw [anchor=north west,fill=white, align=left] (0.66\linewidth, 1\linewidth) node {\bf c)};
\end{scope}

\end{tikzpicture}
     \caption{Effect of the number of attentional steps on the diversity/recognizability framework for $3$ different runs. (\textbf{a}) Effect of the number of attentional steps on the diversity. (\textbf{b}) Effect of the number of attentional steps on the recognizability.
     (\textbf{c}) Simultaneous evolution of diversity and recognizability when one varies the number of attentional steps from $20$ to $90$}
\label{SI:fig_effect_attention}
\end{figure}

In this experiment, we have varied the number of attentional steps from $20$ to $90$. Note that we could not go below $20$ attentional steps to make sure the attentional process is fully covering the entire image. We did not go over $90$ attentional steps because we faced some training instabilities beyond this point. We observe a non-monotonic evolution of the diversity and the recognizability with the increase of the number of attentional steps. This experiment has been conducted with $3$ different seeds (i.e., different network initialization), represented with red, green and blue data points, respectively. For each seed we report $8$ data points. In order to properly assess the type of parametric curves that govern the evolution of the diversity-recognizability space when one varies the number of attentional steps, we have used a least curve fitting method~\citep{grossman1971parametric}. This method involves finding the best polynomial fit (second order in our case) for the $3$ curves (Fig.~\ref{SI:fig_effect_attention}a, b and c) simultaneously. This method is iteratively refining all the fits to minimize the sum of all least square error.

\subsection{Mathematical formulation of the ELBO}

Let us consider a dataset $\boldsymbol{X} = \{\boldsymbol{x^{(i)}}\}_{i=1}^{N}$ composed of $N$ i.i.d samples of a random variable $\boldsymbol{x}$. We assume that $\boldsymbol{x}$ is generated by some random process involving an unobserved random variable $\boldsymbol{z}$. The latent variable $\boldsymbol{z}$ is sampled from a Gaussian distribution (see Eq.~\ref{eq:prior}). The mean of the likelihood is parametrized by $\boldsymbol{\mu}_{\theta}$ (in which $\theta$ denotes the parameters) and its variance is considered constant.
\begin{align}
\boldsymbol{x} \sim p_{\theta}(\boldsymbol{x}\mid\boldsymbol{z}) \quad & \mbox{s.t} \quad p_{\theta}(\boldsymbol{x}\mid\boldsymbol{z}) = \mathcal{N}\big(\boldsymbol{x};\boldsymbol{\mu}_{\theta}(\boldsymbol{z}), \boldsymbol{\sigma}^{2}_x\big) \label{eq:likelihood}\\
 \boldsymbol{z} \sim p(\boldsymbol{z}) \quad & \mbox{s.t} \quad p(\boldsymbol{z}) = \mathcal{N}\big(\boldsymbol{z};\boldsymbol{\mu}_p, \boldsymbol{\sigma}^{2}_p\big) \label{eq:prior}
\end{align}

The Variational Auto Encoder is optimized by maximizing the Evidence Lower Bound (ELBO), as formalized in its simplest form in Eq.~\ref{eq:ELBO_common}:

\begin{align}
ELBO(\boldsymbol{x},\theta,\phi) 
&= \mathbb{E}_{q_\phi(\boldsymbol{z} \mid \boldsymbol{x})}[\log p_{\theta}(\boldsymbol{x}\mid \boldsymbol{z})] - \beta \KL\big( q_{\phi}(\boldsymbol{z} \mid \boldsymbol{x}) \lVert p(\boldsymbol{z}) \big) \label{eq:ELBO_common}
\end{align}

One could observe that the $\beta$ coefficient is tuning the importance of the prior (through the KL). If $\beta>1$, then the latent space will be forced to be closer to the prior distribution but will attenuate the weight of the reconstruction loss. Such a scenario tends to improve the disentanglement of the latent space~\citep{higgins2016beta}. On the contrary, if $\beta$ is low, then the reconstruction loss (i.e., $\mathbb{E}_{q_\phi(\boldsymbol{z} \mid \boldsymbol{x})}[\log p_{\theta}(\boldsymbol{x}\mid \boldsymbol{z})]$) will take over, and then the latent space will be less regularized. Note that in the extreme case where $\beta=0$, the VAE becomes an auto-encoder.

The ELBO loss can be updated to include a latent variable encoding for the context $\boldsymbol{c}$ as in the \VAENS. In this formulation, the context corresponds to a dataset $D$ (see Eq.~\ref{eq:ELBO_VAENS}):
\begin{align}
\label{eq:ELBO_VAENS}
ELBO(\boldsymbol{x},\theta,\phi) 
&= \mathbb{E}_{q_\phi(\boldsymbol{c} \mid D)} \Big[\sum_{\boldsymbol{x}\in D} \mathbb{E}_{q_\phi(\boldsymbol{z} \mid \boldsymbol{c} , \boldsymbol{x})}[\log p_{\theta}(\boldsymbol{x}\mid \boldsymbol{z})] - \beta \KL\big( q_{\phi}(\boldsymbol{z} \mid \boldsymbol{c}, \boldsymbol{x}) \lVert p(\boldsymbol{z} \mid \boldsymbol{c}) \big)\Big] \\
& -\KL\big( q_{\phi}(\boldsymbol{z} \mid D) \lVert p(\boldsymbol{c}) \big)\notag
\end{align}

The ELBO could also be extended to include a sequential generative process as in the \VAESTN. In this case, the latent variable $\boldsymbol{z}$ is time-indexed and is now a sequence of random variables denoted $(\boldsymbol{z_1}, .., \boldsymbol{z_T})$. In Eq.~\ref{eq:ELBO_VAESTN}, $\boldsymbol{z_{<k}}$ indicates the collection of all latent variables from step $t=1$ to $t=k$.
\begin{align}
ELBO(\boldsymbol{x},\theta,\phi) &= \mathbb{E}_{q_\phi(\boldsymbol{z_{1}}, .., \boldsymbol{z_{T}}\mid \boldsymbol{x})}[\log p_{\theta}(\boldsymbol{x}\mid \boldsymbol{z_{1}}, .., \boldsymbol{z_{T}})] - \beta \KL \sum_{k=1}^{T}\big( q_{\phi}(\boldsymbol{z_{k}} \mid \boldsymbol{z_{<k}}, \boldsymbol{x}) \lVert p(\boldsymbol{z_{k}}) \big)
\label{eq:ELBO_VAESTN}
\end{align}

\subsection{Effect of the beta coefficient on the  diversity/recognizability framework}
\label{SI:effect_beta}
\begin{figure}[h!]
\begin{tikzpicture}

\draw [anchor=north west] (0\linewidth, 0.97\linewidth) node {\includegraphics[width=1\linewidth]{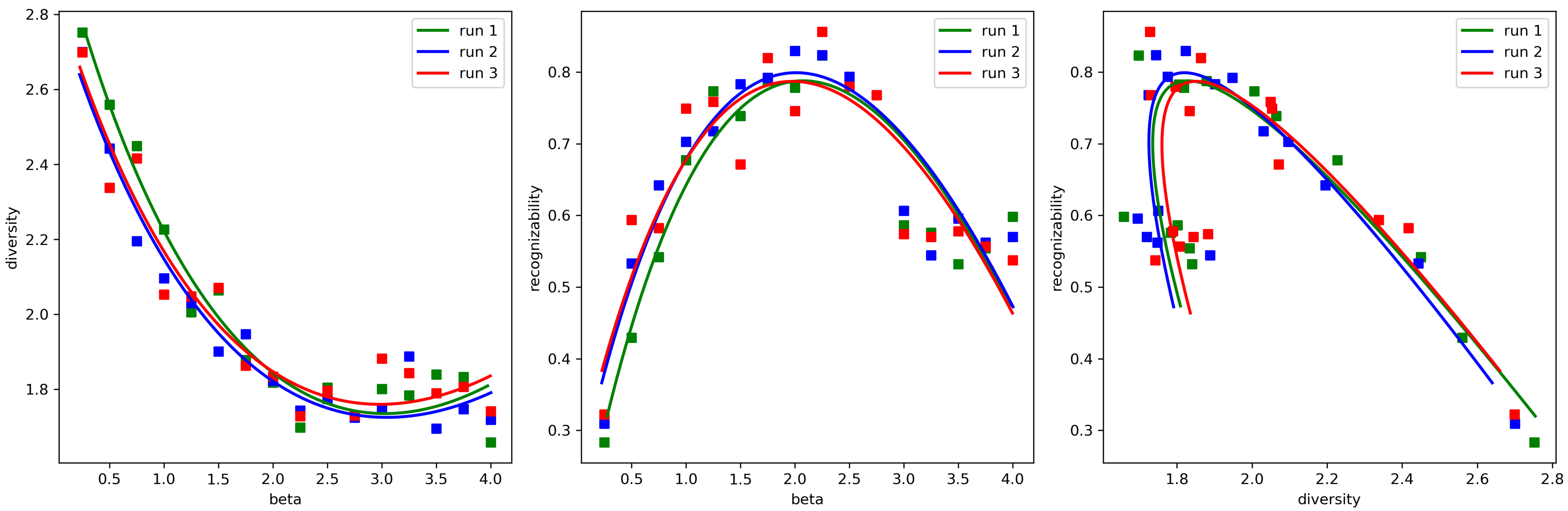}};
\begin{scope}
    \draw [anchor=north west,fill=white, align=left] (0.0\linewidth, 1\linewidth) node {\bf a) };
    \draw [anchor=north west,fill=white, align=left] (0.33\linewidth, 1\linewidth) node {\bf b)};
    \draw [anchor=north west,fill=white, align=left] (0.66\linewidth, 1\linewidth) node {\bf c)};
\end{scope}

\end{tikzpicture}
     \caption{Effect of varying $\beta$ in the \VAESTN~on the diversity/recognizability framework for $3$ different runs. (\textbf{a}) Effect of $\beta$ on the diversity. (\textbf{b}) Effect of $\beta$ on the recognizability. (\textbf{c}) Parametric curve recognizability versus diversity when one varies $\beta$ from $0.25$ from to $4$.}
\label{SI:fig_effect_beta_vae_stn}
\end{figure}

\begin{figure}[h!]
\begin{tikzpicture}

\draw [anchor=north west] (0\linewidth, 0.97\linewidth) node {\includegraphics[width=1\linewidth]{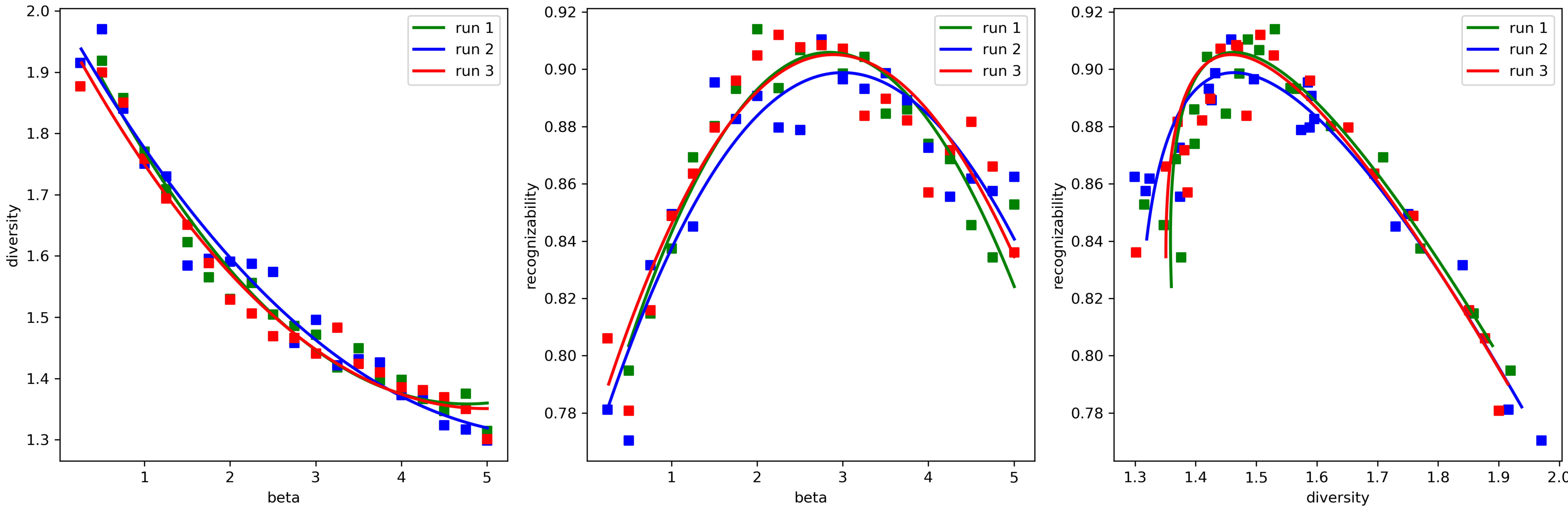}};
\begin{scope}
    \draw [anchor=north west,fill=white, align=left] (0.0\linewidth, 1\linewidth) node {\bf a) };
    \draw [anchor=north west,fill=white, align=left] (0.33\linewidth, 1\linewidth) node {\bf b)};
    \draw [anchor=north west,fill=white, align=left] (0.66\linewidth, 1\linewidth) node {\bf c)};
\end{scope}

\end{tikzpicture}
     \caption{Effect of varying $\beta$ in the \VAENS~on the diversity/recognizability framework for $3$ different runs. (\textbf{a}) Effect of $\beta$ on the diversity. (\textbf{b}) Effect of $\beta$ on the recognizability. (\textbf{c}) Parametric curve recognizability versus diversity when one varies $\beta$ from $0.25$ from to $5$.}
\label{SI:fig_effect_beta_ns}
\end{figure}

In this experiment, we have varied the value of the $\beta$ coefficient from $0.25$ to $4$ for the \VAESTN~and from $0.25$ to $5$ for \VAENS~model. This experiment has been conducted with $3$ different seeds (i.e., different network initialization), represented with red, green and blue data points, respectively. For the \VAESTN~and for each seed, we have collected $16$ data points (see Fig.~\ref{SI:fig_effect_beta_vae_stn}), and $20$ for the \VAENS~(see Fig.~\ref{SI:fig_effect_beta_ns}). We use a similar method than in \ref{SI:effect_attention} to find a polynomial fit (second order in our case) of the curves shown in Fig.~\ref{SI:fig_effect_beta_vae_stn}a, b, and c and Fig.~\ref{SI:fig_effect_beta_ns}a, b, and c. We report a quasi-monotonic decline of the diversity when the beta value is increased (see Fig.~\ref{SI:fig_effect_beta_vae_stn}a and Fig.~\ref{SI:fig_effect_beta_ns}a). In contrast, the recognizability follows a parabolic relationship when the beta value is increased. For the \VAESTN, the maximum recognizability ($\approx80\%$) is reached for a $\beta$ value of $2.25$ (see Fig.~\ref{SI:fig_effect_beta_vae_stn}b). For the \VAENS, the maximum recognizability ($\approx91\%$) is reached for a $\beta$ value of $3$ (see Fig.~\ref{SI:fig_effect_beta_ns}b). Even if the change of amplitude in recognizability and in diversity is larger for the \VAESTN~than for \VAENS, the shapes of the curves are very similar.

\newpage
\subsection{Effect of the size of the latent space on the diversity/recognizability framework}
\label{SI:effect_z}

\begin{figure}[h!]
\begin{tikzpicture}

\draw [anchor=north west] (0\linewidth, 0.97\linewidth) node {\includegraphics[width=1\linewidth]{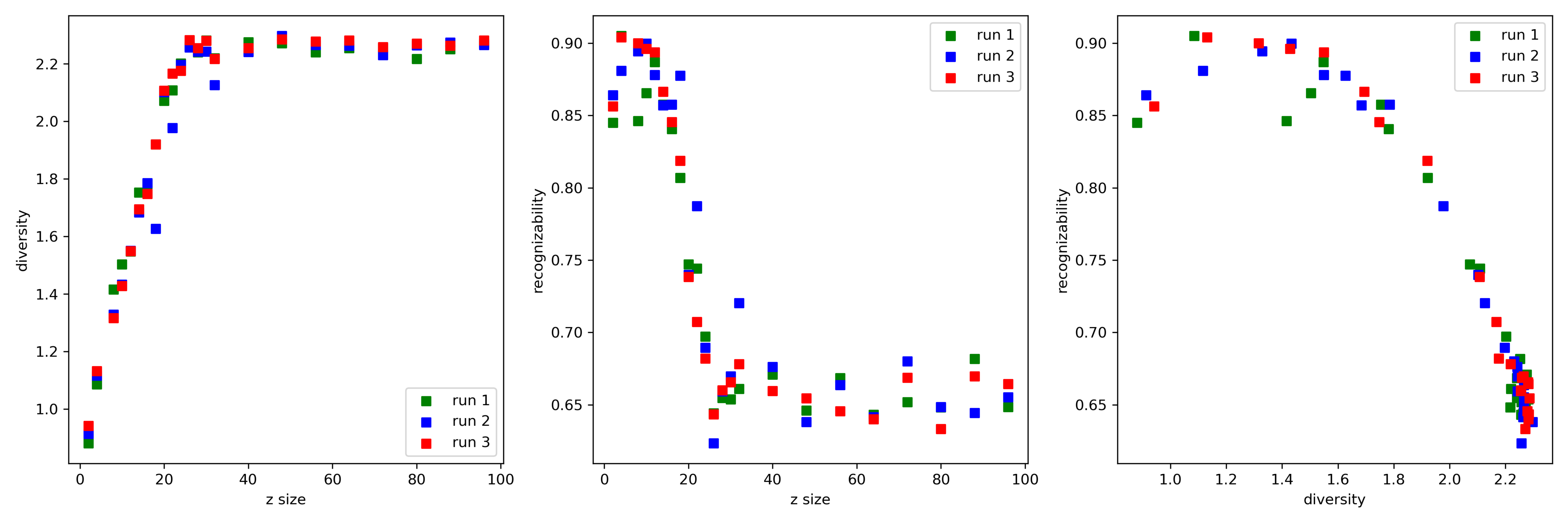}};
\begin{scope}
    \draw [anchor=north west,fill=white, align=left] (0.0\linewidth, 1\linewidth) node {\bf a) };
    \draw [anchor=north west,fill=white, align=left] (0.33\linewidth, 1\linewidth) node {\bf b)};
    \draw [anchor=north west,fill=white, align=left] (0.66\linewidth, 1\linewidth) node {\bf c)};
\end{scope}

\end{tikzpicture}
     \caption{Effect of varying the size of the latent vector ($z$) in the \VAENS~on the diversity/recognizability framework for $3$ different runs. (\textbf{a}) Effect of latent size on the diversity. (\textbf{b}) Effect of the latent size on the recognizability. (\textbf{c}) Parametric curve recognizability versus diversity when one varies $\beta$ from $5$ from to $100$.
     }
\label{SI:fig_effect_z_vaens}
\end{figure}

\begin{figure}[h!]
\begin{tikzpicture}

\draw [anchor=north west] (0\linewidth, 0.97\linewidth) node {\includegraphics[width=1\linewidth]{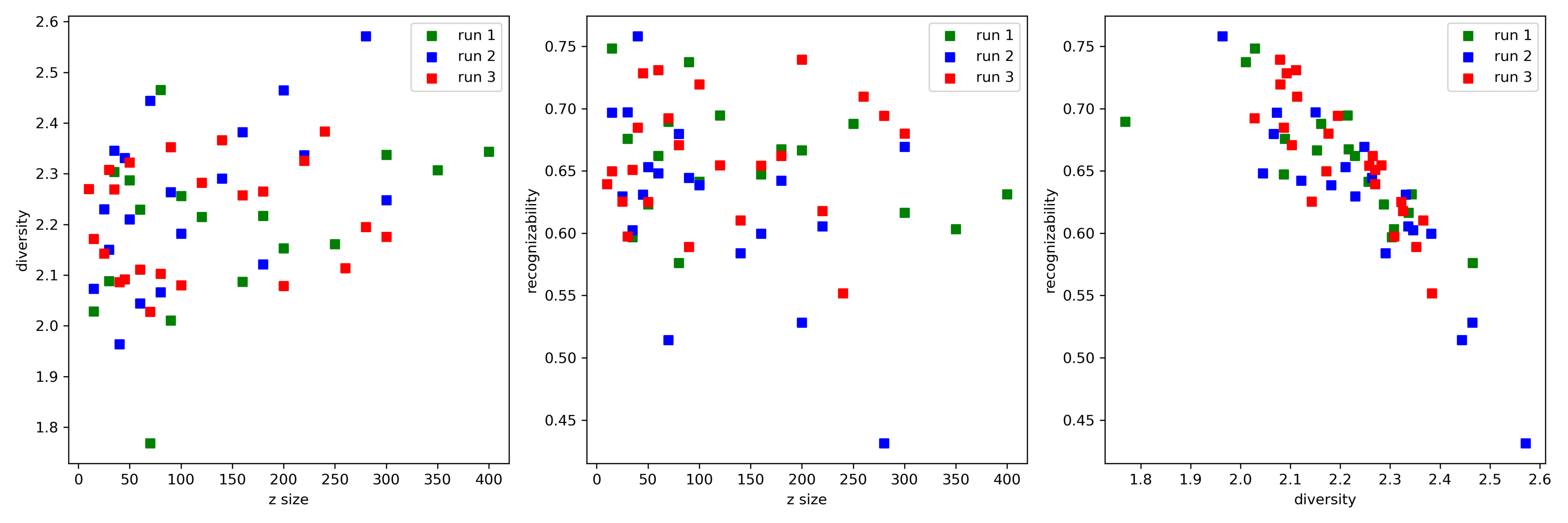}};
\begin{scope}
    \draw [anchor=north west,fill=white, align=left] (0.0\linewidth, 1\linewidth) node {\bf a) };
    \draw [anchor=north west,fill=white, align=left] (0.33\linewidth, 1\linewidth) node {\bf b)};
    \draw [anchor=north west,fill=white, align=left] (0.66\linewidth, 1\linewidth) node {\bf c)};
\end{scope}

\end{tikzpicture}
     \caption{Effect of varying the size of the latent vector ($z$) in the \VAESTN~on the diversity/recognizability framework for $3$ different runs. (\textbf{a}) Effect of latent size on the diversity. (\textbf{b}) Effect of the latent size on the recognizability. (\textbf{c}) Parametric curve recognizability versus diversity when one varies $\beta$ from $5$ from to $400$.
     }
\label{SI:fig_effect_z_vae_stn}
\end{figure}

\begin{figure}[h!]
\begin{tikzpicture}

\draw [anchor=north west] (0\linewidth, 0.97\linewidth) node {\includegraphics[width=1\linewidth]{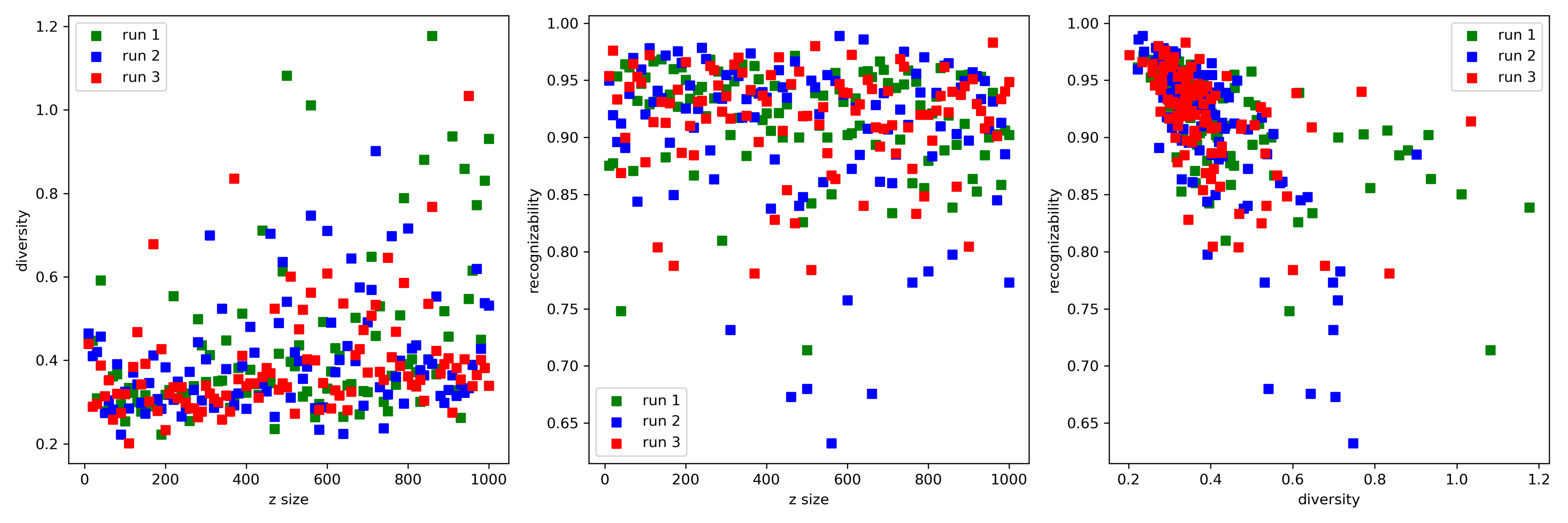}};
\begin{scope}
    \draw [anchor=north west,fill=white, align=left] (0.0\linewidth, 1\linewidth) node {\bf a) };
    \draw [anchor=north west,fill=white, align=left] (0.33\linewidth, 1\linewidth) node {\bf b)};
    \draw [anchor=north west,fill=white, align=left] (0.66\linewidth, 1\linewidth) node {\bf c)};
\end{scope}

\end{tikzpicture}
     \caption{Effect of varying the size of the latent vector ($z$) in the \DAGANUN~on the diversity/recognizability framework for $3$ different runs. (\textbf{a}) Effect of latent size on the diversity. (\textbf{b}) Effect of the latent size on the recognizability. (\textbf{c}) Parametric curve recognizability versus diversity when one varies $\beta$ from $10$ from to $1000$.
     }
\label{SI:fig_effect_z_dagan_un}
\end{figure}

\begin{figure}[h!]
\begin{tikzpicture}

\draw [anchor=north west] (0\linewidth, 0.97\linewidth) node {\includegraphics[width=1\linewidth]{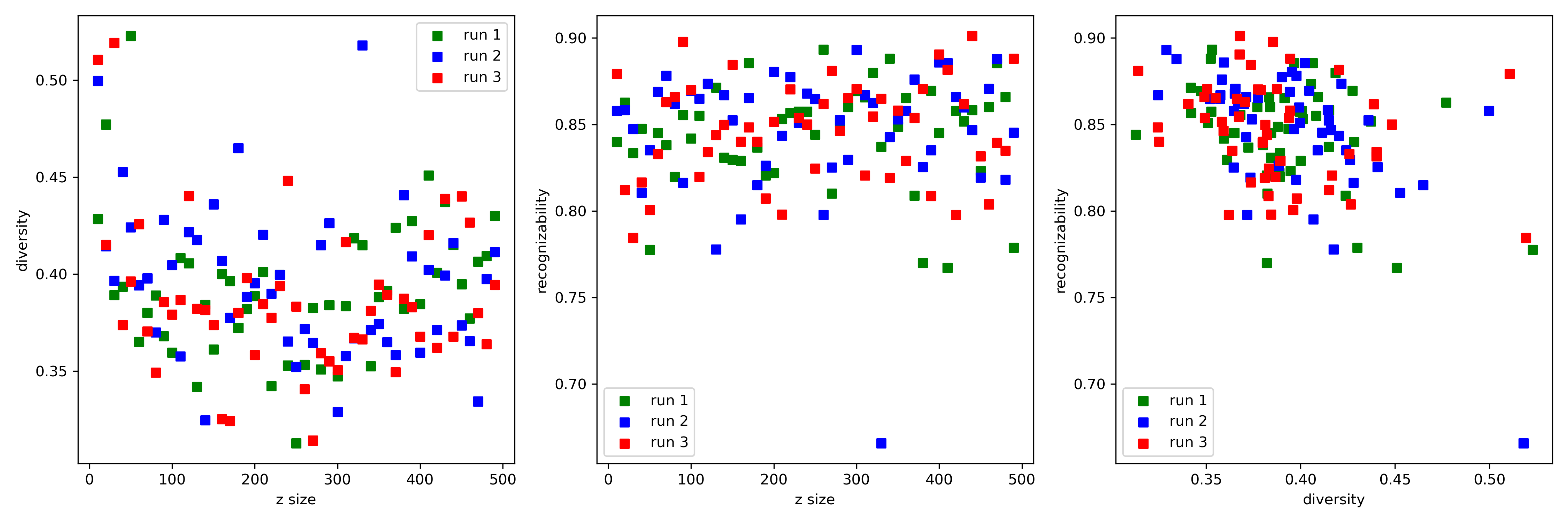}};
\begin{scope}
    \draw [anchor=north west,fill=white, align=left] (0.0\linewidth, 1\linewidth) node {\bf a) };
    \draw [anchor=north west,fill=white, align=left] (0.33\linewidth, 1\linewidth) node {\bf b)};
    \draw [anchor=north west,fill=white, align=left] (0.66\linewidth, 1\linewidth) node {\bf c)};
\end{scope}

\end{tikzpicture}
     \caption{Effect of varying the size of the latent vector ($z$) in the \DAGANRN~on the diversity/recognizability framework for $3$ different runs. (\textbf{a}) Effect of latent size on the diversity. (\textbf{b}) Effect of the latent size on the recognizability. (\textbf{c}) Parametric curve recognizability versus diversity when one varies $\beta$ from $10$ from to $500$.
     }
\label{SI:fig_effect_z_dagan_rn}
\end{figure}








\newpage
\subsection{Overfitting of standard classifier in low-data regime}\label{SI:Overfitting}
Omniglot is a dataset composed of images representing 1,623 classes of handwritten letters and symbols (extracted from 50 different alphabets) with just 20 samples per class. This low number of samples per class makes Omniglot very different from other datasets (e.g. MNIST, CIFAR10...). In such a low-data regime, standard deep learning classifiers are known to overfit to the training data~\citep{brigato2021close} resulting in poor generalization performance. In this section we provide experimental confirmation of such a phenomenon.

\begin{figure}[h!]
\begin{tikzpicture}

\draw [anchor=north west] (0\linewidth, 0.97\linewidth) node {\includegraphics[width=1\linewidth]
{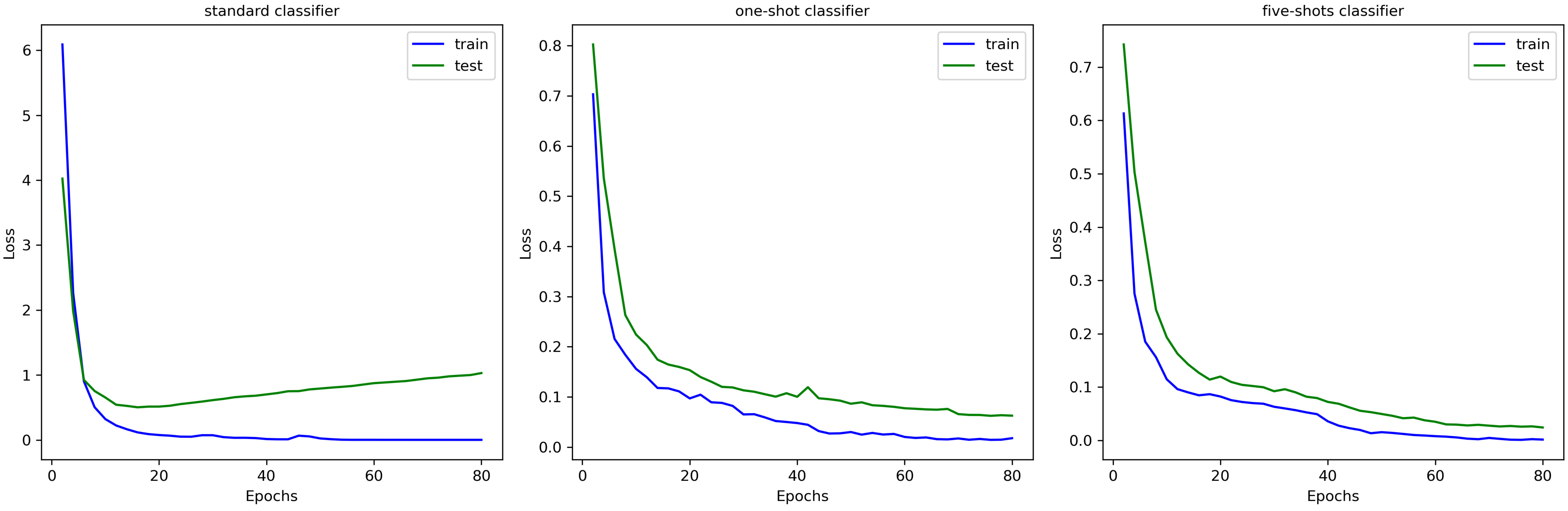}};
\draw [anchor=north west] (0\linewidth, 0.6\linewidth) node {\includegraphics[width=1\linewidth]
{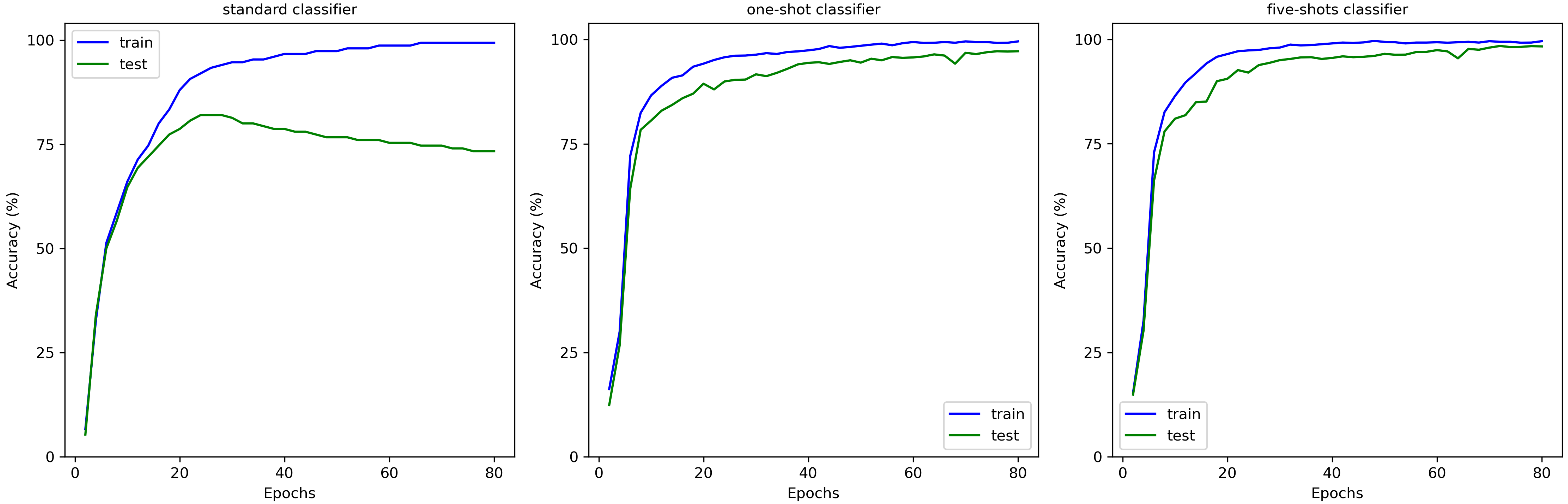}};
\begin{scope}
    \draw [anchor=north west,fill=white, align=left] (0.0\linewidth, 1\linewidth) node {\bf a) };
    \draw [anchor=north west,fill=white, align=left] (0.33\linewidth, 1\linewidth) node {\bf b)};
    \draw [anchor=north west,fill=white, align=left] (0.66\linewidth, 1\linewidth) node {\bf c)};
    \draw [anchor=north west,fill=white, align=left] (0.0\linewidth, 0.63\linewidth) node {\bf d) };
    \draw [anchor=north west,fill=white, align=left] (0.33\linewidth, 0.63\linewidth) node {\bf e) };
    \draw [anchor=north west,fill=white, align=left] (0.66\linewidth, 0.63\linewidth) node {\bf f) };
\end{scope}

\end{tikzpicture}
     \caption{Comparison between different classifiers in low-data regime. Train and test losses at each training epoch for (\textbf{a}) a standard classifier,  (\textbf{b}) a Protypical Net in a one-shot learning setting and (\textbf{c}) a Prototypical Net in a 5-shots learning setting. Train and test classification accuracy at each training epoch for (\textbf{d}) a standard classifier,  (\textbf{e}) a Protypical Net in a one-shot learning setting and (\textbf{f}) a Prototypical Net in a 5-shots learning setting.}
\label{SI:Overfitting_lowdata}
\end{figure}

We have trained 3 different classifiers, all having a similar architecture (the architecture is described in Table~\ref{SI:table1}):
\begin{itemize}
\item \textbf{A standard classifier}. For this classifier, the last linear layer has been changed to have an output activation of size $1623$. Said differently, the layer entitled "Linear(256, 128)" in Table~\ref{SI:table1} has been replaced by "Linear(256, 1623)". We have trained this classifier using $18$ samples per class of the Omniglot dataset.  The testing set is composed of the $2$ remaining samples per class. To summarize, the training set is composed of $29,214$ samples ($1623\times18$) and the training set is composed of $3246$ samples ($1623\times2$). This classifier is trained using a standard back-propagation on a cross-entropy loss (same learning parameters than those described in Section~\ref{SI:ProtoNet}). Train/test loss and classification accuracy are reported for all training epochs in Fig~\ref{SI:Overfitting_lowdata}a and Fig~\ref{SI:Overfitting_lowdata}d, respectively.
\item \textbf{A one-shot classifier}. Both the architecture and the training procedure of this classifier are described in Section~\ref{SI:ProtoNet}. We remind the reader that we use a weak generalization split to train the few-learning networks (i.e. $1473$ classes in the training set and $150$ classes of testing set). Train/test loss and classification accuracy are reported for all training epochs in Fig~\ref{SI:Overfitting_lowdata}b and Fig~\ref{SI:Overfitting_lowdata}e, respectively.
\item \textbf{A five-shot classifier}. This network is the exact same than the one-shot Prototypical Net described before, except that it is trained in a 5-shots settings. Train/test loss and classification accuracy are reported for all training epochs in Fig~\ref{SI:Overfitting_lowdata}c and Fig~\ref{SI:Overfitting_lowdata}f, respectively.
\end{itemize}

For the standard classifier, we observe an increase of the test loss (resp. a decrease of the test accuracy) while the train loss is still decreasing (resp. the train accuracy is still increasing), see Fig~\ref{SI:Overfitting_lowdata}a and Fig~\ref{SI:Overfitting_lowdata}d. It suggests that the network becomes better at classifying the training samples but worst at dealing with the testing samples. The standard classifier is then overfitting on the training set.  Note that the $2$ other few-shots learning networks are not showing such a decrease in the test loss and accuracy. Such an experiment suggests that standard classifiers are not adequate to extract features of samples in a low-data regime.

\subsection{Computational Resources}
\label{SI:computational_resources}
All the experiments of this paper have been performed using Tesla V100 with 16 Gb memory. The training time is dependent on the hyper-parameters, but varies between 4h to 24h per simulation.

\subsection{Broader Impact}
\label{SI:Broader_Impact}
This work does not present any foreseeable negative societal consequences.  We think the societal impact of this work is positive. It might help the neuroscience community to evaluate the different mechanisms that allow human-level generalization, and then better understand the brain. 

\end{document}